%% file: thesis.tex
\documentclass[12pt]{ociamthesis}
\pdfoutput=1
\input{preamb/packages}

\input{preamb/macros}

\input{preamb/title}

\AtBeginDocument{}

\begin{document}
\baselineskip=22pt plus1pt

\setcounter{secnumdepth}{3}
\setcounter{tocdepth}{2}

\input{preamb/frontmatter}

\subincludefrom{intro/}{intro}

\subincludefrom{ch-bg/}{ch-bg}

\subincludefrom{ch-hpylm/}{ch-hpylm}

\subincludefrom{ch-dist/}{ch-dist}

\subincludefrom{ch-pysrcag/}{ch-pysrcag}

\subincludefrom{conclusions/}{conclusions}

\input{preamb/backmatter}

\end{document}

%% file: preamb/packages.tex
\usepackage[utf8]{inputenc} 
\usepackage[T1]{fontenc}    
\usepackage{times}

\usepackage{etex}      

\usepackage{amsthm}    
\usepackage{amsmath}   
\usepackage{amsfonts}  
\usepackage{amssymb}   
\usepackage{bm}        

\usepackage[english]{babel} 
\usepackage[shortcuts]{extdash} 

\usepackage{fixltx2e}
\usepackage{import}
\usepackage{algpseudocode}
\usepackage{algorithm}

\usepackage{multicol,multirow,array} 

\usepackage{booktabs}   
\usepackage{tabularx}   

\usepackage{setspace} 

\usepackage{xspace}

\usepackage[pdftex]{graphicx} 

\usepackage[svgnames]{xcolor}

\usepackage{pgfplotstable}
\usepackage{subcaption} 

\usepackage{tikz}
\usepackage{tikz-qtree}
\usetikzlibrary{positioning,chains,calc}

\usepackage{ragged2e} 
\usepackage{footnote} 

\usepackage[nottoc,notlot,notlof]{tocbibind} 

\usepackage[round]{natbib} 

\usepackage{hyperref}
\hypersetup{%
  colorlinks=false,       
  pdfborder={0 0 0},      
  citecolor=NavyBlue,     
  linkcolor=Maroon,       
  urlcolor=NavyBlue,      
  linktoc=all,            
  pdftitle={Probabilistic Modelling of Morphologically Rich Languages},
  pdfauthor={Jan Abraham Botha},
  pdfdisplaydoctitle,         
  pdffitwindow,
  bookmarksopen,              
  bookmarksopenlevel=1,
  plainpages=false,           
}

%% file: preamb/macros.tex
\newcommand{\captionfonts}{\small}

\makeatletter  
\long\def\@makecaption#1#2{%
  \vskip\abovecaptionskip
  \sbox\@tempboxa{{\captionfonts #1: #2}}%
  \ifdim \wd\@tempboxa >\hsize
    {\captionfonts #1: #2\par}
  \else
    \hbox to\hsize{\hfil\box\@tempboxa\hfil}%
  \fi
  \vskip\belowcaptionskip}
\makeatother   

\let\oldmarginpar\marginpar
\renewcommand\marginpar[1]{
	\-\oldmarginpar[\raggedleft\footnotesize #1]%
	{\raggedright\footnotesize #1}
	}

\def\SpaceGobblingS~{\S{}} 

\DeclareMathOperator*{\argmax}{arg\,max}
	
	\renewcommand{\th}{\textsuperscript{th}\hspace{.25em}} 

\renewcommand{\cite}[1]{\citep{#1}}

\newcommand{\chapoverview}[1]{
    \begin{center}%
      \bfseries Chapter Abstract
    \end{center}\begin{quote}\begin{itshape}#1
   \end{itshape}\end{quote}\begin{center}%
      \par%
      \noindent\rule[0.5ex]{0.75\linewidth}{1pt}%
     \end{center}}
 
\newcommand*{\ie}{i.e.\@\xspace}
\newcommand*{\eg}{e.g.\@\xspace}
\newcommand*{\cf}{cf.\@\xspace}

\newcommand*{\posphrase}{part-of-speech\xspace}

\newcommand{\chref}[1]{Chapter~\ref{#1}}

\newcommand{\mytilde}{\raise.17ex\hbox{$\scriptstyle\mathtt{\sim}$}}

\newcommand*{\iteg}[1]{\emph{#1}}  
\newcommand*{\iten}[1]{(`#1')}     

\newcommand\p[1]{\ensuremath{P\!\left(#1\right)}}
\newcommand\psub[2]{\ensuremath{P_{#1}\!\left(#2\right)}}

\newcommand{\mb}[1]{\mathbf{#1}}

\newcommand{\BigO}[1]{\ensuremath{\mathcal{O}(#1)}}

\newcommand*\tokseq[3]{\ensuremath{#1^{#3}_{#2}}}

\newcommand{\roottri}[3]{#1$\cdot$#2$\cdot$#3}         
\newcommand{\range}[2]{\ensuremath{\langle #1..#2 \rangle}}
\newcommand{\ag}{PYSRCAG\xspace} 
\newcommand{\agcfg}{PY-CFAG\xspace} 
\newcommand{\gram}[1]{\ensuremath{\mathcal{#1}}} 

\newcommand{\model}[1]{{\small\textbf{\textsf{#1}}}}
\newcommand{\cfgmulti}{\model{MConcat}\xspace}
\newcommand{\cfgone}{\model{Concat}\xspace}
\newcommand{\tplquad}{\model{TplR4}\xspace}
\newcommand{\tplonetrilit}{\model{Tpl}\xspace}
\newcommand{\tplmultitrilit}{\model{MTpl}\xspace}
\newcommand{\tplmultitrilitTtoThree}{\model{MTpl3Ch}\xspace}
\newcommand{\tplonetrilitTtoThree}{\model{Tpl3Ch}\xspace}
\newcommand{\tplonetrilitTFour}{\model{Tpl+T4}\xspace}
\newcommand{\tplmultitrilitTFour}{\model{MTpl+T4}\xspace}
\newcommand{\morfessor}{\model{Morfessor}\xspace}

\newcommand{\data}[1]{\textsc{#1}}
\newcommand{\bw}{\data{Bw}\xspace}
\newcommand{\bwvoc}{\data{Bw}\ensuremath{^{'}}\xspace}

\newcommand{\qu}{\data{Qu}\ensuremath{^{'}}\xspace}       

\newcommand{\heb}{\data{Heb}\xspace}

\newcommand{\word}[2]{\emph{#1} (#2)}
\newcommand{\arrowg}{\ensuremath{\rightarrow}}
\newcommand{\nt}[1]{\ensuremath{\text{#1}}}   
\newcommand{\nta}[1]{\ensuremath{\text{\underline{#1}}}}  
\newcommand{\rank}{\ensuremath{\rho}}
\newcommand{\fan}{\ensuremath{\psi}}

\newcommand\morflabel[1]{\textsubscript{\it #1}}
\newcommand\substem{\morflabel{STM}\xspace}
\newcommand\subsuf{\morflabel{SUF}\xspace}
\newcommand\subpre{\morflabel{PRE}\xspace}

\newcommand\GRT[1]{\colorbox{black!20}{#1}} 
\newcommand\HRT[1]{{\bf\color{black!50}#1}}      
\newcommand\HGRT[1]{\textbf{#1}}   
\def\segFP{{\scriptsize$\otimes$}}            
\def\segTP{\textvisiblespace}                        
\def\segFN{$_\uparrow$}            

\newcommand{\ngram}{\mbox{$n$-gram}\xspace}
\newcommand{\ngrams}{\mbox{$n$-grams}\xspace}
\newcommand{\kgram}[1]{\mbox{$#1$-gram}}
\newcommand{\tpr}[1]{\textsl{#1}} 

\newcommand*{\textvector}[1]{\ensuremath{\overrightarrow{\mathrm{#1}}}}

\newcommand{\Set}[1]{\ensuremath{\mathcal{#1}}}
\newcommand{\V}{\Set{V}\xspace}
\newcommand{\F}{\Set{F}\xspace}

\newcommand{\unk}{\textsc{Unk}\xspace}

\newcommand*{\zL}{L}  
\newcommand{\wvoc}{\mathcal{W}}
\newcommand{\mvoc}{\mathcal{M}}

\newcommand*{\pyp}{\ensuremath{\mathcal{PY}}}

\newcommand{\Dmodel}[1]{{\small\Dmodelx{#1}}}
\newcommand{\Dmodelx}[1]{\textsf{#1}}
\newcommand{\lblpp}{\Dmodel{LBL++}\xspace}
\newcommand{\lblpc}{\Dmodel{LBL+c}\xspace}
\newcommand{\lblpo}{\Dmodel{LBL+o}\xspace}
\newcommand{\lbl}{\Dmodel{LBL}\xspace}

\newcommand{\clbl}{\Dmodel{CLBL}\xspace}
\newcommand{\clblpp}{\Dmodel{CLBL++}\xspace}
\newcommand{\clblpc}{\Dmodel{CLBL+c}\xspace}
\newcommand{\clblpo}{\Dmodel{CLBL+o}\xspace}

\newcommand{\nlbl}{\Dmodel{nLBL}\xspace}
\newcommand{\nclbl}{\Dmodel{nCLBL}\xspace}

\newcommand{\mkn}{\Dmodel{MKN}\xspace}

\newcommand{\Lcs}{\data{Cs}\xspace}
\newcommand{\Lde}{\data{De}\xspace}
\newcommand{\Les}{\data{Es}\xspace}
\newcommand{\Len}{\data{En}\xspace}
\newcommand{\Lfr}{\data{Fr}\xspace}
\newcommand{\Lru}{\data{Ru}\xspace}

\newcommand{\corpus}[1]{{\small\textsf{#1}}}
\newcommand{\corpusx}[1]{{\footnotesize\textsf{#1}}}
\newcommand{\tool}[1]{\textsl{#1}}
\newcommand{\bleu}{\textsc{Bleu}\xspace}

\newcommand{\lp}[2]{\data{#1}$\rightarrow$\data{#2}}

\newcommand{\rhotimes}{\mbox{$\rho\hspace{-2pt}\times\hspace{-2pt}100$}\xspace}

\newcommand{\compose}{\emph{compose}\xspace}

\pgfplotsset{
  /pgfplots/table/addPercToNonEmpty/.style={%
    postproc cell content/.append code={ 
          \pgfkeysgetvalue{/pgfplots/table/@preprocessed cell content}\pgfmathresult%
          \ifx\pgfmathresult\empty\relax
          \else%
          \pgfkeys{/pgfplots/table/@cell content/.add={}{\%}}%
          \fi }%
  }%
}

\newcommand{\RID}[1]{{\ensuremath{(\text{#1})}}}  
\newcommand{\mRID}{{\ensuremath{(\emptyset)}}}  

\newcommand{\HHist}[1]{\text{Hist}_{#1}}
\newcommand{\Hist}[2]{\HHist{#1}[#2]}
\newcommand{\lstar}{_{\ast}}

\makeatletter
\newcommand*\betweenfootnoteandscriptsize{%
    \@setfontsize\betweenfootnoteandscriptsize{9}{11.0}%
}
\makeatother

%% file: preamb/title.tex
\title{Probabilistic Modelling of Morphologically Rich Languages}
\author{Jan Abraham Botha}
\college{Lincoln College}
\degree{Doctor of Philosophy}
\degreedate{Hilary Term 2014}
\renewcommand{\crest}{\beltcrest}

%% file: preamb/frontmatter.tex

\begin{frontpages} 

\maketitle

\begin{acknowledgements}
  \input{text/acknowledgements}

\end{acknowledgements}

\begin{abstract}
	\input{text/abstract}

\end{abstract}

\end{frontpages}
	
\begin{romanpages}
\begin{singlespace}
\tableofcontents
\listoffigures
\listoftables
\end{singlespace}
\end{romanpages}


%% file: text/acknowledgements.tex
\begingroup
\setlength{\parskip}{6pt}
Thanks go to my main supervisor, Phil Blunsom, and to Stephen Pulman, who remained an available secondary supervisor. To Phil, I am especially grateful for two things he provided: 1) Continued patience in laying out subtle intricacies of machine learning, mixed with nudges towards interesting applications in language processing; 2) Generous exposure to his own collaborators during the early days when I was new to the NLP field. In this regard, I am also thankful to the input Chris Dyer provided at an early stage and as co-author, which helped transform my thitherto generic explorations into something more concrete.

This project was funded in part by the Rhodes Trust, and I benefited from travel grants from Lincoln College, which aided conference attendance. I would like to acknowledge three administrators in particular who made the papermaze of Oxford’s bureaucracy much more tolerable through the dedicated and pleasant way in which they supported my efforts: Julie Sheppard, Carmella Elan-Gaston and Mary Eaton.

Thanks to my friends and various housemates in Oxford who made the time there so enriching and often unbelievable. In particular, thanks to Bronwyn and Alysia for frequently providing a place to stay in Oxford during the last six months after I moved away.

I give special thanks to my family for their continued support of my endeavours, and for nurturing such a special appreciation of language and languages at home. Thanks to Saskia for her love and support throughout, across continents and Channels, and not across them.
\endgroup

%% file: text/abstract.tex
\begingroup
\setlength{\parskip}{6pt}
This thesis investigates how the sub-structure of words can be accounted for in 
 probabilistic models of language.  Such models
play an important role in natural language processing tasks such as translation
or speech recognition, but often rely on the simplistic assumption that words
are opaque symbols. This assumption does not fit morphologically complex
language well, where words can have rich internal structure and sub-word
elements are shared across distinct word forms.

Our approach is to encode basic notions of morphology into the assumptions of
three different types of language models, with the intention
that leveraging shared sub-word structure can improve model performance and
help overcome data sparsity that arises from morphological processes.

In the context of n-gram language modelling, we formulate a new Bayesian model
that relies on the decomposition of compound words to attain better smoothing,
and we develop a new distributed language model that learns vector
representations of morphemes and leverages them to link together morphologically
related words.  In both cases, we show that accounting for word sub-structure
improves the models' intrinsic performance and provides benefits when applied to
other tasks, including machine translation. 

We then shift the focus beyond the modelling of word sequences and consider
models that automatically learn \emph{what} the
sub-word elements of a given language are, given an unannotated list of words.
We formulate a novel model that can learn discontiguous morphemes in addition to
the more conventional contiguous morphemes that most previous models are limited to.
This approach is demonstrated on Semitic languages, and we find that modelling
discontiguous sub-word structures leads to improvements in the task of
segmenting words into their contiguous morphemes.

\endgroup

%% file: intro/intro.tex
\chapter{\label{ch:introduction}Introduction} 
People can understand sentences and words they have never heard before. We do this by
combining the interpretations of parts of the sentence, such as individual words
or phrases, or parts of words, into an interpretation of the whole. Two key
parts of this
compositional view of language is that the smaller meaningful entities are
easily identifiable and that their meanings are likely already known.
To illustrate this, consider how immediately the following sentence is
understandable to an English speaker:
\begin{itemize}
  \item[] \emph{The king finally abdicated after years of unkingly conduct.}
\end{itemize}
This sentence is understandable despite the unlikeliness that one will
have previously encountered the word \iteg{unkingly}. So not only is the sentence
interpretable as a function of the words it contains, but the word
\iteg{unkingly} is itself
interpretable as a function of the morphemes it contains---one can compose the
meaning of \iteg{king} with the \iteg{-ly} suffix to derive the adjective, while the
prefix \iteg{un-} supplies the negation.

In the field of natural language processing (NLP), a general approach is also to
decompose problems as the manipulation of elements smaller than whole sentences. We thus
face the task of determining what the elementary units of language are that
need to be represented. A simplistic solution is to take punctuation and spaces
as demarcating the boundaries of the elementary tokens we need to compute with,
and it is common to regard the resulting word tokens as opaque symbols without sub-structure.

One measure for the suitability of such tokenisation is how
readily an NLP system would encounter a token not observed in the corpus of text
it was trained on. That is, how much coverage does a corpus provide relative to
the hypothetical set of unique tokens forming the vocabulary of the language
under consideration?
Aside from phenomena such as slang, technical jargon and proper
nouns, coverage depends heavily on what semantic and grammatical information a given
language encodes into individual word forms.
Languages that are closer to the analytic end of the linguistic typology
spectrum tend to construct word forms from single or very few morphemes. At
the other extreme are synthetic languages, which combine many morphemes into a
single word form, either by chaining them together (agglutinative) or fusing
the information of multiple conceptual morphemes into a single affix.

English is a suitable instance of a language that is relatively analytic, as it
uses a fairly simple system of inflection and conjugation.
Grammatical relations are largely encoded by dedicated
function words (\eg \iteg{the} and \iteg{of}) and syntactic structure. 
Few grammatical distinctions are marked directly on
words---for example, English nouns are inflected only for their grammatical
number, with the plural typically indicated by an \iteg{-s} suffix.
In more analytic languages the assumption of opaque symbols is therefore quite
reasonable.  Although as the
derivational example of \iteg{unkingly} shows,
this assumption is an oversimplification even for morphologically `simple' English.

The same assumption is not well-suited to more synthetic languages. Czech, for
example, marks two number and seven case distinctions on nouns using a single
fused suffix, giving rise to ten different inflectional variants of the word
\iteg{král} \iten{king}, including 
\iteg{krále}, \iteg{králi}, \iteg{králem}, \iteg{králové},
etc.\footnote{From \url{http://en.wiktionary.org/wiki/král},
    accessed 15/04/2014.}
Or in the case of strongly agglutinative languages, it is possible to create new
word forms by the repeated addition of suffixes, theoretically without bound.
Turkish is a classic example of this:\footnote{\RaggedRight{From
    \url{http://en.wikipedia.org/wiki/Turkish_language}, accessed 
    15/04/2014, verified with the analyser of \citet{ColtekiTurkishMorphAnalyser}.}}
\begin{itemize}
  \item[] {%
\begin{tabular}{ll}
  \iteg{ev}             & \iten{house} \\
  \iteg{eviniz}         & \iten{your (pl.) house} \\
  \iteg{evinizdeyim}    & \iten{I am at your house} \\
  \iteg{evinizdeymişim} & \iten{I was apparently at your house}
\end{tabular}
}
\end{itemize}
Certain Germanic languages follow a similar regime in the formation of compound
nouns, as in the German example:
\begin{itemize}
  \item[] {%
\hspace*{-1em}\begin{tabular}{ll}
  \iteg{Regen}                             & \iten{rain}                  \\
  \iteg{Regenschirm}                       & \iten{umbrella}              \\
  \iteg{Regenschirmhersteller}             & \iten{umbrella manufacturer} \\
  \iteg{Regenschirmherstellergewerkschaft} & \iten{umbrella manufacturers trade
  union}
\end{tabular}
}
\end{itemize}
To assume that the words in these examples are atomic symbols ignores the very
apparent ways in which they share sub-structure.
A computational system that ignores sub-word structure is likely to obtain poor coverage
of the vocabulary of a language from a corpus, since morphological processes
readily create forms that are very rare or non-existent in a given corpus. 

A central theme in this dissertation is therefore to develop methods that account
more explicitly for the observed grammatical and semantic links between word
forms that share morphemes in their orthographic representations.
The notion of training language processing systems from example corpora
provides a bridge to the other defining element of the dissertation, which we turn to
next.

A computational system that processes language in a useful way is tasked with
more than merely representing the right symbols.  Approaches to the larger
question of casting the rules and idiosyncrasies of natural languages in
algorithmic form can be characterised based on two extremes.  At the one extreme
lie knowledge-intensive approaches that explicitly specify the symbols and
rules relevant to a language processing task, \eg morphological analysis or
translation, according to linguistic theories or hand-crafted heuristics.  This
can yield systems that work predictably and to a high standard, especially when
the application domain is very limited. But this approach has various downsides,
foremost of which is that it requires substantial additional labour to transfer
a solution to new domains or other languages. Moreover, a reliance on
dictionaries and prescriptive rules does not acknowledge the dynamic nature of
human language, where new words and expressions arise continually. 

At the other extreme are approaches that try to equip computer systems with a
learning ability, rather than supplying them directly with linguistic abilities.
The premise is that an appropriate combination of learning algorithm and data
can produce a level of language processing that is useful. In many instances,
we can assume that examples of language use (data) arise faster and more
organically than linguistic expertise or linguists, which speaks in favour of
approaches that rely on data. This notion has been a driving force in the
success and prominence of statistical approaches to machine translation and
speech recognition over the past two decades.

Model-based, data-driven approaches overcome some of the disadvantages of
knowledge-intensive methods. In principle, they scale more readily across
different languages and domains if adequate data is available.  This latter
condition is key. To clarify it further, the performance of NLP systems in this
general category is a function of both model assumptions and data. The
transferability, however, also makes it tempting to accept that a given set of
model
assumptions apply sufficiently well to multiple languages, which is not in
general the case.

A particular case where a one-size-fits-all scheme is problematic is that of
variations in morphological complexity.
Many statistical models of language adopt the token-based view that we argued is
ill-suited for morphologically rich languages. The fundamental problem is that,
at the level of word tokens, many legitimate word forms occur very rarely, or do
not occur at all, in a training corpus. This data sparsity makes
vocabulary coverage poor and hampers the robustness of parameter estimates, both
factors which constrain the performance of a model when processing new data.

This dissertation argues in favour of integrating high-level intuitions about
morphology into the assumptions of statistical language models in order to
address the aforementioned challenges posed by token-based processing of
morphologically rich languages.
The guiding principle is that morphologically related words should share statistical
strength---for example, separate observations of the words \iteg{king} and
\iteg{unlikely} should trigger the ability of a model to produce an informed
response to an unknown word like \iteg{unkingly}, and instances of \iteg{králi}
and \iteg{králové} should inform system behaviour towards \iteg{králem}.

\section{Problem statement}
The preceding section framed this dissertation in terms that are meant to be
understandable to a broader audience, which necessarily introduces some
imprecision.  In this section we now state and motivate the thesis in more
precise terms.

Our thesis is that basic linguistic intuitions about how words are formed
in morphologically rich languages can be exploited in the formulation of
probabilistic models that perform better at their tasks.
Probabilistic modelling offers a coherent mathematical setting in which to
express assumptions. It allows difficult tasks to be reformulated accurately in
terms of simpler tasks, \eg ``What is the probability of observing this
sentence?'' can change to the more approachable ``What is the probability of
observing each word in the sentence?''.  Similarly, probabilistic models can be
extended and combined in a principled way, limiting the number of dials that
have to be set manually in heuristic approaches.

We deal with language modelling tasks of two kinds. The first is the narrow
sense that \emph{a language model (LM) assigns probabilities to word sequences}.
This is the sense in which the term \emph{language model} is widely understood
in the literature, and which will henceforth be implied in this dissertation
unless otherwise stated.
A token-centric view in language models is at odds with morphologically rich
languages, where sub-word elements are correlated in meaningful ways across
different words.
By accounting for such correlations in the design of LMs, making use of
morphological information that we assume is given, we show in this
dissertation how the sparsity in the training data of morphologically rich
languages can be mitigated to improve their predictive performance.
Predictive performance is quantified primarily in two ways. The first is to
measure a LM's intrinsic ability to predict word sequences in held out test
data, which we do in terms of the perplexity metric.
The second is to consider what effect our morphology-targeting LM extensions
have when situated in a machine translation system. In this context, LMs play a
crucial role in biasing system output toward translations that are more fluent
renditions of the target language. LMs that encode patterns at the sub-word level
are by definition better suited for this role when translation is into 
morphologically rich languages. We thus measure translation quality in such
settings by comparing system output to reference translations, using the
automated quality metric \bleu.

The second type of modelling task we consider
is that of inducing meaningful sub-word structures from raw
text or word lists. Whereas the previously outlined task is primarily one of sequence-based
density estimation, this second sense of modelling natural language focuses on
the unsupervised learning of word structure, independent of actual word usage in
sentences.
A dominant approach in this task is to model words as sequences of morphemes,
with the goal of learning what those morphemes are and how to segment a given
word accordingly. This type of language modelling (in the broad sense) is of
practical interest in various other sub-problems in NLP, including
part-of-speech-tagging, syntactic parsing, or
indeed, for integrating sub-word level information into LMs, as we do in the
first part of this dissertation.
The idea that basic linguistic intuitions can inform and improve the design of
a model finds expression in morphology learning by observing that some
morphological processes involve discontiguous sub-word elements. We exploit this
to devise a model that goes beyond linear segmentation to capture elements of
non-concatenative morphology as well. As an example, in Arabic, the words \iteg{kitab}
and \iteg{kutub} share the same root \roottri{k}{t}{b}. We show that modelling such
discontiguous morphemes can improve the segmentational component of a model, and
that it can identify part of a lexicon of such morphemes given merely an
unannotated list of words.

\section{Dissertation outline}
The dissertation is structured as follows.

\chref{ch:bg} provides further background central to understanding the rest of
the material, and motivates the technical approaches followed.

Chapters~\ref{ch:hpylm} and \ref{ch:dist} document two different approaches to
integrating morphological information into probabilistic LMs. \chref{ch:hpylm}
introduces a hierarchical Bayesian model that accounts for the productive
compound word formation attested in certain Germanic languages. \chref{ch:dist}
introduces a method for integrating morphological information into probabilistic
LMs based on distributed feature representations.  The method is demonstrated to
be effective across a range of languages featuring different levels of
morphological complexity.

In \chref{ch:pymcfg}, the focus shifts to language modelling in the broader
sense, and we present an unsupervised approach to learning contiguous and
discontiguous morphemes from raw text. The approach is applied primarily to
Semitic languages, which make use of both kinds of morphemes.

Finally, \chref{ch:conclusion} draws together the work of the preceding three
chapters to summarise our main findings and present possible avenues for further
research.

The remainder of this chapter highlights the specific research contributions
presented in this dissertation.

\section{Main research contributions}
\begin{itemize}
  \item \textbf{New generative model of productive compounding}\\
We present a new language model where the formation of compound words in terms
of their theoretically unbounded number of components is part of the generative
process. This is done by extending the Hierarchical Pitman-Yor Language model
with an additional back-off level where compound heads are conditioned on the
sentential context, a decision based on grammatical considerations, while
compound modifiers are generated by an additional language model. The effect is
to reduce data sparsity due to compounding, and it results in higher precision
outputting of compounds from a translation system using our language model. This
work was originally presented at the 2012 EACL student workshop
\citep{BothaEACL} and published in the proceedings of COLING \citep{Botha2012}.

\item \textbf{Unsupervised learning of distributed morpheme representations}\\
We present a simple technique for learning distributed feature representations
for morphemes (given an external source of morphological segmentation) as part of a
distributed language model. We show that the resulting morpheme vectors can be
composed through addition to form word vectors that perform well in
replicating human judgements of the semantic similarity of pairs of words. The
benefit lies in being able to go beyond the vocabulary of the original source of
the word vectors.

\item \textbf{Distributed language modelling for rich morphology}\\
The model within which the aforementioned morpheme representations are obtained
is a novel variant of the log bilinear language model.
Its key property is to provide a soft tying of parameters for words that
share morphological content. We show that this is particularly beneficial for
modelling rare words.

\item \textbf{Integration of normalised distributed language model into a
    machine translation decoder}\\
We provide the first demonstration, to our knowledge, of integrating a
normalised distributed language model into a machine translation decoder.
This and the previous two contributions listed here form part of work accepted for
presentation and publication at ICML 2014 \citep{Botha2014}.

\item \textbf{Joint representation of contiguous and discontiguous morphemes}\\
We formulate a novel approach to the problem of representing concatenative and
non-concatenative morphology in a unified fashion. This is done by applying
mildly context-sensitive simple Range Concatenating Grammars (SRCGs), and we
illustrate the approach on Semitic morphology.
This and the next contribution listed here were originally reported in the
proceedings of EMNLP 2013 \citep{Botha2013}.

\item \textbf{Formulation of mildly-context sensitive adaptor grammars}\\ 
In order to do unsupervised learning within our SRCG-based approach to modelling
morphology, we formulate an extension of adaptor grammars to the SRCG-formalism.
This provides an additional method for inducing SRCG grammars and their
equivalents, which could be useful in applications beyond morphology.

\end{itemize}

%% file: ch-bg/ch-bg.tex
\chapter{\label{ch:bg}Background} 
\chapoverview{%
This chapter introduces background material that the rest of the dissertation
builds on.
We formulate the key aspects of sequence-based statistical language models (LMs)
and their evaluation. We present two alternative perspectives on language
modelling by introducing back-off \ngram models and distributed language models.
This sets up a short discussion of some existing methods for integrating
sub-word level information into those LMs, and what their limitations are.
The latter half of the section introduces aspects of Bayesian probability theory
and non-parametric modelling that are relevant in subsequent chapters.
Although most terminology and notation will be introduced when needed, we begin
this section with some preliminaries.
}

\section{Preliminaries}
\label{sec:bg:prelims}
This dissertation relies heavily on probability theory, especially as defined
for discrete events.
A discrete random variable $X$ can take on any value from an event space 
$\Omega=\{x_1,x_2,\dots\}$, which encodes the potential outcomes of a stochastic
operation or process. A simple example is that of tossing a coin, where
$\Omega=\{\text{heads},\text{tails}\}$. In general, $\Omega$ can be a countably
infinite set.

A probability mass function $P:\Omega \to [0,1]$ maps a particular value $x_i$ of
$X$ to the probability of that the event $X=x_i$ occurs, namely $P(X=x_i)$. We
will sometimes abbreviate this as $P(x_i)$, taking the random variable
assignment as implicit. This function characterises the distribution of the
random variable, and accordingly we usually refer simply to a ``probability
distribution''. The event space $\Omega$ is also referred to as the support of
the distribution.

\section{Statistical language modelling}
\label{sec:bg:SLM}
The canonical task in statistical language modelling is to assign a probability
to a sequence of
words, \eg \[P(\tpr{the~sky~is~grey}) =\; ?\]
Such probabilities are inferred by using statistical estimates from data, in
other words, the model is \emph{trained} on a corpus of text in the language of
interest.

LMs play a crucial role in applications like speech recognition, machine
translation and predictive text input, where an intended utterance must be
decoded from an ambiguous signal. LMs that better capture the regularities and
idiosyncrasies of natural language are better at discriminating among
alternative output candidates that arise in these applications. They are not
directly concerned with the input signal (audio features, a foreign language, or
a sequence of keystrokes), but instead reward output that is a more fluent and
coherent rendition of the target language.

The fundamental challenge in assigning probabilities to word sequences is to
contend with the fact that the set of possible word sequences is infinite, while
any given data set used for estimation is finite.  The thesis investigated here
is premised on the observation that morphological processes have a strong effect
on the sparsity of training data. Broadly speaking, more complex morphological
processes exacerbate data sparsity. The overall approach of modelling sub-word
elements is therefore intended to help overcome data sparsity. The motivation is
that this is likely a more viable long term solution than estimating
morphologically blind models from ever more data, effective as that may be for
particular languages in the short term \cite{Brants2007}.

The dominant strategy\footnote{Other strategies also exist, \eg the
  whole-sentence model of \citet{Rosenfeld2001}.} in LM estimation is to
decompose the probability \p{\mb{w}} of a word sequence
$\mb{w}=w_1 \dots w_L$, using the chain-rule, as
\begin{equation}
\p{\mb{w}} = \prod_{i=1}^L \p{w_i \mid w_1, \dots, w_{i-1}}.
\end{equation}
The task of estimating probabilities for entire word sequences of arbitrary
length is thereby simplified into one of estimating probabilities for individual
words, conditioned on all the preceding words in the sequence. An approximation
that further simplifies this is based on the assumption that a particular word
only depends directly on the ones immediately preceding it. A word sequence can
accordingly be modelled as a Markov chain of order $n-1$:
\begin{equation}\label{eq:bg:ngram_prob}
\p{\mb{w}} \approx \prod_{i=1}^L \p{w_i \mid w_{i-n+1}, \dots, w_{i-1}}.
\end{equation}

The sequence of $n$ words $(w_{i-n+1}, \dots, w_i)$ is commonly referred to as
an \mbox{\emph{$n$-gram}}, with \emph{history} $h=(w_{i-n+1}, \dots, w_{i-1})$. We use
the short-hand notation for a word sequence, $w_i^j\equiv(w_i,\dots,w_j)$.
Alternative terms that we use are that $h$ is the \emph{context} from which the
\emph{target word} $w_i$ must be predicted.  The support of the conditional
distribution $\p{w_i \mid \tokseq{w}{i-n+1}{i-1}}$ is the vocabulary \V, the
discrete set of word types to be modelled.

Sequence boundaries require some special handling in Markov chains. Common
practice is to assume that the conditioning context 
of the first token $w_1$ in a sequence is provided by a special padding symbol,
while the use of a designated symbol marking the end of the sequence allows for
coherent modelling of variable-length sequences.

We often need to compute the probability of a set of word sequences, such as
sentences in a corpus. If $\{\mb{w}_1,\dots,\mb{w}_N\}$ are $N$ 
independent word sequences, then the joint probability (using the aforementioned
Markov assumption) is
\[\p{\mb{w}_1,\dots,\mb{w}_N} 
  = \prod\limits_{j=1}^N \p{\mb{w}_j} 
  = \prod\limits_{j=1}^N \prod\limits_{i=1}^{|\mb{w}_i|} \p{w_{j,i} \mid
    \tokseq{w}{j,i-n+1}{j,i-1}},
\]
In subsequent formulations we refer to a corpus simply as one sequence of
tokens, \tokseq{w}{1}{M}, taking these mechanics of boundary symbols and
sentence independence as being implied.

The reliance on the conditional probability distributions
$\p{\tokseq{w}{i-n+1}{i-1}}$ makes parameter estimation more tractable and is
advantageous for the dynamic programming employed by machine translation (MT)
decoders. We consider
language model estimation in \S\ref{sec:bg:two-main-approaches}.
For now, we assume an estimated model $P_{LM}$ is given and
consider how its quality can be evaluated.

\subsection{Language model evaluation}
\label{subsec:bg:lm_eval}
A common evaluation method is to measure how well the language model $P_{LM}$
predicts previously unseen test data $w_1^N$.
\emph{Perplexity} is a useful metric for this purpose, as it is well-defined for
any length $N$ and allows for a fair comparison of models in spite of
differences in their assumptions or internal workings. The
perplexity of a model $P_{LM}$ on the test
sequence \tokseq{w}{1}{N} is
\begin{align}
\textrm{ppl} &= 2^{H(P_{LM})}, 
\label{eq:bg:perplexity} \\
\intertext{where $H$ is the cross-entropy}
 H(P_{LM}) &= \sum_{i=1}^N -\frac{1}{N} \log_2 \p{w_i \mid \tokseq{w}{1}{i-1}}.
\label{eq:bg:xent}
\end{align}
The intuitive interpretation of a perplexity value of $k$ is that the
uncertainty of the model with respect to the test data is equivalent to the
uncertainty in the outcome of throwing a $k$-sided fair die. A model is thus on
average $k$-ways perplexed by the observation of each test word. Lower
perplexity corresponds to lower uncertainty, which is the mark of a better
model.\footnote{For a basic mathematical derivation of perplexity in terms of
  entropy, see \citep[][ch.~7]{Koehn2010}.} Note that perplexity, as defined
above, is independent of model assumptions such as the Markov assumption applied
in the formulation of \ngram models.

A comparison of two models $P_{LM_1}$ and $P_{LM_2}$ in terms of their perplexity
on a given data set is only fair if the support of the probability
distributions is equivalent, in other words, if they model the same vocabulary
\V. This matter requires special attention when going beyond the standard
LM assumption that \V is a finite set, as we do in \chref{ch:hpylm}.

\subsection{Two dominant approaches to Markov-based LMs}
\label{sec:bg:two-main-approaches}
The focus now turns to two dominant  approaches to modelling the categorical
conditional distributions \p{w_i \mid \tokseq{w}{1}{i-1}} used in Markovian language
models.

1) One approach is to infer these distributions by using exact-matching of
\ngrams, thereby directly representing the high-dimensional discrete space of
\ngrams.  For historical reasons, this class of models are referred to simply as
\emph{\ngram language models}.

2) The other major approach is to embed words in a low-dimensional, continuous
vector space and learn a function that uses the arising word vectors to map an
\ngram history into a probability for the target word $w_i$. These
\emph{distributed language models} are also known by the terms neural, connectionist or
continuous-space language models. 

We instantiate the thesis that basic linguistic intuitions about word formation
are beneficial in modelling morphologically rich languages within each of these
two approaches. The next two sections present them in more detail.

\subsubsection{\ngram language models}
A discrete \ngram LM is in fact a collection of conditional multinomial
distributions like \p{w_i \mid \tokseq{w}{i-n+1}{i-1}}, estimated directly from the
empirical distribution of \ngrams in training data.  Maximum likelihood
estimates suffer a variety of problems: Some \ngrams are not observed in the
training data yet a model must still be able to score them. \ngrams that do
appear in the data may do so with a low frequency that is not representative of
their true distribution in the language \cite{Good1953}.
To address these and other more subtle challenges arising from data sparsity, a
wide variety of techniques have been devised to smooth the empirical \ngram
distributions by redistributing probability mass \cite{Chen1998}. 

The interpolated modified Kneser-Ney (MKN) model
\cite{KneserNey1995,ChenGoodman1999CSL} is considered the state-of-the art
discrete \ngram LM \cite{ChelbaOneBnLMBenchmark2013}. It serves as a baseline model in the empirical evaluations
reported later in this dissertation.  We briefly introduce key smoothing
concepts it relies on, some of which have analogues in the models we
subsequently develop.

The first concept is that of basing probability estimates of a certain random event on
more general random events.
Following on from the initial example in this chapter, 
we can regard the observation of the word \emph{grey} in the context \emph{sky~is}
as a more specific event than observing it in the shorter context \emph{is}.
That is, the bigram \emph{is grey} is a generalisation of the trigram 
\emph{sky~is~grey}. Thus to overcome the data sparsity inherent in estimating \ngram LM
parameters, an effective technique is to let
\psub{LM}{\tpr{grey} \mid \tpr{sky}, \tpr{is}} be a
function of \p{\tpr{grey} \mid \tpr{is}}.

More formally, defining \psub{LM}{w_i \mid \tokseq{w}{i-n+1}{i-1}} as a function of
lower-order conditional distributions \psub{LM}{w_i \mid \tokseq{w}{i-n+2}{i-1}}
can be done through interpolation \cite{JelinekMercer1980} or by strictly
switching to the lower-order model \cite{Katz1987}.
Interpolation typically works better \cite{Chen1998}, so we restrict the
attention to its functional form:
\begin{equation} \label{eq:bg:interpolated-backoff}
\psub{LM}{w_i \mid \tokseq{w}{i-n+1}{i-1}} = %
\alpha\!\left(\tokseq{w}{i-n+1}{i}\right) + \gamma\!\left(\tokseq{w}{i-n+1}{i-1}\right)
\psub{LM}{w_i \mid \tokseq{w}{i-n+2}{i-1}},
\end{equation}
where $\alpha\!\left(\tokseq{w}{i-n+1}{i}\right)$ denotes a smoothed version of
the \ngram conditional distribution (see below), while 
the $\gamma$ parameters are set to normalise the overall probability
distribution, $\psub{LM}{w_i \mid \tokseq{w}{i-n+1}{i-1}}$. This procedure can be
followed recursively and bottom out with a reliance on unigram model
$\psub{LM}{w_i}$.

The matter of defining $\alpha(\cdot)$ provides the connection to the second
important smoothing concept of \emph{discounting}. 
Instead of
relying directly on the actual number of times an \ngram is observed in the
training data, $C(\tokseq{w}{i}{j})$, estimation is done using
counts that are adjusted down, to give a discounted frequency
\mbox{$C_{disc}(\tokseq{w}{i}{j}) \leq C(\tokseq{w}{i}{j})$}. 
MKN uses absolute discounting \cite{NeyEssenKneser1994} to adjust
an \ngram count 
by a constant $D_m$,
\begin{equation}\label{eq:bg:MKN-discount}
C_{MKN}(\tokseq{w}{i-n+1}{i}) = \max \left(C(\tokseq{w}{i-n+1}{i}) - D_m,0\right),
\end{equation}
where $m \in \{1,2,3+\}$ designates one of three discount levels according to
whether the \ngram occurs once, twice, or more than twice.

MKN combines discounting and (interpolated) back-off in a particular way.
$\alpha(\cdot)$ is the discounted maximum likelihood estimate 
\begin{equation}\label{eq:bg:MKN-discount-and-interp}
\alpha\!\left(\tokseq{w}{i-n+1}{i}\right)
= \frac{C_{MKN}(\tokseq{w}{i-n+1}{i})}{C(\tokseq{w}{i-n+1}{i-1})}
\end{equation}
The remaining element for using \autoref{eq:bg:interpolated-backoff} is to specify the
 back-off distribution \psub{LM}{w_i \mid \tokseq{w}{i-n+2}{i-1}}.
 Kneser-Ney smoothing uses purpose-built lower-order
distributions that exploit the intuition that the diversity of contexts a word
appears in often represents a more accurate view of its true probability than the
\ngram count itself. That is, the empirical prevalence of the \ngram \emph{Los
  Angeles} overstates the true probability of the word \emph{Angeles}, so Kneser-Ney
smoothing bases the estimate instead on the number of unique \ngram histories
the word appears in.

\ngram language models, and MKN smoothing in particular, have been relied on
heavily in the domain of machine translation in the last~15 years, and have more
recently received significant amounts of engineering attention in order to scale
to vast training data sizes \cite{Heafield2011,Heafield2013}.  A main limitation
of MKN is that it is fundamentally heuristic-based, albeit heuristics
derived from rigorous empirical work.  This makes it challenging to incorporate
new smoothing ideas. \citet{Teh2006} proposed a Bayesian approach to \ngram
language modelling which implements some of the same principles that MKN depends
on, but offers a framework that better lends itself to modifications of the
basic smoothing behaviour. In \chref{ch:hpylm} we use that framework to create a
more sophisticated smoothing method which exploits simple linguistic intuitions
about word formation.

\subsubsection{Distributed language models}
\label{sec:bg:distributed}
A salient feature of the \ngram models in the previous section is that they 
parametrise a high-dimensional discrete space directly, based on the identities of
words. This amounts to a harsh binary perspective on the way words are
correlated in text by requiring that a test \ngram (or its backed-off version)
exactly match an event in the training data. 

An alternative approach that acknowledges a more multifaceted notion of
correlation among word occurrence is to represent words in a multidimensional
vector space. Thus instead of each word type $v$ in the vocabulary \V being
merely a
symbol in a discrete set, each $v$ has an associated real-valued
\emph{distributed feature vector} $\mb{r}_v \in \mathbb{R}^d$. This immediately
enables reasoning about the relations among words in terms of how they relate in
vector space, with the intuition that ``similar'' words are closer to each
other, using a metric such as the angle or Euclidean distance between their
representation vectors. As an example, the words of the vocabulary fragment
$\V=\{\text{blue},\text{clouds},\text{grey},\text{skies},\text{sky},\dots\}$ could
have feature representations that look as follows when plotted in two-dimensional space:
\begin{center}%
\input{img/word-vec}

\end{center}%

In general, the different dimensions have no intrinsic meaning, although this
may depend on the method used for obtaining distributed representations.  For
example, in the area of distributional semantics, vector representations of
words have been constructed directly from corpus-based co-occurrence statistics,
using the most frequent words as basis vectors \citep[][\emph{inter
  alia}]{BullinariaLevy2007,MitchellLapata2008}.  However, we consider models
where the word vectors are learnt as part of LM training, and the different
dimensions simply capture different correlates.

Distributed language models (DLMs) apply distributed feature representations in
assigning probabilities to word sequences. The general idea can be stated as
follows in terms of \ngram modelling: An \ngram history \tokseq{w}{i-n+1}{i-1}
is mapped to a vector $\mb{x} \in \mathbb{R}^{d'}$ as some function $f(\cdot)$ of
its word vectors, \ie $\mb{x} = f(\mb{r}_{w_{i-n+1}},\dots,\mb{r}_{w_{i-1}})$,
where $d'$ is free to differ from $d$.
A further function $g(\cdot)$ then transforms $\mb{x}$ into a conditional
probability: $\psub{LM}{w_i \mid \tokseq{w}{i-n+1}{i-1}} = g(w_i, \mb{x})$. 
A variety of neural network architectures and related methods have been used to model the functions
$f$ and $g$ while jointly learning the representation vectors
\cite{XuRudnicky2000,Bengio2003,Morin2005,Mnih2007,Mikolov2010}.

The formulation above is a simplification of the \emph{neural probabilistic language model}
\cite{Bengio2003}, but is sufficient to highlight the key advantages of DLMs
over the discrete \ngram LMs of the previous section. The main benefit is that,
by making the conditional probabilities smooth functions of the \ngram context,
generalisation to new \ngrams is more automatic. It becomes possible for two
lexically distinct phrases such as \emph{the grey clouds} and \emph{the blue
  sky} to effect similar conditioning on an immediately following word, assuming
the word vectors involved are pair-wise ``nearby'' in the distributed
representation space as demonstrated in the preceding diagram.  This obviates the
need for explicit back-off strategies.
In more general terms, DLMs mitigate the sparsity problem associated with the
high-dimensional parameter space of discrete \ngram LMs (which is in principle
$|V|^n$) by operating in a lower dimensional continuous space. In particular, $d
\ll |\V|$, and values of 50~to~1000 are typical for $d$, whereas typically
$|\V|>10^5$, or much larger for morphologically rich languages.

This dissertation addresses two issues in DLMs. The first is to
consider how they can be improved for morphologically rich languages by
using the internal structure of words, and the second is to demonstrate the
 full integration of DLMs into a machine translation system (\autoref{ch:dist}).

\subsection{Language modelling at the sub-word level}
\label{sec:bg:subwordLMs}
Both approaches to language modelling presented above usually consider words as
the fundamental modelling unit. They seek to mitigate data sparsity in different
ways, but do not directly address a major source of that data sparsity, which is
that of complex morphology. Morphological complexity comes in many guises, but
the common outcome is that languages with more complicated rules for forming
words tend to have larger vocabularies of surface word forms.
This exacerbates the data sparsity to which \ngram LMs are prone to by
definition, while DLMs are also stretched by having fewer training instances of
individual word forms from which to learn suitable representation vectors and
the accompanying probability function.

A further implication of rich morphology is that test data can be expected to
contain relatively many word forms not appearing in the training
vocabulary.\footnote{Mismatches in the domains of text are another major source
  of out-of-vocabulary words in test data.} The standard practice for dealing
with this out-of-vocabulary (OOV) problem is to set the model up in a way that
reserves some probability mass for a generic ``unknown-word'' symbol, which we
will denote as \unk. This has the shortcoming of treating truly novel words,
such as proper nouns, the same as previously unseen morphological variants of a
known word, ignoring the obvious relation among them.

Both these problems can be addressed by shifting the fundamental unit of
representation to be at the sub-word level. The sought-after effects is that
morphologically related word forms should share statistical strength in some way
and that some OOVs can be handled in terms of their sub-word structure.

Previous approaches to integrating morphological information into language models have been
proposed.\footnote{The focus here is on the types of language models introduced
  in \S\ref{sec:bg:two-main-approaches}, but we note that maximum-entropy
  language models also offer a useful way of integrating morphological 
information, \eg \cite{Minkov2007}.}

A naïve strategy is to segment words into their morphemes in a preprocessing
step, and then construct otherwise unmodified language models from the modified
tokenisation.  As an example of decomposition, the sentence \\
\indent \emph{The\quad forecasters\quad  predicted\quad  more\quad  flooding} \\
may be segmented into morphemes to give \\
\indent \emph{The\quad  forecaster\quad  s\quad  predict\quad  ed\quad  more\quad  flood\quad  ing} \\
This technique reduces the number of unique token types, which directly
addresses sparsity, but introduces its own problems: It begs the question of how
aggressively to segment---this often opens the door to ad-hoc heuristics that do
not transfer reliably across multiple tasks or languages. And it disrupts the
contextual conditioning in back-off \ngram models by making a hard switch away
from modelling sentential coherence, say via \p{\tpr{predicted} \mid \tpr{the},
  \tpr{forecasters}}, toward modelling fragments, say \p{\tpr{ed} \mid \tpr{s},
  \tpr{predict}}. This strategy has been used in speech-recognition
\citep{Geutner1995,Ircing2001a,Hirsimaki2006}, and machine translation
\cite{Virpioja2007}.

A more systematic approach is that of \emph{factored language models} (FLMs)
\citep{Bilmes2003FLM,FLMTutorial2008}. 
FLMs cast each word as a bundle of $K$ features
$w_i=(f_i^{(1)},\dots,f_i^{(K)})$ over which a variety of conditional
probability distributions are defined. To clarify the notation, a standard
\ngram model trivially uses word form as its single feature ($K=1$) and the
\ngram conditional distribution
is written as $\p{f_i^{(1)} \mid f_{i-n+1}^{(1)},\dots,f_{i-1}^{(1)}}$.
The intention is to allow for the integration of information beyond word
forms.
For example, a word's \posphrase and stem could be used as additional features,
$w_i=(f^{\text{surface}_i}, f^{\text{PoS}_i}, f^{\text{stem}_i})$,
as in $\textvector{cloudy}=(cloudy, noun, cloud)$ ($K=3$).

This factored representation bears superficial resemblance to the distributed
feature representations introduced in \S\ref{sec:bg:distributed}. The main
difference is that the FLM features are discrete random variables over which
arbitrary directed graphical models are defined. Furthermore, feature values are
likely set with reference to some external source and not trained as part of the
FLM.

A FLM that goes beyond standard \ngram
conditioning might incorporate a sub-model $\p{f_i^{\text{surface}} \mid
  f_{i-1}^{\text{stem}} }$, which conditions the target word 
on the stem of the preceding word.
Multiple conditional distributions of this kind can be combined, similar to the
way standard \ngram models rely on lower-order conditionals. But with the richer
information encoded, an important question becomes how to prioritise the
back-off sequence.
\citet{Bilmes2003FLM} proposed solving this by generalising back-off as a procedure that happens on a
graph (instead of the linear procedure of dropping context words as in back-off
\ngram models).
They rely on genetic algorithms to acquire the structure of the back-off graph
over the $K$ feature dimensions and $n$ sequence positions in an \ngram approach,
while the individual conditional distributions use standard smoothing or
discount algorithms.

The factored language model approach has also been applied to neural
probabilistic language models \cite{Alexandrescu2006}, which, as pointed out
before, remove the necessity to deal with back-off paths. Associations among
different features are learnt as part of the network weights rather than having
to be specified explicitly as present or absent.
These \emph{factored neural language models} (FNMLs) in effect associate a
distributed representation with each feature component of a word, so the overall
word representation $\mb{r}_w=[\mb{f}^{(1)};\dots;\mb{f}^{(K)}]$ is constructed
by concatenating the feature vectors $\mb{f}^{(k)} \in \mathbb{R}^{d(k)}$, where
$1 \leq k \leq K$ and the dimensionality $d(k)$ may vary across components.
 
FLMs are a suitable vehicle for integrating external information into language models, including
morphology.\footnote{As an example of FLMs using non-morphological information, \citet{Adel2013a}
applied them successfully to the problem of code-switching by incorporating a language-ID factor for
each word.} They can lower perplexity \cite{Bilmes2003FLM,Alexandrescu2006,Wu2012} 
and improve speech recognition accuracy \cite{Vergyri2004}.

The main shortcoming of FLMs is that the number of factors $K$ is a fixed constant for all
words. Therefore they cannot naturally encode variable-length information, such as the surface-level
morphemes of a word. In largely fusional languages, such as Czech, there is
likely no loss in assuming a fixed-length segmentation, say prefix + stem +
suffix.\footnote{Another option is to apply FLMs to morpheme-segmented data, so that
each morpheme is assigned a set of factors \citep{Mousa2011}.}
But a better solution is needed for agglutinative languages or phenomena where
the number of morphemes per word can vary substantially across the distinct
surface words of a given language.

In Chapters~\ref{ch:hpylm}--\ref{ch:dist}, we present two separate language modelling
approaches that might be regarded as variations on the idea of FLMs but that
have inherent support for variable-length decompositions of words.

\subsection{Summary}
The preceding section set up some terminology and definitions used in statistical
language modelling. We introduced the two widely used approaches of back-off
based \ngram LMs and vector-based DLMs. Finally, we discussed an existing
general approach to integrating morphological information into both types of
language models, and pointed out that it does not naturally support strong
agglutination.

\section{Non-parametric Bayesian modelling }
\label{sec:bg:npbayes}
This section introduces the key elements of the mathematical framework employed
in Chapters~\ref{ch:hpylm}~and~\ref{ch:pymcfg}.

Bayesian modelling centres around two concepts that address the questions of
inferring the parameters $\theta$ of a model given observed data $D$, and making
predictions about a new data point $\tilde{x}$ under the model.
First, instead of working with point-estimates of $\theta$, the Bayesian
approach quantifies the uncertainty about the value of $\theta$ by means of
probability distributions over its potential values. 
The \emph{prior distribution} $P(\theta \mid \alpha)$ expresses our beliefs about
what the value of $\theta$ might be in absence of observed training data. The
hyperparameters $\alpha$ control the behaviour of the prior distribution.

The objective, however, is to make inferences in light of observed data. The
\emph{posterior distribution} $P(\theta \mid D, \alpha)$ quantifies the
uncertainty in the values of $\theta$ following the observation of training data
$D$. 
The prior and posterior distributions are related by Bayes' law as
\begin{equation}\label{eq:bg:posterior}
P(\theta \mid D, \alpha) \propto P(D \mid \theta) \, P(\theta \mid \alpha),
\end{equation}
where the term $P(D \mid \theta)$ is referred to as the \emph{likelihood
  distribution} of the data $D$ conditioned on the model parameters $\theta$.
This factorisation is often useful in problem settings where 
the terms on the right-hand side of \eqref{eq:bg:posterior} are computed more
easily than directly computing the posterior probabilities on the left-hand
side.

Our focus falls instead on the modelling convenience afforded by the
specification of a prior distribution. Judicious choice of prior distribution
provides a principled way to encode external knowledge or intuitions relevant to
a problem into a probabilistic model. The effect of the prior on the posterior
distribution is strongest when little or no data has been observed. The latter
two properties of Bayesian modelling makes it especially suitable for our
aims of overcoming the data sparsity problem in morphologically rich languages
by encoding sub-word information into models, which we can do through
constructing appropriate Bayesian priors.

The second central concept of the Bayesian approach is posterior inference.
The prediction of a new data point $\tilde{x}$ is approached by 
relying on all possible values of the parameters $\theta$ rather than a point-estimate.
That is, the \emph{posterior predictive distribution} $P(\tilde{x} \mid D,\alpha)$ is determined by 
marginalising the posterior distribution over $\theta$:
\begin{align}
P(\tilde{x} \mid D, \alpha) &= 
\int P(\tilde{x} \mid \theta) P(\theta \mid D, \alpha)\,\mathrm{d}\theta.
\label{eq:bg:posteriorpredictive}
\end{align}
For many types of models, this integral cannot be solved analytically nor
computed tractably. The two major solutions to this impasse are to approximate
the posterior distribution with a simpler function which enables exact numerical
optimisation \citep{Beal2003}, or to use Markov chain Monte Carlo (MCMC) techniques to approximate
the integral with a finite number of samples drawn from the true distribution
\citep{Neal1993,Gilks1996}. Our work follows the latter option.

\subsection{Pitman-Yor process prior}
\label{subsec:bg:PYP}
A central component of the Bayesian modelling work in this dissertation is the
use of the Pitman-Yor process \citep[PYP;][]{Pitman1995,Pitman1997,Ishwaran2001}
to construct prior distributions that encode sub-word structure.
This follows widespread and successful application of the PYP in NLP tasks
including unsupervised learning of grammars
\citep{CohnBlunsomGoldwaterTSGs,Levenberg2012,CohenBleiSmith2010}, word segmentation
\citep{Mochihashi2009a,Neubig2010}, morphology 
\citep{Goldwater2006NIPS}, parts of speech \citep{BlunsomCohn2011POS} and
\ngram language modelling \citep{Teh2006,Wood2009}.

The suitability of the PYP in these diverse applications, and indeed in this
dissertation, arises from two properties. The first is that it falls within the
class of non-parametric statistical models \citep{Wasserman07AllNPStats}, which
means that the number of model parameters (or more generally, the model complexity) is
driven by the observed data instead of being fixed \emph{a~priori}.
Non-parametric modelling matches the boundlessness of many linguistic objects
very well---words, morphemes and syntactic structures all seem to constitute
countably infinite sets. 
The second key property of the PYP is that it naturally produces power-law
distributions that model the distribution of word frequencies
\cite{Zipf1932} and that are useful for other linguistic elements
\citep{Goldwater2011}.

The PYP is a stochastic process that generates
distributions over exchangeable random partitions of a countably infinite set of elements.
A PYP is specified by a \emph{base distribution} $P_0$, and two
hyperparameters: the \emph{discount} $a$ and \emph{strength} $b$, with
$0 \leq a < 1$ and $b>-a$. We write it as $\pyp(a, b, P_0)$.
Setting $a=0$ reduces the PYP to the more familiar Dirichlet process
\cite{Ferguson1973} with concentration parameter $\alpha=b$.  The role of the base
distribution is considered in the next section, but we note that it determines
the support of a draw $G$ from the PYP, where $G$ is itself an infinite
dimensional probability distribution.

The PYP has various characterisations. An infinite dimensional distribution $G$
sampled from a PYP can be constructed by a stick-breaking procedure
\cite{Sethuraman1994,Ishwaran2001}.
However, in our applications it is more useful to work with draws from $G$.
The Pitman-Yor Chinese Restaurant Process \citep[PYCRP;][]{Ishwaran2003}
provides a means of obtaining such a draw $\mb{y} \sim G$ and circumvents the problem of representing the infinite dimensional object.

The PYCRP can be defined through the metaphor of an unbounded sequence of
customers entering a restaurant one-by-one and being seated at tables with
unbounded seating capacity and of which there is a limitless supply.
The table allocation of customer $x_i$ is recorded as table index $z_i \in\mathbb{Z}^+$.
Let $\mb{z}_n$ denote the sequence of table indices after $n$ customers have
been seated, and let $n_k$ denote the number of customers seated at the $k$\th of
the $m$ non-empty tables in the restaurant. Hence $n=\sum_{k=1}^m n_k$.
The first customer $x_1$ sits at the first table by definition, \ie $z_1=1$.
Given $n\geq1$ previously seated customers, the next customer $x_{n+1}$ is
seated at table index $z_{n+1}\in [1,m+1]$, sampled from the probability
distribution
\begin{equation} \label{eq:bg:seatconditional}
  \p{z_{n+1} = k \mid \mb{z}_n, a, b} =
\begin{cases}
  \dfrac{n_k - a}{n+b} & \text{ if } 1 \leq k \leq m       \\[3ex]
  \dfrac{am + b}{n+b} &  \text{ if } k=m+1.
\end{cases}
\end{equation}

The dependence on $n_k$ in the first case means customers are more likely to be
assigned to tables that already have relatively more customers. However,
assignment to a previously unoccupied table (case~$k=m+1$) always carries
non-zero probability, and is itself dependent on the number of occupied tables
$m$.  The interaction of this preferential attachment dynamic and the positive
attitude towards novelty induces a power-law distribution over the number of
customers $n_k$ at each table, provided $0<a<1$ \cite{Goldwater2011}.

The joint probability of a particular seating assignment $\mb{z}_n$ under the
PYCRP is obtained as the product of the individual seating decisions:
\begin{align}
  &\p{\mb{z} \mid a, b} %
  = \p{z_1 \mid a,b} \cdot \prod\limits_{i=2}^n \p{z_i \mid \mb{z}_{i-1}, a, b}  \\
  \label{eq:bg:p-z-joint}
  &= 1 \cdot \frac{(a(1)+b)\dots(a(m-1)+b)\times
    \prod\limits_{k=1}^m \big( (1-a)\dots(n_k-1-a) \big) } {(1+b)\dots(n-1+b)} \\
  \label{eq:bg:p-z-joint-expand} 
  &= \frac{\Gamma(1+b)}{\Gamma(1+b)}
        \cdot \frac{\prod\limits_{k=1}^{m-1} (ak+b)}{(1+b)\dots(n-1+b)} \notag \\
  & \qquad \qquad \qquad \qquad
          \times \prod\limits_{k=1}^m \left( (1-a)\dots(n_k-1-a)
            \frac{\Gamma(1-a)}{\Gamma(1-a)} \right)  \\
  &= \frac{ \Gamma(1+b)}{ \Gamma(n+b)}
  \left( \prod\limits_{k=1}^{m-1} (ak+b) \right)
  \left( \prod\limits_{k=1}^m \frac{ \Gamma(n_k-a) }{ \Gamma(1-a)} \right).
\end{align}
The numerator in \eqref{eq:bg:p-z-joint} accounts separately for the initial occupation
of the tables and for the assignment of subsequent customers to each table. 
The final form is efficient for computation, making use of the Gamma function which is defined as
$\Gamma(x)=\int_0^\infty u^{x-1} e^{-u}\,\mathrm{d}u$ for $x>0$, with the
identity $\Gamma(x) = (x-1) \Gamma(x-1)$ for real-valued $x>0$.

\subsection{Pitman-Yor process base distributions}
A sequence of table assignments $\mb{z}_n$ under the PYCRP is fundamentally just
a partition of a sequence of integers, $\{1,\dots,n\}$, into $m$ clusters such
that they exhibit certain frequency distributions. For concrete modelling
applications, the usefulness of this depends on what meaning is assigned to the
customers and their tables. The role of the base distribution $P_0$ of the PYP
is to provide that meaning. In the restaurant metaphor, the first customer $x_i$
to be seated at table with index $k$ triggers the sampling of a dish $\ell_k$ from
the base distribution, $\ell_k \sim P_0$. All subsequent customers to join a
non-empty table~$k$ share in the same dish $\ell_k$ previously chosen for that
table.

If we define the identity $y_i$ of a customer $x_i$ at table $z_i$ to be the
table label $\ell_{z_i}$, the effect is to produce a sequence $\{y_i\}$ distributed
as follows:
\begin{align*}
  y_i \mid G       &\sim G \\
  G \mid a, b, P_0 &\sim \pyp(a, b, P_0) 
\end{align*}

The conditional probability of the identity $y_{n+1}$ for an additional customer
$x_{n+1}$, given labels $\bm{\ell}$ for the $n$ customers seated at $m$
tables according to the seating arrangement $\mb{z}_n$, is
\begin{align}
  P(y_{n+1} = y \mid \mb{z}_n, \bm{\ell}, a, b)
 &= \sum\limits_{k=1}^m 
      \delta_y(\ell_k) \frac{n_k - a}{n+b} 
      + \frac{am + b}{n+b} P_0(y) \notag \\
      &= \frac{N_y - a m_y   + (a m + b)P_0(y)}{n+b},
      \label{eq:bg:predictive-conditional}
\end{align}
where $\delta$ is the Kronecker delta indicator function, $N_y$ is the number of
customers seated at tables labelled $l_k=y$, and $m_y$ is the number of such tables.

As an example,\footnote{This example derives from the connections between
  language model smoothing and the PYP pointed out by
  \citet{Goldwater2006NIPS,Teh2006}.}
the PYP can be used to define a simple unigram language model
over a closed vocabulary \V. 
Let the base distribution $P_0=\mathrm{Uniform}(|\V|)$ and regard a customer $x$ as a
generic word token. A label $\ell$ drawn from the base distribution and assigned
to the token via the table it joins in the associated restaurant then supplies the token its identity as a
particular word type from the vocabulary \V. Note that the same label $\ell$ 
can be drawn repeatedly from $P_0$.
The term $N_y-am_y$ in the numerator of
Equation~\eqref{eq:bg:predictive-conditional} can be
understood to discount the frequency $N_y$ with which a word $y$ occurs in a corpus by
an amount $a m_y$, while the second term provides interpolation with the uniform
base distribution, in analogy to the language model smoothing described by
Equation~\eqref{eq:bg:interpolated-backoff}, p\pageref{eq:bg:interpolated-backoff}.
This smoothing ability forms the basis of the language model
\cite{Teh2006} that we extend in \chref{ch:hpylm}.

The base distribution can, however, also be used to generate more complex labels
than word identities. In particular, they can be used to generate trees of
various kinds \cite{CohnBlunsomGoldwaterTSGs,Johnson2007a,Levenberg2012},
 a line of work we extend to another grammar formalism in \chref{ch:pymcfg}.

\section{Summary}
This chapter described the two approaches to sequence-based statistical language
modelling that subsequent chapters extend by integrating information about
sub-word structures.
In the second part of the chapter, we introduced the aspects of Bayesian
modelling with the non-parametric Pitman-Yor process priors that we employ in
the next chapter and the penultimate chapter of this dissertation.

%% file: ch-bg/img/word-vec.tex
\begin{tikzpicture}[scale=1, every node/.style={font=\footnotesize}]
    \draw [<->,thick] (0,2) node (yaxis) [above] {$x_2$}
        |- (3,0) node (xaxis) [right] {$x_1$};
   \node (origin) at (0,0) {};

   \fill (0.3,1) circle (2pt) node[right](p1) {grey};
   \draw [-latex] (origin.center) -- ($(0.3,1) - (0,0.1)$);

   \fill (0.15,1.3) circle (2pt) node[right](p1) {blue};
   \draw [-latex] (origin.center) -- ($(0.15,1.3) - (0,0.1)$);

   \fill (1.5,1.2) circle (2pt) node[right](p1) {sky};
   \draw [-latex] (origin.center) -- ($ (1.5,1.2)- (0.1,0.05)$);

   \fill (1.8,1.0) circle (2pt) node[right](p1) {skies};
   \draw [-latex] (origin.center) -- ($ (1.8,1.0) - (0.1,0.05)$);

   \fill (2.2,0.6) circle (2pt) node[right](p1) {clouds};
   \draw [-latex] (origin.center) -- ($ (2.2,0.6)- (0.1,0)$);
\end{tikzpicture}

%% file: ch-hpylm/ch-hpylm.tex

\chapter{\label{ch:hpylm}A Hierarchical Bayesian Language Model for Compounding} 

\chapoverview{%
In this chapter we introduce an \ngram-based LM that exploits 
the sub-structure of single-token compound words to attain better smoothing.
The LM extends the Hierarchical Pitman-Yor language model \citep{Teh2006} with
the ability to provide probabilities for out-of-vocabulary compound words.
This chapter is an extension of material originally published as
\citep{BothaEACL} and \citep{Botha2012}.}

\emph{Compounding} is a process that forms words by combining two or more other
words. For example, in English, the independent words \iteg{space} and
\iteg{ship} may combine to form the noun \iteg{spaceship}, while \iteg{inspiring} and
\iteg{awe} form the hyphenated adjective \iteg{awe-inspiring}.
Many languages make use of compounding as an important vehicle for novel word formation. 
Our focus on compounding is motivated by the prevalence of \emph{closed
  compounds} in certain languages. Such compound words are written as single
orthographic units, and can pose problems to LMs, and NLP systems generally,
that regard punctuation-delimited tokens as their elementary units of analysis.
Closed compounds are especially problematic in languages that exhibit highly productive
compounding, such as German, Dutch and Afrikaans. In these instances,
compounding can occur recursively, theoretically without bounds. Word-based
LMs are thus prone to suffer from sparse data effects that can be attributed to
compounds specifically. The aim of the model we propose is to overcome data
sparsity due to compounding by accounting for compound structure.

A compound is said to consist of a \emph{head} component and one or more
\emph{modifier components}, with optional \emph{linking elements} between
consecutive components \cite{Goldsmith98}.

\paragraph{Examples of German compounds}
\begin{itemize}
\item A basic noun-noun compound: 
  \subitem \iteg{Auto} + \iteg{Unfall} = \iteg{Autounfall} \iten{car crash}
\item \emph{Linking elements} can appear between components
  \subitem \iteg{Küche} + \iteg{Tisch}
  = \iteg{Küche\underline{n}tisch} \iten{kitchen table}
\item Components can undergo stemming
  \subitem \iteg{Schul\underline{e}} + \iteg{Hof} = \iteg{Schulhof} \iten{schoolyard}
\item Compounding is recursive
  \subitem (\iteg{Regen} + \iteg{Schirm}) + \iteg{Hersteller}
    \subitem  = \iteg{Regenschirmhersteller} \iten{umbrella manufacturer}
\item Compounding extends beyond noun components
  \subitem \iteg{Zwei-Euro-Münze} \iten{two Euro coin} 
  \subitem \iteg{Fahrzeug} \iten{vehicle} -- from \iteg{fahren} \iten{to drive}
  + \iteg{Zeug} \iten{stuff/thing}
\end{itemize}


\section{Structural dependencies}
\label{sec:hpylm:struct-dep}
The linguistic intuition that we propose to exploit in our language model is
that the head of a compound word determines the morphosyntactic properties of the whole
word \cite{Williams1981}. We focus our investigation on German, which is
predominantly right-headed.
The right-most component of a compound thus determines the agreement
requirements that needs to be fulfilled between the compound itself and other
words in the sentence.
For example, the \iteg{Bahn} \iten{track/rail} in the compound \iteg{Eisenbahn}
\iten{railway} identifies the word as singular feminine, which determines the
requirements for its agreement with verbs, articles and adjectives.

A language model could therefore give a reasonable assessment of the syntactic
fluency of a sequence of German words by abstracting away the non-head components of
compounds.  For example, the sentence \iteg{I'm going by train} can be rendered in
German as either of the following:
\begin{itemize}
\item Ich fahre mit der Eisenbahn.
\item Ich fahre mit der Bahn.
\end{itemize}

Collapsing all compounds to their heads and ignoring modifiers would decrease
sparsity and allow more robust \ngram probabilities to be estimated from data.
But such a strategy would not be probabilistically sound as a generative model
of a corpus.  Moreover, a model that ignores modifiers would assign the same
probability value to \iteg{Eisenbahn} and the empirically much rarer
\iteg{Bobbahn} \iten{bobsled}, which would be unsatisfactory in a task where the language model
plays a discriminative role.  The model needs to account for the non-head
components in some way.  We expect the identity and number of modifier
components to be strongly correlated with the identity of the head.  In
particular, the conditional distributions of modifier given head will be sharply
peaked.  A simple approximation is thus to assume that, conditioned on the head,
modifiers are generated by a reverse \ngram model:
\begin{equation*}
P(\tpr{eisenbahn}\mid \tpr{mit der})
\equiv P(\tpr{bahn}\mid \tpr{mit der}) \times
P(\tpr{eisen}\mid \tpr{bahn}) \times
P(\tpr{\$}\mid \tpr{eisen}) 
\end{equation*}

The end symbol \$ indicates the word boundary and doubles as a control on the
number of modifiers.  In general, we will use this process as a \emph{back-off}
strategy, i.e., when the trigram \emph{mit der Eisenbahn} is unobserved. Note that
this is markedly different from linguistically na\"{\i}ve back-off models that
would score the unobserved trigram \emph{mit der Eisenbahn} by falling back on
bigram or unigram estimates.  In our model, we instead permit the model to back
off to this decomposition before dropping valuable context information.

\begin{figure}[tbp]
\centering
\includegraphics[width=110px]{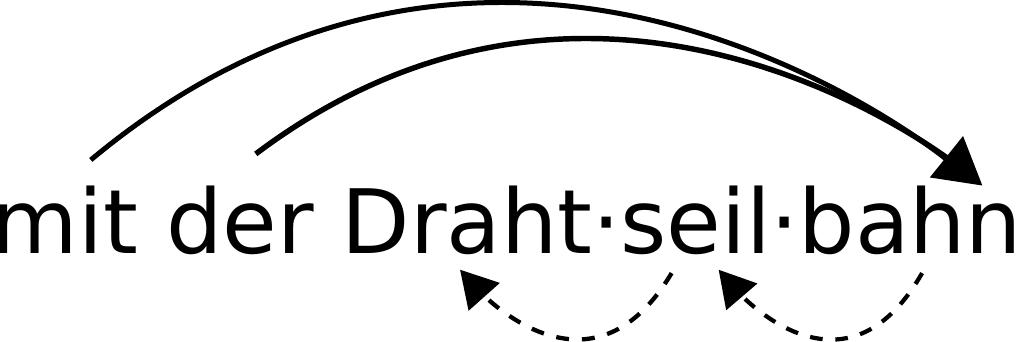}
\caption[Intuition behind generative process for compound words]%
{Intuition for the proposed generative process of a compound word: The
context generates the head component, which generates a modifier component,
which in turn generates another modifier. (Literal translation: ``with the cable
car'')}
\label{fig:hpylm:generative_intuition}
\end{figure}

\section{An \ngram model with productive compounding}
\label{sec:hpylm:model}
This section defines a hierarchical Bayesian model that embeds the
aforementioned intuitions about compound formation into an \ngram language
model. The approach is to extend the Pitman-Yor language model
\citep[HPYLM;][]{Teh2006} by modifying its base distributions to include a
sub-model for generating a compound's modifiers conditional on its head.

A seemingly obvious alternative method of integration would be to to use two
distinct word-level and compound-level \ngram models, independently built using
existing heuristic smoothing algorithms.  However, the advantage of the unified
hierarchical Bayesian model proposed here is that it can learn an 
interpolation between those levels, obviating the need to introduce and tune an
extraneous interpolation scheme between sub-models, while opening the door for
future extensions, \eg analysing compounds occurring in the \ngram history.

\subsection{Hierarchical Pitman-Yor Language Model (HPYLM)}
\label{sec:hpylm:hpylm}
The HPYLM applies the technique of iteratively shortening the context of an
\ngram to affect smoothing over an \ngram conditional distribution,
as used by conventional \ngram LMs. 
Let $\mb{u}$ denote the context of an \ngram \tokseq{w}{1}{n}, \ie
$\mb{u}=[w_1,\dots,w_{n-1}]$, and define $\pi(\mb{u})$ as a function that
truncates $\mb{u}$ by dropping the left-most context-word.

The HPYLM defines the distribution $G_{\mb{u}}$ of a word $w$ given its context
$\mb{u}$ as being drawn from a Pitman-Yor process with hyperparameters
$a_{|\mb{u}|}$ and $b_{|\mb{u}|}$.
The base distribution
is taken to be the
distribution $G_{\pi(\mb{u})}$ governing the probability that the word $w$
follows the truncated context
$\pi(\mb{u})$, as in the case of back-off models.
Since this latter distribution $G_{\pi(\mb{u})}$ likewise needs to be modelled,
we assume it is itself drawn from another PYP, which has hyperparameters
$a_{|\mb{u}|-1}$ and $b_{|\mb{u}|-1}$. This procedure is
applied recursively until reaching the unigram distribution $G_\emptyset$, for
which a final PYP prior is used, with its base distribution $G_0$ being the uniform
distribution over the vocabulary, $P(w)=1/|\Set{W}|$.

Conditioned on the context $\mb{u}$, a word $w$ is generated as follows:
\begin{align}
  w &\sim G_{\mb{u}} \\
  G_{\mathbf{u}} &\sim \pyp(a_{|\mathbf{u}|}, b_{|\mathbf{u}|},G_{\pi(\mathbf{u})}), \\
G_{\pi(\mathbf{u})} &\sim \pyp(a_{|\mathbf{u}|-1}, b_{|\mathbf{u}|-1},G_{\pi \circ \pi(\mathbf{u})}) \\
&\vdots \notag \\
G_{\mathbf{\emptyset}} &\sim \pyp(a_0, b_0, G_0)  \\
G_0 &= \mathrm{Uniform}(|\wvoc|) 
\end{align}

\subsection{HPYLM + compounds (HPYLM+c)}
\label{sec:hpylm:hpylmc}
We define a compound word $\tilde{w}$ as a sequence of components $[c_1, \dots,
c_\zL]$, plus an end symbol \$ marking either the left or the right boundary
of the word, depending on the direction of the model.  To maintain generality
over this choice of direction, let $\Lambda$ be an index set over the positions,
such that $c_{\Lambda_1}$ always designates the head component.

Following the motivation in \S\ref{sec:hpylm:struct-dep}, we introduce an
additional level of back-off where the head component $c_{\Lambda_1}$ is
conditioned on the context $\mathbf{u}$ of the word, while the remaining
components $c_{\Lambda_2},\dots,c_{\Lambda_L}$, as well as an end-of-sequence
symbol \$, are generated by a modifier model
$F$, independently of the context $\mathbf{u}$. We use a bigram HPYLM for $F$.

\begin{figure}[tb]
\centering
\begin{minipage}[b]{0.35\textwidth}
\centering
\includegraphics[width=1\columnwidth]{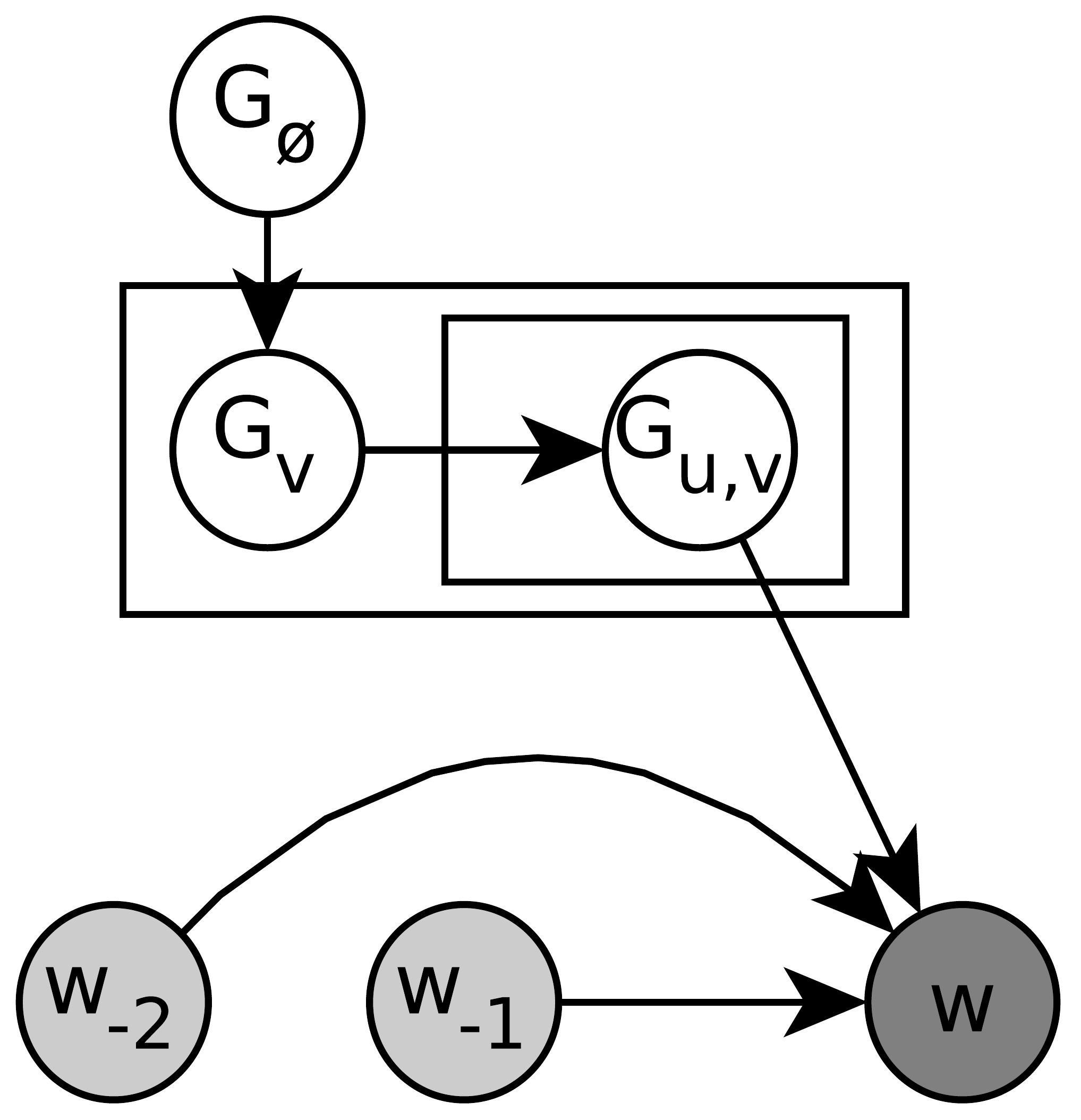}
\subcaption{Trigram HPYLM generating word $w$.}
\label{fig:hpylm:plate_hpylm} 
\end{minipage}
\quad \quad
\begin{minipage}[b]{0.55\textwidth}
\centering
\includegraphics[width=1\columnwidth]{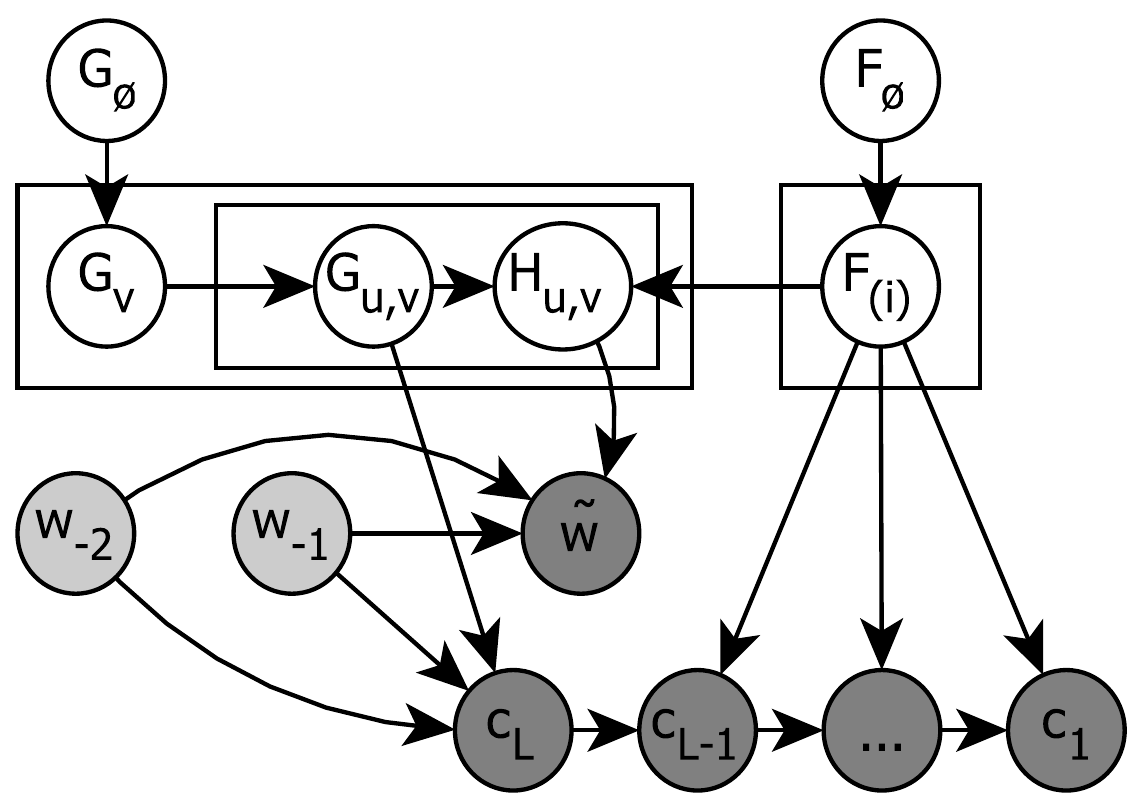}
\subcaption{Trigram HPYLM+c generating word $\tilde{w}$ according
  to the conditioning scheme $\Lambda^{\text{ling}}$, such that $c_\zL$ is the
  head and $c_{1}\dots c_{\zL-1}$ are the modifiers. }
\label{fig:hpylm:plate_hpylmc} 
\end{minipage}
\caption[Plate diagrams for HPYLM and HPYLM+c models]%
{Plate diagrams illustrating the generative structure for two trigram models
 HPYLM (left) and HPYLM+c (right).
Variables $w_{-2}$ and $w_{-1}$ denote the two words in the trigram context. 
Hyperparameters and their priors are omitted for clarity.}
\end{figure}

The two main modifications we make to the word-based HPYLM to obtain the HPYLM+c are as follows:
\begin{enumerate} 
\item Replace the support of the base distribution at the root of the hierarchy with the reduced vocabulary $\mvoc$, the set of unique
elementary components $c$.
$\mvoc$ also includes word types consisting of a single component to begin with.
\item Add an additional level of conditional distributions $H_{\mathbf{u}}$
(with $|\mathbf{u}|=n-1$) where items from $\mvoc$ combine to form the observed
compound words.
\end{enumerate}
The process for generating a word from the HPYLM+c is as follows (see also
Figure~\ref{fig:hpylm:plate_hpylmc}):
\begin{align}
\textit{Draw head: }\quad c_{\Lambda_1}    &\sim G_\mb{u} \\
G_{\mathbf{u}} &\sim \pyp(a_{|\mathbf{u}|}, b_{|\mathbf{u}|},G_{\pi(\mathbf{u})}),\\
G_{\pi(\mathbf{u})} &\sim \pyp(a_{|\mathbf{u}|-1}, b_{|\mathbf{u}|-1},G_{\pi \circ \pi(\mathbf{u})}) \\
&\vdots \notag \\
G_{\mathbf{\emptyset}} &\sim \pyp(a_0, b_0, G_0)  \\
G_0              &= \mathrm{Uniform}\left(|\mvoc|\right)  \\
\textit{Draw modifiers: }\quad c_{\Lambda_i} &\sim F_{c_{\Lambda_{i-1}}}, \quad
2 \leq i \leq L \quad \text{(\ie until \$ is drawn)} \\
F_{c} &\sim \pyp(a'_1, b'_1, F_\emptyset) \\
F_{\emptyset} &\sim \pyp(a'_0, b'_0, F_0) \\
F_0    &= \mathrm{Uniform}\left(|\mvoc|\right) \\
\textit{Create word: } \quad \tilde{w}&=\mathrm{concatenate}(c_1,\dots,c_L) \\
 \tilde{w}        &\sim H_{\mathbf{u}} \\
\label{eq:hpylm:draw-H-from-PYP}
H_{\mathbf{u}}   &\sim \pyp(a''_{|\mathbf{u}|}, b''_{|\mathbf{u}|}, G_{\mathbf{u}} \times F)
\end{align}

So the base distribution for the prior of the word \ngram distribution
$H_{\mathbf{u}}$ is the product of a distribution $G_{\mathbf{u}}$ over compound
heads, given the same context $\mathbf{u}$, and the modifier model $F$ which
conditions on the head component.

For German, the linguistically motivated choice for conditioning in $F$ is
$\Lambda^{\text{ling}}=[\zL, \zL-1, \dots, 1]$ such that $c_{\Lambda_{1}}$ is the
true head component; \$ is drawn from $F (\cdot | c_{1})$ and marks the left
word boundary.

In order to see if the correct linguistic intuition has any bearing on the
model's extrinsic performance, we will also consider the reverse, supposing that
the left-most component were actually more important in this task, and letting
the remaining components be generated left-to-right.  This is expressed by
$\Lambda^{\text{inv}}=[1,\dots,\zL]$, where \$ this time marks the right word
boundary and is drawn from $F (\cdot | c_{\zL})$.

\subsection{Example}
\label{sec:hpylm:tree-example}
This section provides an example of how the models are instantiated.
We illustrate the trigram case and focus on the structures related to the
following fragment of a hypothetical corpus:
\begin{center}
  \begin{small}
    \begin{tabular}{lp{0.5cm}l}
      {\bf Observed \ngram{}s} && {\bf Segmentation} \\ 
        \iteg{dem alten schulhof} \iten{the old schoolyard} && schul$\cdot$hof  \\
        \iteg{dem alten friedhof} \iten{the old cemetery} && fried$\cdot$hof  \\
        \iteg{des alten regierungschefs} \iten{the old head of government} && regierungs$\cdot$chefs \\
        \qquad \dots
    \end{tabular}
  \end{small}
\end{center}

\begin{figure}[t]
  \begin{minipage}[b]{0.48\textwidth} 
    \includegraphics[width=\textwidth]{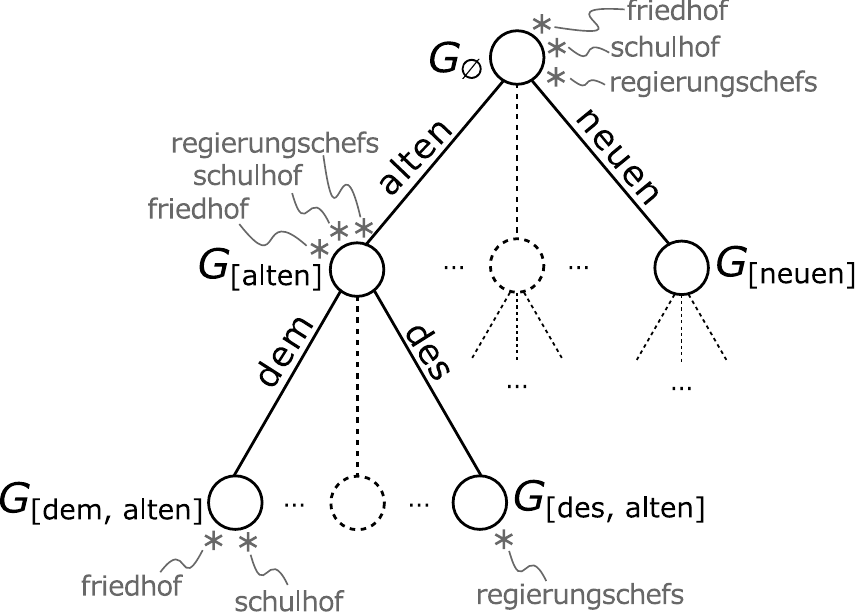}
    \vspace{4.6cm}
    \subcaption{HPYLM}
    \label{fig:hpylm:baseline-tree}
  \end{minipage}
  \hspace{0.01\textwidth}
  \begin{minipage}[b]{0.48\textwidth} 
    \includegraphics[width=\textwidth]{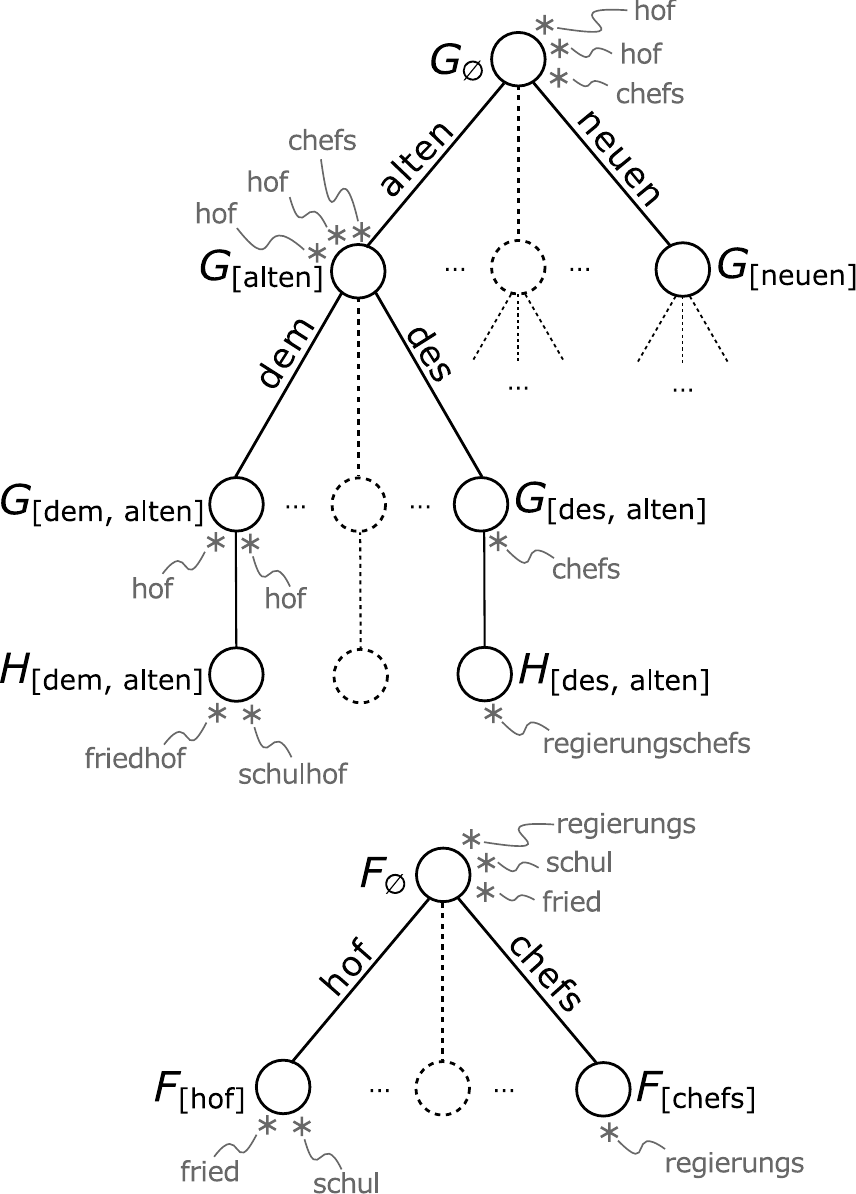}
    \subcaption{HPYLM+c}
    \label{fig:hpylm:my-tree}
  \end{minipage}
  \caption[Instantiation examples for HPYLM and HPYLM+c models]%
  {This visualisation
    illustrates the tree structures formed by the hierarchical priors in the
    HPYLM and HPYLM+c models in representing the corpus fragment given in
    \S\ref{sec:hpylm:tree-example}.
    Grey elements represent some of the customers seated within the various
    restaurants and each asterisk denotes a single customer. Only the minimal
    number of customers relevant to the example are shown explicitly.
  }
  \label{fig:hpylm:tree-structure}
\end{figure}

The hierarchies induced by these \ngram{}s in both models can be visualised as
trees, as shown in \autoref{fig:hpylm:tree-structure}.  The key thing to notice
is how the HPYLM+c combines the statistics of distinct compounds
(\iteg{schulhof} and \iteg{friedhof}) sharing a head, as indicated by the
presence of two \emph{hof}-customers in the restaurants of both
$G_{\text{[dem, alten]}}$ and $G_\emptyset$, unlike in the HPYLM.

Under the HPYLM, the probability of the word \iteg{schulhof} given the context \iteg{dem~alten},
conditioned on a particular seating arrangement across the hierarchy, is
computed recursively using \autoref{eq:bg:predictive-conditional}
(p.~\pageref{eq:bg:predictive-conditional}), as follows:
\begin{flalign}
  \;&P_{\text{HPYLM}}(\tpr{schulhof} \mid \tpr{dem alten}) &\\
  &=  G_{\text{[dem, alten]}}(\tpr{schulhof}) &\\
  &= \frac{N_{\text{schulhof}}^\RID{d.a.} - a_2 m_{\text{schulhof}}^\RID{d.a.} 
        }{ n^\RID{d.a.} + b_2}
      +
      \frac{\left(a_2 m^\RID{d.a.} + b_2 \right) G_{\text{[alten]}}(\tpr{schulhof} )
        }{n^\RID{d.a.} + b_2 } &
\end{flalign}
  with
\begin{align}
   G_{\text{[alten]}}(\tpr{schulhof}) 
  &= \frac{N_{\text{schulhof}}^\RID{a.} - a_1 m_{\text{schulhof}}^\RID{a.} 
        }{ n^\RID{a.} + b_1}
      +
      \frac{\left(a_1 m^\RID{a.} + b_1 \right) G_{\emptyset}(\tpr{schulhof} )
        }{n^\RID{a.} + b_1 }  \\
  G_{\emptyset}(\tpr{schulhof}) 
  &= \frac{N_{\text{schulhof}}^\mRID - a_0 m_{\text{schulhof}}^\mRID 
        }{ n^\mRID + b_0}
      +
      \frac{\left(a_0 m^\mRID + b_0 \right) G_0(\tpr{schulhof} )
        }{n^\mRID + b_0 }, 
\end{align}
where $G_0(\tpr{schulhof}) = 1/|\Set{W}|$.
The notation here uses superscripts to identify the associated restaurant in the
hierarchy. For instance, $N_{\text{schulhof}}^\RID{d.a.}$ is the number of customers
seated at tables labelled $\emph{schulhof}$ in the restaurant for the context
$\text{d.a.} \equiv \text{[dem, alten]}$, while
$m_{\text{schulhof}}^\RID{a.}$ is the number of tables
labelled $\emph{schulhof}$ in the restaurant for the context
$\text{a.} \equiv \text{[alten]}$.

Under the HPYLM+c, the same \ngram probability is computed as follows, with the main
difference being the product in the base distribution probability:
\begin{flalign}
  &P_{\text{HPYLM+c}}(\tpr{schulhof} \mid \tpr{dem alten}) &\\
  &=  H_{\text{[dem, alten]}}(\tpr{schulhof}) &\\
  &= \frac{N_{\text{schulhof}}^{\RID{d.a.}''} - a''_2 m_{\text{schulhof}}^{\RID{d.a.}''}
        }{ n^{\RID{d.a.}''} + b''_2 }  
         \\
  & \qquad  +  \frac{\left(a''_2 m^{\RID{d.a.}''} + b''_2 \right)
        \Big(G_{\text{[dem, alten]}}(\tpr{hof}) 
          \times F_{\text{[hof]}}(\tpr{schul}) 
          \times F_{\text{[schul]}}(\tpr{\$}) 
        \Big)
        }{ n^{\RID{d.a.}''} + b''_2 }. &
\end{flalign}
$G_{\text{[dem, alten]}}(\tpr{hof} )$ is computed similarly as before, but
now has the argument $\tpr{hof}$ and will bottom out with a term
$G_0(\tpr{hof})=1/|\Set{M}|$.
The cascade of computations in the modifier model has a similar form, for
instance:
\begin{align}
   F_{\text{[hof]}}(\tpr{schul}) 
   &= \frac{N_{\text{schul}}^{\RID{h.}'} - a'_1 m_{\text{schul}}^{\RID{h.}'} 
       }{ n^{\RID{h.}'} + b'_1}
      +
      \frac{\left(a'_1 m^{\RID{h.}'} + b'_1 \right) F_{\emptyset}(\tpr{schul} )
      }{n^{\RID{h.}'} + b_1 }  \\
  F_{\emptyset}(\tpr{schul}) 
  &= \frac{N_{\text{schul}}^{\mRID'} - a'_0 m_{\text{schul}}^{\mRID'}
    }{ n^{\mRID'} + b'_0}
      +
      \frac{\left(a'_0 m^{\mRID'} + b'_0 \right) F_0(\tpr{schul} )
      }{n^{\mRID'} + b'_0 }, 
\end{align}
where $\text{h.}\equiv\text{[hof]}$ and $F_0(\tpr{schul})=1/|\Set{M}|$.

\clearpage
\section{Inference}
\label{sec:hpylm:inference}

For ease of exposition we describe inference with reference to the trigram
HPYLM+c model, but the general case should be clear.
The model is specified by the latent variables 
$\mathcal{L}=(G_{[\emptyset]}, G_{[v]}, G_{[u,v]}, H_{[u,v]}, F_{\emptyset}, F_{c})$, 
where \mbox{$u,v \in \wvoc$}, \mbox{$c \in \mvoc$}, 
and hyperparameters
$\Omega=( a_i,b_i, a'_j,b'_j ,a''_2,b''_2 )$,
where $i=0,1,2$; $j=0,1$; single primes designate the hyperparameters in
$F$ and double primes those of $H_{[u,v]}$. 
By marginalising out the latent variables in $\mathcal{L}$, we can use
a collapsed Gibbs sampler to do inference via the hierarchical variant of the
Pitman-Yor Chinese Restaurant Process (PYCRP) \cite{Teh2006Report}.

When the prior of $G_{\mathbf{u}}$ has a base distribution $G_{\pi(\mathbf{u})}$
that is itself PYP-distributed, as in the HPYLM, the restaurant metaphor changes
slightly from the description in \autoref{ch:bg}.  In general, each context
$\mb{u}$ has an associated restaurant.  Whenever a customer is seated at a
previously unoccupied table in some restaurant $R$ which has a PYP-distributed
base distribution, a dish must be sampled from that base distribution.  That is,
a further customer is spawned and joins the parent restaurant $\text{pa}(R)$
associated with the context $\pi(\mb{u})$.  This induces a consistency
constraint over the hierarchy: the number of tables $m_\ell$ labelled with dish
$\ell$ in restaurant $R$ must equal the number of customers $N_\ell$ at tables
serving dish $\ell$ in the parent restaurant $\text{pa}(R)$.

We take care to satisfy this constraint in the HPYLM+c model, where a restaurant
associated with the distribution $H_{\mb{u}}$ has
as base distribution a product of PYP-distributed distributions, $F$ and
$G_{\mb{u}}$. Here, when a dish $\ell$ is drawn for a previously unoccupied table 
 in trigram restaurant $H_{[u,v]}$, a customer $c_{\Lambda_1}$
joins the corresponding trigram restaurant $G_{[u,v]}$, \emph{and} customers
$c_{\Lambda_{2}}, \dots, c_{\Lambda_{\zL}},\$$ are sequentially sent to the restaurants
associated with
$F_{c_{\Lambda_{1}}},\dots,F_{c_{\Lambda_{\zL}}}$, respectively.

\subsection{Posterior inference}
The Chinese restaurant representations effectively allows us to replace the
collection of Pitman-Yor priors with a collection of seating arrangements, $S$.
Marginalising over the seating arrangements $S$ and parameters $\Omega$ yields
the posterior predictive probability of a word $w$ given the
training data $\Set{D}$:
\begin{equation}
P(w \mid \mathcal{D}) = \int_{S,\Omega} P(w \mid S,\Omega) \, P(S,\Omega \mid \mathcal{D})\,dS\,d\Omega,
\end{equation}
This integral can be approximated by averaging over a number of posterior
samples $(S,\Omega)$ generated using Markov chain Monte Carlo methods. 
In particular, we can use a Gibbs sampler that explores the space of possible
seating assignments by iteratively resampling a single model variable conditioned on
all other model variables.
Given a current seating arrangement $S$, a new configuration $S'$ is sampled by
individually reseating each customer $x_i$ in the trigram restaurants
$H_{[u,v]}$. Reseating a customer $x_i$ means removing them
from their current table $z_i$ and assigning another table $k'$
drawn from the conditional distribution $P(z_i \mid \mb{z}^{-i})$.  Here,
$\mb{z}^{-i}$ denotes the table allocations
$(z_1,\dots,z_{i-1},z_{i+1},\dots,z_{|\mb{z}|})$ of the other customers in the given
restaurant.

If the original table $z_i$ only seated
a single customer, the removal of that customer triggers further customer
removals from the base distribution restaurants in accordance with the hierarchy
consistency constraints.

In the absence of any strong intuitions about appropriate values for the
hyperparameters, we place vague priors over them and use 
slice sampling \cite{Neal2003} to update their values during generation of the posterior
samples: $a \sim \text{Beta}(1,1)$ and $b \sim \text{Gamma}(10, 0.1)$.\footnote{%
    We used the slice sampler implementation released by Mark Johnson, \\
    {\scriptsize \url{http://web.science.mq.edu.au/~mjohnson/Software.htm}}
}

Lastly, we follow the pragmatic course of using a single sample
$(S,\Omega)$ to approximate the integral over the posterior.\footnote{%
    Preliminary experiments (C. Dyer, p.c.) indicated that the posterior over the
    latent model structure is quite sharply peaked, so that a single sample
    constitutes a low-variance estimator of the posterior predictive distribution.
}

\subsection{Sampler description}
\label{sec:hpylm:sampler-description}
The previous section outlines the Gibbs sampler in broad terms. In this section,
we detail the algorithms involved.

The scheme described above poses the challenge of keeping track of a table
assignment for each individual customer in every restaurant. Given that language
models need to be trained on substantial amounts of data, this bookkeeping can
be problematic, both in terms of memory and implementation.

\citet{Blunsom2009} introduced an efficient solution to this problem, which
leverages the exchangeability property of the hierarchical Dirichlet process and
which carries over straightforwardly to the PYP. Instead of tracking what table
each customer is seated at, the authors suggest maintaining histograms of table
occupancy. When a customer has to be added or removed from a restaurant, it is
done by sampling directly from the histogram. A histogram $\HHist{w}$ maps a table occupancy
value $t$ to the number of tables labelled with $w$ and having $t$ customers,
$\Hist{w}{t}$.

As a brief example, suppose we have an assignment of 6 customers to tables as
$z_1$=1, $z_2$=1, $z_3$=2, $z_4$=1, $z_5$=3, $z_6$=4 and the tables are labelled as
$\ell_1$=$\ell_3$=$\textit{sun}$, $\ell_2$=$\ell_4$=$\textit{moon}$. The equivalent
histogram representation is
$\Hist{\textit{sun}}{3}$=1, $\Hist{\textit{sun}}{1}$=1
and
$\Hist{\textit{moon}}{1}$=2.

The two basic operations required for updating the seating arrangement in the
HPYLM+c are \textsc{AddWordToModel} 
and \textsc{RemoveWordFromModel}, as defined in Algorithms~\ref{algfloat:add}--\ref{algfloat:remove}.
To facilitate reading the algorithms in conjunction with the mathematical
definitions given before, the psuedocode retains the quantities $N_w$, $m_w$,
and $m$, but we note that these can all be reconstructed from the histogram
representations as follows:
\begin{align}
N_w &= \sum_{t \in \HHist{w}} t \times \Hist{w}{t} \\
m_w &= \sum_{t \in \HHist{w}} \Hist{w}{t} \\
\text{and } m &=\sum_w m_w \text{, as stated earlier.}
\end{align}

\input{algs}

\clearpage
\section{Experiments}
We evaluate the proposed compound model HPYLM+c on German in terms of its intrinsic
performance (\S\ref{sec:hpylm:intrinsic_eval}), and by its influence when used as part of a machine translation
system (\S\ref{sec:hpylm:mteval_into}). 
In both settings, we compare against baselines obtained with the standard HPYLM
as well as the Modified Kneser-Ney (MKN) LM.

The next two sections describe further methodology details.

\subsection{Data preparation} 
\label{sec:hpylm:dataprep}
We use corpus data released as part of the 2011 ACL Workshop on Machine Translation's
shared task\footnote{{\scriptsize \url{http://www.statmt.org/wmt11/}}}.
Standard data preprocessing steps included normalising punctuation, tokenising and lowercasing all words.

For language model training, we used the combination of the corpora
\corpus{Europarl}, \corpus{news-commentary} and \corpus{news-2011}, filtered to
exclude duplicate sentences. This amounts to 59 million word tokens.

\begin{table}[tb]
\centering
\begin{minipage}[b]{0.42\textwidth}
\begin{tabular}{lccc} \toprule
                & En & De & De LM\\ \midrule
Sentences       & 1.7m    & 1.7m   & 2.4m \\
Word Tokens          & 49m     & 38m    & 59m  \\
Word Types     & 112k     & \multicolumn{2}{c}{351k} \\ \bottomrule
\end{tabular}
\subcaption{Statistics of training corpora.} 
\label{tbl:hpylm:corpus_stats}
\end{minipage}
\quad\quad
\begin{minipage}[b]{0.48\textwidth}
\begin{tabular}{cr} \toprule
Parts/Compound     & \multicolumn{1}{c}{\# Compound Types} \\ \midrule
2         & 197233 \\
3         & 25128 \\
4         & 1194 \\
$\geq$5   & 59 \\ \midrule
\multicolumn{1}{r}{\emph{Total number:}}     & 223614 \\ \bottomrule
\end{tabular}
\subcaption{Compound types by length.}
\label{tbl:hpylm:compound_lengths}
\end{minipage}
\caption[Statistics of training data and compound words contained therein]%
{Summary of training data and compound segmentation. The combined reading is that 223614 of the 351k word types are designated as compounds.}
\end{table}

The language model vocabulary $\Set{W}$ consists of all word types appearing in the target-side of
the bitext. As the machine translation
system used in the subsequent extrinsic evaluation cannot output word types
outside of this vocabulary, the choice was made not to model anything beyond
that.\footnote{See \S\ref{sec:hpylm:mt_dataprep} for details on bitext
  preprocessing.}
LM training tokens outside of that vocabulary were replaced by the \unk symbol.
Models were trained using all resulting $4$-grams; \ie no pruning was done on
low-frequency \ngrams.

Test data for language model evaluation comprised the combination of the
news-test corpora of 2008--2010, released with WMT11.
Statistics are provided in \autoref{tbl:hpylm:corpus_stats}.

\subsection{Compound segmentation method}
\label{sec:hpylm:segmentation-method}
For this evaluation, we used a deterministic, \emph{a priori} segmentation of
compound words into their parts. This means we assume a single, fixed analysis
of a compound regardless of the context it occurs in, which is necessitated by
the fact that our probabilistic model does not specify a step for choosing an
analysis.  To construct a segmentation dictionary, we used the one-best
segmentation output of a supervised
maximum-entropy-based compound splitter
\cite{Dyer2009} on all the words in the training vocabulary.
In addition, word-internal hyphens were also taken as
segmentation points.\footnote{Our choice of a supervised compound splitter 
  is based on a desire to focus the evaluation on the language model given high
  quality segmentations. The splitter of \citet{Dyer2009} obtained a segmentation
  F-score of \mytilde95\% in their evaluation.
  Unsupervised methods could also be used with our model.}
We regard the set of compounds $\mathcal{C} \subset {\wvoc}$ as those word types that are split into two or more parts by this procedure.  Similarly, $\mvoc_\mathcal{C} \subset \mvoc$ is the set of unique word parts that result from segmenting the compounds $\mathcal{C}$.
The majority of compounds thus identified consist of two or three parts
(Table~\ref{tbl:hpylm:compound_lengths}).

In our method, linking elements do not constitute items in the vocabulary of
word components $\mvoc$. 
Regarding linking elements as components in their own right would
sacrifice important contextual information and disrupt the conditional
bigram distributions, $F( \cdot | c_{\Lambda_{i-1}})$, generating compound modifiers.
That is, faced with the compound
küche$\cdot$\underline{n}$\cdot$tisch \iten{kitchen table}, we want the generation of the modifier to
be according to $P(\text{küche} |
\text{tisch})$, and not $P(\text{küche} | \text{n})$.
To account for linking elements we follow the pragmatic option
of merging any linking element onto its adjacent component---for
$\Lambda^{\text{ling}}$, merging happens onto the preceding component (\ie using
$P(\text{küchen} | \text{tisch})$), while for  $\Lambda^{\text{inv}}$, the
linking element is merged
onto the succeeding component (\ie using $P(\text{ntisch} | \text{küche})$).
This ensures that the set of head components  $\{c_{\Lambda_{1}}\}$ is not fragmented.\footnote{%
    It is worth noting that for German the presence and identity of linking elements
    between consecutive components $c_i$ and $c_{i+1}$ are largely governed by
    the first component $c_i$ \cite{Goldsmith98}. A more sophisticated model could incorporate this
    bias. }
\autoref{tbl:hpylm:seg} summarises the effect of these different merging options on
the size of the vocabulary $\Set{M}$ of word components modelled.

\begin{table}[tbp]
\centering
\begin{tabular}{lrrl} \toprule
  \textbf{Linking element handling} 
  & $|\mvoc_\mathcal{C}|$ 
  & $|\mvoc|$ & \textbf{Segmented example} \\ \midrule
\emph{(no segmentation, \ie $\mvoc=\wvoc$)}
              & 223614 & 350997 & Küchentisch \\
delete 
              & 45183 & 144358 & Küche$\cdot$tisch\\
merge left  ($\Lambda^{\text{ling}})$ 
              & 50868 & 152323 & Küchen$\cdot$tisch\\
merge right ($\Lambda^{\text{inv}})$ 
              & 63315 & 164095 & Küche$\cdot$ntisch\\ 
      \bottomrule 
\end{tabular}
\caption[Linking elements and base distribution support]%
{Effect of linking elements on the number of unique word components
  modelled. The second column is the number of components measured only over the set of compounds $\mathcal{C}$, while the third
  column is over all words $\wvoc$, showing what the size of $\mvoc$, the support of the base distribution $G_{\mathbf{\emptyset}}$,
would be under the alternative schemes for handling linking elements.} \label{tbl:hpylm:seg}
\end{table}

\subsection{MCMC sampler configuration}
\label{sec:hpylm:sampler-config}
We initialised the samplers used for the HPYLM and HPYLM+c with random seating
assignments instead of basing the initial configuration on the actual
conditional distributions for adding customers to restaurants. This worked
satisfactorily and was convenient in our implementation, so we did not experiment
with alternative initialisation schemes.

\begin{figure}[tb]
  \centering
  \includegraphics[trim=0 3ex 0 0,clip,%
                    width=0.8\textwidth]{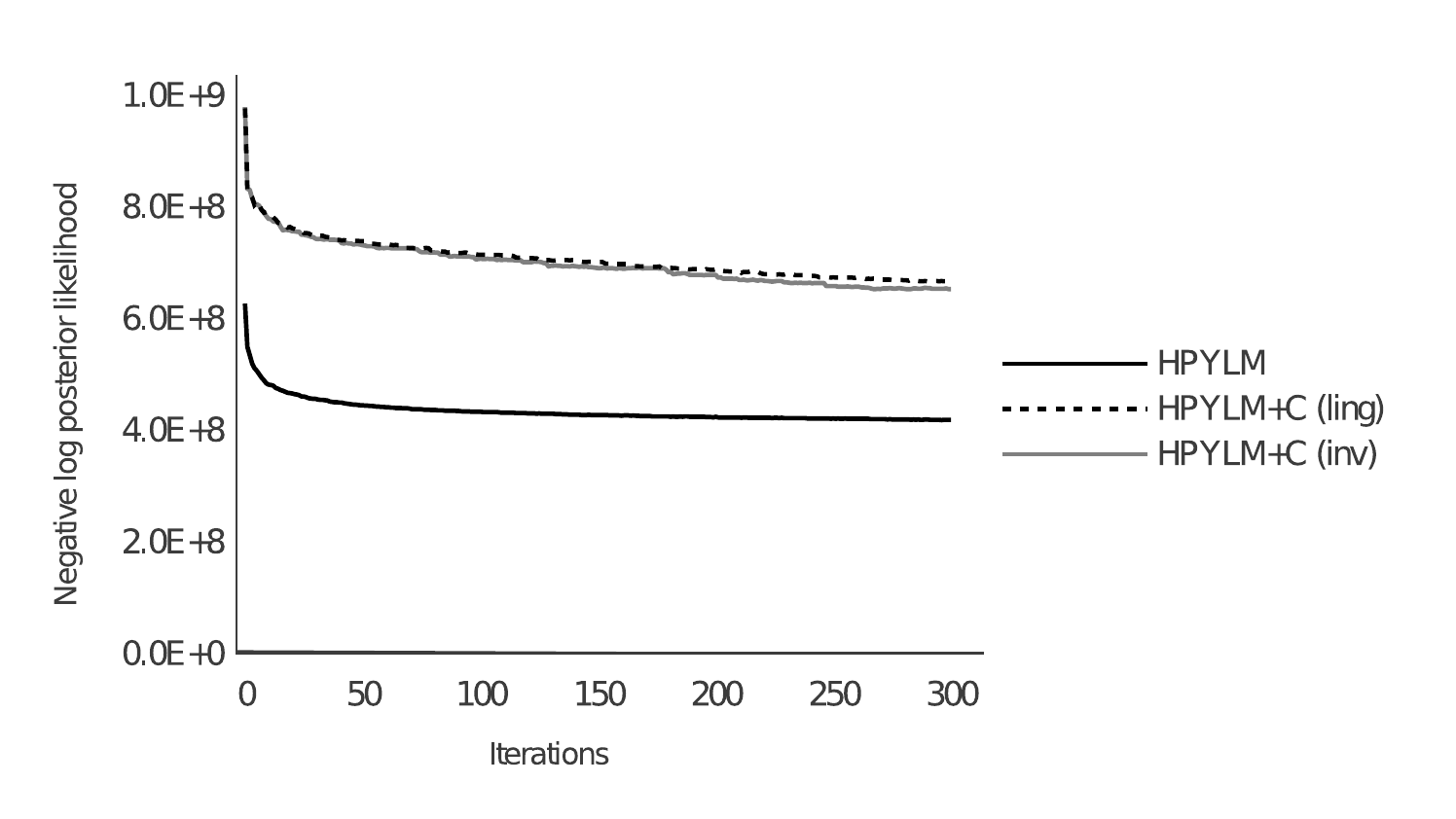}
  \caption[Sampler burn-in]%
  {Sampler burn-in for the \kgram{4} models.  }
  \label{fig:hpylm:LLplot}
\end{figure}

Each sampler was then run for 300 iterations of burn-in, and we gauged
convergence toward the underlying stationary distribution at the hand of the
posterior likelihood (\autoref{fig:hpylm:LLplot}).
This approach takes a cue from earlier work on the HPYLM on non-trivial
amounts of training data, where a similarly low number of iterations was used \cite{Huang2010}.
One consideration for why this works reasonably well---in light of the favourable
comparisons of the HPYLMs to traditional back-off LMs---is that the model does
not need to infer complex latent structure, as may be the case in other MCMC
applications where vastly more iterations are standard practice. The identities
of word tokens are fully observed, so that the sampler primarily has to explore
different seating arrangements across the hierarchy of Chinese restaurant
processes. Moreover, the training data tends to be sparse, so that
the range of possible configurations for the restaurants corresponding to rare
\ngrams are limited by a paucity of customers.

\subsection{Intrinsic evaluation}
\label{sec:hpylm:intrinsic_eval}
The modelling approach proposed in this chapter is premised on the hypothesis that better
smoothing of \ngram distributions can be obtained by analysing compounds into their
components.
This intrinsic evaluation aims to test that hypothesis and to explore the
relative performance of the model variants that make different assumptions
around the conditional dependencies of compound parts.

The HPYLM+c models were compared against two baseline methods, one using MKN smoothing
and the other being the HPYLM, both without any special handling of compounds.
\footnote{The SRILM toolkit was used for training and testing the MKN models
  \cite{Stolcke02srilm}.}

\subsubsection{Results}

\begin{table}[tbp] 
\centering
\input{data/main_ppl_B}

\caption[Test-set perplexities for primary model variants]%
  {Test-set perplexity results for two variants of the proposed
  compounding language model and two baseline language models. The last two
  columns give the percentage perplexity reduction obtained by the model of the
given row relative to each baseline.} 
\label{tbl:hpylm:result_mono}
\end{table} 

Test-set perplexities are summarised in \autoref{tbl:hpylm:result_mono}.
We find that the HPYLM obtains a perplexity of 294.0 on our test-set, lower than
the MKN perplexity of 299.9. This outcome is consistent with earlier results 
that the HPYLM outperforms the MKN model for training data sets of non-trivial sizes 
\citep{Huang2009}.

The HPYLM+c that assumes right-headed compounds ($\Lambda^{\text{ling}}$)
virtually matches the performance of the HPYLM baseline, a result that we
consider in more detail shortly.  The assumption of left-headed compounds
($\Lambda^{\text{inv}}$), which is \emph{inconsistent} with the known properties of
German, leads to a marked degradation in perplexity. The HPYLM+c with this
assumption achieves perplexity of 305.5, which is higher than the perplexity of
both baseline models.  This result confirms the validity of our strategy to
condition actual compound heads on sentential context.

The aforementioned comparison features a bias toward the word-based baseline
models, since the perplexity of different models on a given data set is directly comparable
only if the models are defined over the same vocabulary.
This is not the case here---according to the generative process of the HPYLM+c
(\S\ref{sec:hpylm:hpylmc}), the generation of an additional modifier component
$c\in \mvoc$ for any partially formed token $\tilde{w}$ always carries non-zero probability.
HPYLM+c thereby defines a distribution over a countably infinite set of word
types, compared to the finite vocabularies modelled by the word-based HPYLM and
MKN. The preceding results therefore establish a type of upper bound on the
perplexities of the HPYLM+c models on the test-set.

To obtain a fairer comparison, we performed the costly renormalisation of
the right-headed HPYLM+c over the finite vocabulary of words modelled by the
baseline LMs. We find that the test-set perplexity of the renormalised model is 292.2,
which is below the 294.0 of the HPYLM baseline. The fact that the difference in
perplexity between the infinite model and its renormalisation is not too great
suggests that the HPYLM+c reserves a reasonable amount of probability mass for
words outside the finite vocabulary; it would not be ideal if too large a
portion of mass was allocated beyond that vocabulary, since ``most'' items in
the countably infinite domain of the model would not constitute well-formed compounds
that could practically occur in the language.

\subsubsection{Effect of Markov order}
The baseline word-based language models deal with unseen test \ngrams by progressively
truncating the context to rely on lower-order distributions for which the
training data is less sparse. The simplifying assumption encoded by this
strategy is that novelty in \ngrams arises strictly from the high likelihood of
encountering a new element from the space of $|\wvoc|^n$ possible \ngrams.

In the HPYLM+c, the back-off procedure for handling unseen \ngrams is first to 
decompose the predicted compound into its components, and thereafter to truncate
context. The underlying assumption is therefore that compounding is the primary driver of
novelty in \ngrams, and that the space of possible \ngrams should be viewed as
$|\wvoc|^{n-1}\times |\mvoc|^\zL$, where the number of components in a
compound, $\zL$, is in principle unbounded but in practice likely to be bounded at the order of
10.

The aforementioned assumption in the HPYLM+c is violated when compounding is
\emph{not} responsible for an unseen \ngram. The model would then prematurely
decompose the predicted word when in fact immediately truncating the context
without decomposing the predicted word would result in matching some observed
$k$-gram ($1\leq k \leq n$).

The impact of this effect is measured in a further experiment by training
language models with lower maximum orders, $n=2$ and $n=3$.
At lower $n$, there should be fewer test tokens for which premature
decomposition could occur, in principle.
The results in Table~\ref{tbl:hpylm:order_scaling} show 
that the relative margin of improvement by HPYLM+c over HPYLM increases as $n$
is decreased from 4 to 2. This outcome implies that prioritising generalisation
through compound decomposition higher than generalisation through context
truncation tends to constrain the model's performance.

\begin{table}[tb]
\centering
\input{ch-hpylm/data/ppl_vary_order}
\caption[Test-set perplexities for different \ngram orders]%
{Test-set perplexity for models trained with different \ngram orders, along with
  relative differences at the bottom. HPYLM+c refers to the variant using $\Lambda^{\text{ling}}$.
}
\label{tbl:hpylm:order_scaling}
\end{table}

\paragraph{Convergence of lower-order models}
For $n=2$ and $n=3$, the sampler had not fully converged after 300 iterations.
We suspect this is due to the higher entropy in the distributions governing the
seating assignments: If $n=2$, there should be more customers (hence more
possible seating configurations) in the average restaurant for a particular
context-length $j<n$ than in the same restaurant if $n$ is larger.  This did not
affect perplexity, which was stable when evaluating with different individual
samples from the posterior around 300 iterations.

\subsubsection{Breakdown by different test subsets}
\label{sec:hpylm:pplbreakdown}
A more fine-grained view of model performance can be obtained by partitioning test
tokens into meaningful subsets and computing the perplexity over the test tokens
in each subset.
In this section, the subsets are characterised by the type of token
(compounds vs.\ non-compound) and the ``hit-length'' $h$, where $h=1$ means the
token appears in an unseen context and the language model has to rely on the
unigram distribution to score it, through to $h=n$, for which the full \ngram
occurred in training data. Note that $h$ is defined from the view-point of a
standard \ngram model, using surface-level matching, and does not include
decomposition.

\begin{figure}[tb]
  \centering
  \includegraphics[width=0.7\textwidth]{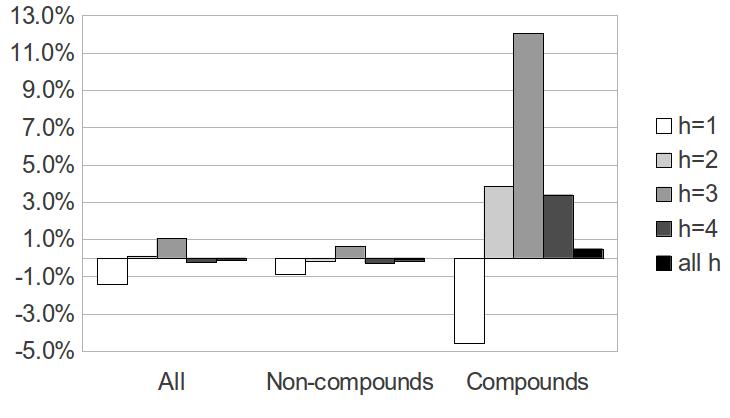}
  \caption[Improvements broken down by context hit-length and compound status]%
{Percentage relative cross-entropy reduction of HPYLM+c with respect
    to the baseline HPYLM, broken down by test token hit-length $h$ and compound
    status (defined in \S\ref{sec:hpylm:pplbreakdown}). Both are 4-gram models. Bars going down imply improvement over the
    baseline. 
  }
  \label{fig:hpylm:rel_xent_hpylmc_vs_hpylm}
\end{figure}

\begin{figure}[tb]
  \centering
  \includegraphics[width=0.5\textwidth]{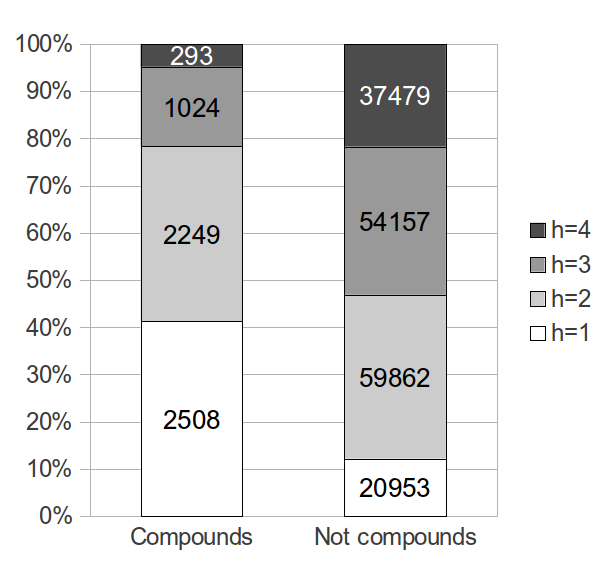}
  \caption[Test-set broken down by context hit-length and compound status]%
  { Composition of test-set in terms of number of tokens with hit-length $h$ and compound
    status. The total 178525 test tokens breaks down into 6074 compounds and
    172451 non-compounds, the latter including the end-of-sentence padding and
    OOVs, which are handled with mapping to an \unk symbol. 
  }
  \label{fig:hpylm:testset-composition}
\end{figure}

The results of this breakdown are reported in terms of relative cross-entropy
reductions in Figure~\ref{fig:hpylm:rel_xent_hpylmc_vs_hpylm} and the sizes of
the different subsets are summarised in \autoref{fig:hpylm:testset-composition}.\footnote{This analysis
  uses cross-entropy rather than perplexity because the dynamic range of
  perplexity varies drastically across the test subsets, which impedes
  interpretation when presenting results on a single visual scale.}

The breakdown indicates that HPYLM+c improves over the baseline
HPYLM only for test tokens appearing in fully unseen contexts ($h=1$); this
improvement is larger for compounds (-4.6\%) than for non-compounds (-0.9\%).
This improvement for non-compounds with $h=1$ is consistent with the
fact that many compound heads appear as words in themselves; their prediction
thus benefits from the way the HPYLM+c shares statistical strength across
instantiations of a given head.

The categories where the HPYLM+c performs worse than the baseline are for
trigram hits generally, and for compounds at hit-lengths $h\neq1$.

\subsubsection{Qualitative analysis}

\begin{table}[bt]
\centering
\small{
\begin{tabular}{rc} \toprule
HPYLM+c better & $\Delta$ \\ \midrule
gegen die umstrittene {\bf wieder+wahl}          &  0.058 \\
aufbau der afghanischen {\bf sicherheits+kräfte} &  0.036 \\
dessen zentralen {\bf gesichts+punkten}          &  0.035 \\
in annapolis , {\bf mary+land}                   &  0.035 \\
wochen vor den {\bf präsidentschafts+wahlen}     &  0.032 \\
dieses vertrauen nicht {\bf miss+brauchen}       &  0.030 \\
für psychiatrie und {\bf psycho+therapie}        &  0.028 \\
tage vor den {\bf parlaments+wahlen}             &  0.028 \\
reduktion der {\bf treibhausgas+emissionen}      &  0.025 \\
in einem unblutigen {\bf militär+putsch}         &  0.021 \\ \midrule
Baseline (MKN) better & $\Delta$ \\ \midrule
, newt {\bf ging+rich}                           &  0.511 \\  
nächtlichem {\bf flug+lärm}                      &  0.449 \\
generalsekretär ban {\bf ki-+moon}               &  0.423 \\
in st. {\bf peters+burg}                         &  0.420 \\
im 17. {\bf jahr+hundert}                        &  0.419 \\
saalpublikums in st. {\bf peters+burg}           &  0.359 \\
militanten klerikers moqtada {\bf al-+sadr}      &  0.352 \\
un-hochkommissarin für {\bf menschen+rechte}     &  0.286 \\
schwebt in {\bf lebens+gefahr}                   &  0.231 \\
der akademie der {\bf wissen+schaften}           &  0.212 \\ \bottomrule
\end{tabular} }
\caption[Compound words with greatest difference between model scores]%
 { Compounds from the monolingual test set for which HPYLM+c outperforms
  MKN by the largest margin (top) and \emph{vice versa} (bottom). We define the margin
  $\Delta$ as the difference in probability that the models assign to the given
  test \ngram.
 }
\label{tbl:hpylm:canned_mono} 
\end{table}

For a more qualitative insight into the model performance, we did a further
direct comparison of the HPYLM+c model and the MKN baseline by ranking test set
compounds by the difference in probability value that each model assigns to the
\ngram.  The test compounds where the compound model does best
(Table~\ref{tbl:hpylm:canned_mono} top) are all words for which an analysis into a
context-dependent head and modifiers should clearly be beneficial. For example,
in scoring the phrases \iteg{wochen vor den präsidentschaftswahlen} \iten{weeks before
  the presidential elections} and \iteg{tage vor den parlamentswahlen} \iten{days before
  the parliamentary elections}, the head \emph{wahlen} is having a mutually
reinforcing effect.

In contrast, we find that the cases where the MKN baseline model does best
(Table~\ref{tbl:hpylm:canned_mono} bottom) feature various words that are not strictly
speaking compounds, and feature over-segmentation by the supervised compound
segmentor, \eg \emph{ging+rich} and \iteg{wissen+schaften} \iten{sciences}, or through the greedy
splitting on hyphens, \eg \emph{ki-+moon}.  These are words where our compound
model's smoothing is hurting performance, since it allocates some probability
mass toward observing other modifiers with these heads, which in the case of these
proper nouns will not happen.  This is evidence of success on the part of our
model's underlying mechanism, but demonstrates that more care should be taken
with the particular segmentation method used.

\subsubsection{Effect of training data size}
\begin{figure}[tb]
\centering
\includegraphics[width=0.9\textwidth]{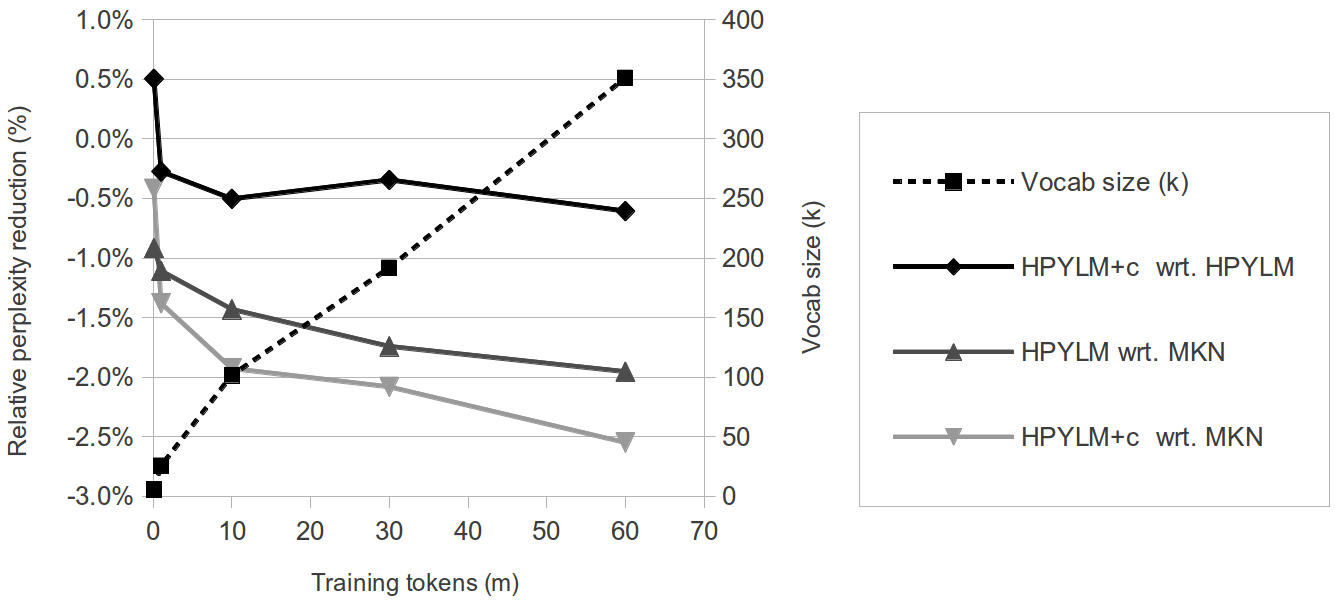}
\caption[Perplexity as a function of training data size]%
{ Relative perplexity differences of HPYLM+c against the two baseline
  models as the number of training tokens is varied from 100k to 60m tokens.
  $4$-gram models are used throughout.}
\label{fig:hpylm:data-scaling}
\end{figure}

We performed a scaling experiment to investigate the model's sensitivity to
training data size. The vocabulary at each training data size is determined
directly by the data (excluding singletons in order to learn reasonable
conditional distributions involving the unknown word symbol). Perplexity is
therefore not directly comparable from one size to another, hence we report
relative perplexity differences in \autoref{fig:hpylm:data-scaling}.

The results show that HPYLM+c provides a small but consistent improvement over
the HPYLM baseline as training data size is varied over two orders of
magnitude. The two Bayesian models outperform the MKN baseline by an
increasing relative margin as data size grows, and the additional benefit of
HPYLM+c over HPYLM is additive.
These trends are consistent with our intuition that productive compounding
affects language model performance not just at the smallest of data sizes;
extrapolation suggests consistent benefit could be derived from accounting for compounding could
at larger data sizes as well.

\subsection{Extrinsic machine translation evaluation}
\label{sec:hpylm:mteval_into}
Our intrinsic evaluation established that the HPYLM+c obtains small improvements
over the MKN and HPYLM baseline models.
This section reports on a machine translation experiment, where we compare the
quality of the output produced by the translation system when using different
language models. 

We measure translation quality with the automated metric \bleu \cite{Papineni2001}.

\subsubsection{System details}
\label{sec:hpylm:mt_dataprep}
This evaluation uses the hierarchical phrase-based translation system
implemented by the \tool{cdec} decoder \cite{Dyer2010}.

The English-German bitext consisted of Europarl and news commentary, filtered
after preprocessing to
exclude sentences longer than 50 tokens, resulting in 1.7 million parallel
sentences; word alignments were inferred from this using the Berkeley
Aligner
\cite{Liang2006} and used as a basis from which to extract a 
synchronous context-free grammar \citep[SCFG;][]{Chiang2007}.

The weights of the log-linear translation model were optimised towards the \bleu
metric using \tool{cdec}'s  implementation of Minimum
Error-rate Training \citep[MERT;][]{Och2003a}.  The development set
\corpus{news-test2008} (2051 sentences) was used for this weight tuning.
The evaluation result is \bleu measured on the test set \corpus{newstest2011}
(3003 sentences, 171460 tokens). 
To counter the known instability of MERT+\bleu \cite{JonClark2011}, three repetitions
were done and the results averaged.
 
\subsubsection{Language model feature function}
\label{sec:hpylm:MT_LM_feat}
The independent variable in this translation experiment is the language model.
The two baseline systems respectively use the 4-gram MKN and HPYLM baseline
language models from the preceding monolingual evaluation as a feature function in the
log-linear translation model.  To test the effect of the compounding LM on
translation, this feature function is replaced by the HPYLM+c variant in
question.  Each translation system therefore uses a single language model
feature.

\subsubsection{Translation results}
\label{sec:hpylm:MT_results}
\begin{table}[tbp]
\centering 
\begin{tabular}{lccl} \toprule
                                              & {\bf PPL} & \textbf{\bleu} & std.dev  \\ \midrule
MKN                                           & 299.9     & 13.9          & $\pm$0.3 \\
HPYLM                                         & 294.0     & 13.9          & $\pm$0.3 \\
HPYLM+c ($\Lambda^{\text{ling}}$)         & 294.1     & 13.9          & $\pm$0.1 \\
HPYLM+c ($\Lambda^{\text{inv}}$)          & 305.5     & 13.7          & $\pm$0.2 \\ \bottomrule
\end{tabular}
\caption[Machine translation results]%
{ Machine translation results obtained with different language models.
  The last two columns give \bleu mean and standard deviation over three runs.
For reference, we also show the perplexity of each (4-gram) language model on the monolingual test set.}
\label{tbl:hpylm:result_trans}
\end{table}

In terms of \bleu score, we do not find a meaningful difference between the
various systems (Table~\ref{tbl:hpylm:result_trans}).  The system using HPYLM+c
matches the two baselines, a result that indicates our more expressive modelling
is not sacrificing any performance in this task. This is an important outcome,
as it means we avoid a common pitfall whereby a new model is proposed to target
some specific phenomenon only to sacrifice performance
globally. Consistent with the intrinsic evaluation, the use of the left-headed
model leads to a small decrease in \bleu.

\subsubsection{Compound-specific evaluation}
\label{sec:hpylm:compound_PRF}
Next, we turn to a more fine-grained look at the translation output.  The \bleu
metric is likely to miss small improvements in translation quality.  Moreover,
in the reference translations of the test set, only 2652 of the 72661 tokens are
compounds; a moderate improvement in generating them is unlikely to have a big
impact on the \bleu.  

To assess the quality of the translation system in terms of compound words,
we use precision, recall and balanced F-score of hypothesised compounds with
respect to compounds in the reference sentences.
Specifically, let $R_i$ (resp.\ $H_i$) be the set of compound words occurring in
the $i$\th reference sentence (resp.\ hypothesis).  Precision $P$ and recall $R$ is
calculated through micro-averaging as
\[
  P = \dfrac{\sum_i |H_i \cap R_i|}{\sum_i |H_i|}
  \qquad \text{and} \qquad
  R = \dfrac{\sum_i |H_i \cap R_i|}{\sum_i |R_i|},
\]
while the F-score is the harmonic mean of precision and recall,
$F_1=\frac{2\times P\times R}{P+R}$.
These metrics are computed per individual system run and then averaged across
repetitions to obtain the percentage-based results shown in Table~\ref{tbl:hpylm:result_trans_acc}.

The right-headed HPYLM+c obtains an average improvement in compound precision 
of 9\%--13\% relative to the two baseline systems (2--3 absolute percentage
points).
It does so without a decrease in recall, hence the gain in precision is not
achieved simply by the system being more conservative about outputting compounds
in the first place.
The left-headed HPYLM+c performs marginally below the baseline methods in this
evaluation as well.

\begin{table}[tbp]
\centering 
\input{data/compound-PRF}

\caption[Evaluation of compounds in translation output]%
{ Precision, Recall and F-score for compounds in the translation output,
  relative to the reference translations containing 2652 compounds in total.
  Standard deviations across the three MT repetitions are given next to the
  measurements.
  }
\label{tbl:hpylm:result_trans_acc}
\end{table}

\subsubsection{Qualitative analysis of translated compounds}
\label{sec:hpylm:example_translations}
The translation output was further analysed to find specific examples of
precision improvements attained by the HPYLM+c, compared to the HPYLM baseline.
Three examples are given in Table~\ref{tbl:hpylm:example_translations}
on p.~\pageref{tbl:hpylm:example_translations}.

In the Example~1, the baseline system only partially produced the translation of
the input phrase \iteg{prostate cancer}, while the HPYLM+c system succeeded in
producing the full compound.

Example~2 illustrates an improved lexical choice associated with the HPYLM+c
system: while \iteg{agenda} is a valid translation of the orthographically
equivalent English input word, the more idiomatic translation
\iteg{tagesordnung}
is produced by the HPYLM+c system, thereby matching the reference.

Finally, in Example~3 the baseline system incorrectly posits the compound
\iteg{selbst\-mord\-attentate} \iten{suicide attacks}, while the HPYLM+c system correctly
produces the less specific compound. 

In the first two examples, the sentential context plays to the strength of the
main assumption of the HPYLM+c (\ie ``context generates head''), offering a
potential explanation for the improved output.
However, in the final example, this is not the case; the compound's context in
the hypotheses is not particularly informative, and the improvement has to be
put down to a more complex discriminating role fulfilled by the language model
during decoding.

\begin{table}[tb]
\input{data/ML1_moreright_than_ML4.tex}
\caption[Translation examples]%
{Translation examples where compounds in the hypotheses are more correct
  with HPYLM+c than HPYLM as language model.  Roman numerals mark the MT
  repetitions, which are shown individually only when variation occurred around
  the word of interest.}
\label{tbl:hpylm:example_translations}
\end{table}

\section{Related work}
\label{sec:hpylm:related}
Another common approach for addressing the sparsity effects of productive compounding
\cite{Koehn2003a,Koehn2008,Stymne2009,Berton}, and rich morphology
\cite{Habash2006,Geutner1995}, has been to use pre/post-processing with an
otherwise unmodified translation system or speech recognition system.  This
approach views the existing machinery as adequate and shifts the focus to
finding a more appropriate segmentation of words into tokens, i.e. compounds
into parts or words into morphemes, thus achieving a vocabulary reduction.  The
downside of such a method is that training a standard \ngram language model on
pre-segmented data introduces unwanted effects: in the case of German compounds,
the split-off modifiers would take precedence in the head's \ngram
context, and during back-off the actual word-context information is discarded
first. As illustrated earlier in the ``forecasters predicted'' example in \S\ref{sec:bg:subwordLMs}, the problem is similar when modelling sequences of morphemes as \ngrams,
and earlier work in speech recognition has shown that taking steps against this
effect can improve recognition accuracy \cite{Ircing2001a}.

Pre-processing also often requires heuristics to guard against
over/under-seg\-men\-ta\-tion, which do not generalise well to different settings or
languages.  The modelling approach introduced in this chapter is also subject to
the whims of our compound segmentation method, but the model is more robust
since it retains the original surface form of the word---recall that the
decomposition step amounts to interpolated back-off.

\citet{Baroni2002a} proposed basic models of German compounds for use in
predictive text input, exploiting the same link between right-headedness and
context as we have, although their focus was restricted to compounds with two
components.

\label{correction:pyp-related-work}
Considering language modelling with PYPs more generally, two pieces of work in
particular warrant discussion. \citet{Wood2009} generalised hierarchical PYPs
beyond tree hierarchies to directed acyclic graph structures. They modelled
domain adaptation by tying together multiple domain-specific HPYLMs and a
latent ``background'' HPYLM. The intuition for modelling a given \ngram is that
a domain-specific model is able to back-off either by shortening the context or
by switching to the background model while retaining the same context.
Importantly, this choice is available at all context lengths and
is executed by using a base distribution that is a mixture of PYPs.
In contrast, our HPYLM+c ties together a head-generating HPYLM ($G_\ast$) and a
modifier-generating HPYLM ($F_\ast$) only at the level of full contexts, and does so
via a base distribution (of $H$) that is product instead of a mixture.
It could therefore be interesting to apply the graphical LM of \citet{Wood2009}
to the compounding problem, as it would allow back-off through context-dropping
and back-off through modifier-dropping at different context-lengths.

Finally, the PYP has recently also been used to devise language models that can
leverage the output of traditional finite-state morphological analysers.
\citet{Chahuneau2013}
define an open-vocabulary PYP-based word-model that generates words from
morphemes. This word-model is then integrated into an HPYLM by replacing the
uniform base distribution $G_0$, and performs well in their evaluation.
Though related in the sense of also modelling sub-word elements, the HPYLM+c
contrasts to their approach by seeking to model cross-word dependencies.

\section{Summary}
\label{sec:hpylm:summary}
In this chapter we formulated a Bayesian \ngram language model capable of
scoring compound words in terms of their components, which mitigates the data
sparsity caused by productive compounding.
We exploited the linguistic notion that the compound head is the most relevant
part for syntactic coherence within a sentence. Our results on German
indicate that this is a useful property to take into account, as the perplexity
degraded when setting up the model to ignore the right-headedness of German.
In an extrinsic evaluation in a machine translation system, the proposed
language model contributed to outputting compound words with higher precision.

%% file: ch-hpylm/algs.tex
\begin{algorithm}
  \caption{Adding an \ngram $(\mb{u}, w)$ to the HPYLM+c.}
\label{algfloat:add}
  \begin{footnotesize}
  \begin{algorithmic}[1]
  \algnotext{EndIf} 
  \algnotext{EndFor} 
  \Function{AddWordToModel}{$w$, $\mb{u}$}
  \State $\mb{c} \gets \textsc{SegmentCompound}(w)$ \Comment{for example, as detailled in \S\ref{sec:hpylm:segmentation-method}.}
  \State $P_0^w \gets G_{\mb{u}}(c_{\Lambda_1})
                        \times \prod\limits_{i=2}^{L} F_{c_{\Lambda_{i-1}}}(c_{\Lambda_i})
                        \times F_{c_{\Lambda_{L}}}(\$)$
                        \Comment{base probability, cf.\ \autoref{eq:hpylm:draw-H-from-PYP}}
  \If {\textsc{AddCustomerToRestaurant}($w$, $P_0^w$, $H_\mb{u}$, $a''_{|\mb{u}|}$, $b''_{|\mb{u}|}$)}
  \State \textsc{RecursiveAdd}($c_{\Lambda_1}$, $\mb{u}$, $G\lstar$, $\mb{a}$, $\mb{b}$)
    \For{$i \gets 2,L$}
      \State \textsc{RecursiveAdd}($c_{\Lambda_i}$, $c_{\Lambda_{i-1}}$, $F\lstar$ ,$\mb{a}'$, $\mb{b}'$)
    \EndFor
    \State \textsc{RecursiveAdd}(\$, $c_{\Lambda_L}$, $F\lstar$, $\mb{a}'$, $\mb{b}'$)
    
  \EndIf
  \EndFunction
  \\

  \Function{AddCustomerToRestaurant}{$w$, $P_0^w$, $X$, $a$, $b$} \\
  \Comment{$\HHist{w}$ is the table occupancy histogram for customer type $w$ for the restaurant $X$.}
    \State isNewTable $\gets$ false
    \State $p_\mathit{share} =  \max(N_w - a \times m_w$, 0) \Comment{based on \dots}
    \State $p_\mathit{new} = (a \times m+b)\times P_0^w$ \Comment{\dots
                        \autoref{eq:bg:seatconditional}\&\ref{eq:bg:predictive-conditional}}
    \State $r \gets random(0, p_\mathit{share} + p_\mathit{new})$
    \If{$r \leq p_\mathit{new}$ }
      \State $\Hist{w}{1} \gets \Hist{w}{1} + 1$
      \State isNewTable $\gets$ true
    \Else 
      \State $r \gets random(0, p_\mathit{share})$
      \ForAll {$t \in \HHist{w}$}\Comment{sample a table occupancy from histogram}
        \State $r \gets r - (\Hist{w}{t}\times(t-a))$
        \If {$r \leq 0$} \Comment{add to a table currently containing $t$ customers}
          \State $\Hist{w}{t+1} \gets \Hist{w}{t+1} + 1$
          \State $\Hist{w}{t} \gets \Hist{w}{t} - 1$
          \State \textbf{break}
        \EndIf
      \EndFor
    \EndIf
    \State $N_w \gets N_w + 1$
    \State \Return isNewTable
  \EndFunction
  \\

  \Function{RecursiveAdd}{$w$, $\mb{u}$, $X\lstar$, $\mb{a}$, $\mb{b}$}
    \If{$\mb{u}=\emptyset$ }
      \Return 
    \EndIf
    \State $P_w^0 \gets X_{\pi(\mb{u})}(w)$
    \If {\textsc{AddCustomerToRestaurant}($w$, $P_0^w$, $X_{\mb{u}}$,
                          $\mb{a}_{(|\mb{u}|)}$, $\mb{b}_{(|\mb{u}|)}$)}
      \State {\textsc{RecursiveAdd}($w$, $\pi(\mb{u})$, $X\lstar$, $\mb{a}$, $\mb{b}$)}
    \EndIf
  \EndFunction

\end{algorithmic}
\end{footnotesize}
\end{algorithm}

\begin{algorithm}
\caption{Removing an \ngram $(\mb{u}, w)$ from the HPYLM+c.}
\label{algfloat:remove}
\begin{footnotesize}
\begin{algorithmic}[1]
  \algnotext{EndIf} 
  \algnotext{EndFor} 
  \Function{RemoveWordFromModel}{$w$, $\mb{u}$}
  \State $\mb{c} \gets \text{segment($w$)}$
  \If {\textsc{RemoveCustomerFromRestaurant}($w$, $H_\mb{u}$)}
    \State \textsc{RecursiveRemove}($c_{\Lambda_1}$, $\mb{u}$, $G\lstar$)
    \For{$i \gets 2,L$}
      \State \textsc{RecursiveRemove}($c_{\Lambda_i}$, $c_{\Lambda_{i-1}}$, $F\lstar$)
    \EndFor
    \State \textsc{RecursiveRemove}(\$, $c_{\Lambda_L}$, $F\lstar$)
  \EndIf
  \EndFunction
  \\

  \Function{RemoveCustomerFromRestaurant}{$w$, $X$}\\
  \Comment{$\HHist{w}$ is the table occupancy histogram for customer type $w$ for the restaurant $X$.}
  \State isTableEmptied $\gets$ false
    \State $r \gets random(0, N_w)$
    \ForAll {$t \in \HHist{w}$} \Comment{sample a table occupancy from the histogram}
      \State $r \gets r - \Hist{w}{t}\times t$
      \If {$r \leq 0$} \Comment{remove from a table currently containing $t$ customers}
        \State $\Hist{w}{t} \gets \Hist{w}{t} - 1$
        \If{$t>1$}
          \State $\Hist{w}{t-1} \gets \Hist{w}{t-1} + 1$
        \Else
          \State isTableEmptied $\gets$ true
        \EndIf
        \State \textbf{break}
      \EndIf
    \EndFor
    \State $N_w \gets N_w - 1$
    \State \Return isTableEmptied
  \EndFunction
  \\

  \Function{RecursiveRemove}{$w$, $\mb{u}$, $X\lstar$}
    \If{$\mb{u}=\emptyset$ }
      \Return 
    \EndIf
    \If {\textsc{RemoveCustomerFromRestaurant}($w$, $X_{\mb{u}}$)}
    \State {\textsc{RecursiveRemove}($w$, $\pi(\mb{u})$, $X\lstar$)}
    \EndIf
  \EndFunction
\end{algorithmic}
\end{footnotesize}
\end{algorithm}

%% file: ch-hpylm/data/main_ppl_B.tex
\pgfplotstabletypeset[col sep=comma,
  every head row/.style={before row=\toprule,after row=\midrule},
  every last row/.style={after row=\bottomrule},
  every row no 1/.style={after row=\midrule},
  numCol/.style={numeric type, precision=1, fixed, fixed,zerofill},
  percColumn/.style={numCol, showpos,addPercToNonEmpty},
  display columns/0/.style={string type, column type=l, column name={Model} },%
  columns/ppl/.style={column name=PPL, numCol},%
  columns/RelVsMKN/.style={column name={vs.\ MKN},percColumn },%
  columns/RelVsHPYLM/.style={column name={vs.\ HPYLM},percColumn},%
  multicolumn names  
]{ 
,             ppl,    RelVsMKN, RelVsHPYLM
MKN,          299.86, ,
HPYLM,        294.01, -1.95,
HPYLM+c ling, 294.05, -1.94,    0.02
HPYLM+c inv,  305.46, 1.87,     3.89
}

%% file: ch-hpylm/data/ppl_vary_order.tex
\begin{tabular}{l rrr} \toprule
&\multicolumn{1}{c}{n=2} & \multicolumn{1}{c}{n=3} & \multicolumn{1}{c}{n=4} \\
\midrule
MKN                & 394.5 & 307.2 & 299.9 \\
HPYLM              & 396.6 & 303.3 & 294.0 \\
HPYLM+c            & 390.0 & 299.3 & 294.1 \\
\midrule
HPYLM relative to MKN      & $+0.5$\% & $-2.6$\% & $-1.9$\% \\
HPYLM+c relative to HPYLM  & $-1.6$\% & $-1.3$\% & $+0.0$\% \\
\bottomrule    
\end{tabular}

%% file: ch-hpylm/data/compound-PRF.tex
\pgfplotstabletypeset[col sep=comma,
  every head row/.style={before row=\toprule,after row=\midrule},
  every last row/.style={after row=\bottomrule},
  every row no 1/.style={after row=\midrule},
  numCol/.style={numeric type, precision=1, fixed, fixed,zerofill},
  every odd column/.style={numCol}, 
  every even column/.style={numCol, 
    column name ={},
    column type/.add={@{}>{\footnotesize \ensuremath \pm}}{} 
    },
  display columns/0/.style={string type, column type=l, column name={} },%
  multicolumn names  
]{ 
model,             P,        Psd,   R,        Rsd,   F,        Fsd
MKN,               25.49319, 1.461, 17.21971, 0.986, 20.52788, 0.765
HPYLM,             24.49132, 2.400, 17.54651, 0.737, 20.38019, 0.377
HPYLM+c ($\Lambda^{\text{ling}}$) ,      27.66231, 2.264, 17.40824, 0.660, 21.32176, 0.370
HPYLM+c ($\Lambda^{\text{inv}}$),       23.97813, 2.612, 17.18200, 0.400, 19.96250, 0.645
}

%% file: ch-hpylm/data/ML1_moreright_than_ML4.tex
\begin{footnotesize}%
\noindent%
\fbox{ \parbox{\textwidth}{
\begin{tabular}{lp{0.75\textwidth}}
  \multicolumn{2}{l}{\bf Example 1} \\
  Src: &
 \textcolor{black!70}{in the period 1994-2009 she along with her colleagues
monitored more than 1,500 patients with} 
\textbf{\mbox{prostate cancer}} %
\textcolor{black!70}{in britain and in parallel over three thousand healthy
men~.} \\
Ref: &
 \textcolor{black!70}{im zeitraum von 1994-2009 verfolgte sie mit ihren
kollegen in britannien mehr als 1500 patienten mit} 
\textbf{\mbox{prostatakrebs}}%
\textcolor{black!70}{~, gleichzeitig aber auch 3000 gesunde männer~.}\\
HPYLM (i--ii): &
 \textcolor{black!40}{im zeitraum 1994-2009 sie zusammen mit ihren kollegen
überwacht mehr als 1.500 patienten mit} 
\textbf{\mbox{prostata-}} %
\textcolor{black!40}{in großbritannien und parallel über 3~000 gesunde
männer~.}\\
HPYLM (iii): &
 \textcolor{black!40}{in den jahren 1994-2009 sie zusammen mit ihren kollegen
überwacht mehr als 1.500 patienten mit} 
\textbf{\mbox{prostata- krebs}} %
\textcolor{black!40}{in großbritannien und gleichzeitig über 3~000 gesunde
männer~.}\\
HPYLM+c (i--iii): &
 \textcolor{black!40}{in den jahren 1994-2009 sie zusammen mit ihren kollegen
überwacht mehr als 1.500 patienten mit} 
\textbf{\mbox{prostatakrebs}} %
\textcolor{black!40}{in großbritannien und parallel über~3 000 gesunde
männer~.}\\
\end{tabular}
}} 

\noindent%
\fbox{ \parbox{\textwidth}{
\begin{tabular}{lp{0.75\textwidth}}
  \multicolumn{2}{l}{\bf Example 2} \\
Src: &
 \textcolor{black!70}{co-chairs alan simpson and erskine bowles have taken on
one of the most difficult jobs in washington and have~, in their own
estimation~, forced the issue of the deficit onto the public} 
\textbf{\mbox{agenda}} %
\textcolor{black!70}{in ways that can~'t be ignored~.}\\
Ref: &
 \textcolor{black!70}{die stellvertretenden vorsitzenden alan simpson und
erskine bowles haben einen der schwierigsten jobs in washington übernommen und
haben~, gemäß ihrer eigenen einschätzung~, so sehr darauf gedrängt~, die
defizitfrage auf die öffentliche} 
\textbf{\mbox{tagesordnung}} %
\textcolor{black!70}{zu setzen~, dass diese nicht mehr ignoriert werden
können~.}\\
HPYLM (i--iii): &
 \textcolor{black!40}{ko-vorsitzenden alan simpson und erskine bowles auf sich
genommen haben eines der schwierigsten jobs in washington und haben~, in ihrer
eigenen einschätzung~, die frage des defizits auf der öffentlichen} 
\textbf{\mbox{agenda}} %
\textcolor{black!40}{in einer weise~, die nicht ignoriert werden kann~.}\\
HPYLM+c (i--iii): &
 \textcolor{black!40}{ko-vorsitzenden alan simpson und erskine bowles haben
eines der schwierigsten jobs in washington und haben~, in ihrer eigenen
einschätzung , die frage des defizits in einer art und weise auf die
öffentliche}
\textbf{\mbox{tagesordnung}}~%
\textcolor{black!40}{, die nicht ignoriert werden kann~.}\\
\end{tabular}
}} 

\noindent%
\fbox{ \parbox{\textwidth}{
\begin{tabular}{lp{0.75\textwidth}}
  \multicolumn{2}{l}{\bf Example 3} \\
Src: &
 \textcolor{black!70}{the majority of members of the committee concluded that
the message conveyed was that} 
\textbf{\mbox{suicide}} %
\textcolor{black!70}{should not be seen as a solution of life 's problems~.}\\
Ref: &
 \textcolor{black!70}{die mehrheit der aufsichtsratsmitglieder hatte die
schlussfolgerung gezogen~, die vermittelte botschaft sei hierbei~, dass} 
\textbf{\mbox{selbstmord}} %
\textcolor{black!70}{nicht als lösung von lebenskrisen angesehen werden
dürfe~.}\\
HPYLM (i--iii): &
 \textcolor{black!40}{die mehrheit der mitglieder des ausschusses zu dem
schluss~, dass die botschaft übermittelt wurde~, dass}
\textbf{\mbox{selbstmordattentate}} %
\textcolor{black!40}{sollte nicht als eine lösung der probleme des lebens
betrachtet werden~.}\\
HPYLM+c (i--iii): &
 \textcolor{black!40}{die mehrheit der mitglieder des ausschusses zu dem
schluss~, dass die botschaft übermittelt wurde~,} 
\textbf{\mbox{selbstmord}} %
\textcolor{black!40}{sollte nicht als eine lösung der probleme des lebens
betrachtet werden~.}\\
\end{tabular}
}} 
\end{footnotesize}%

%% file: ch-dist/ch-dist.tex
\chapter{\label{ch:dist}Rich morphology in distributed language models}

\chapoverview{%
In this chapter we develop a distributed language model that composes morpheme
vectors into word vectors so that morphologically related words are linked in the model.
We show that this leads to substantial improvements in perplexity and that the
acquired morpheme vectors successfully enable the construction of
representations for unknown words. A further contribution is to demonstrate the
full integration of a distributed language model into a machine translation
decoder, which leads to improvements in the translation quality metric.
Material in this chapter was originally published as \citep{Botha2014}.
}

In the previous chapter we leveraged existing morphological information in the
form of compound segmentations to improve the smoothing of \ngram distributions.
While this lead to some performance improvements, our results highlighted the
challenge in devising suitable back-off schemes that correctly prioritise the
different types of generalisation (\eg compound decomposition versus context
truncation).

This provides part of the motivation for shifting the focus to distributed
language models (DLMs), which circumvent the need for carefully constructing
back-off schemes by modelling arbitrary correlations between the feature
representations of context words and their subsequent target word.

Word-based distributed representations have been found to capture some
morphological regularity automatically \cite{Mikolov2013a}, but we contend that
there is a case for building \emph{a~priori} morphological awareness into the
language models' inductive bias.
Conversely, compositional vector-space modelling has recently been applied to
morphology to good effect \cite{Lazaridou2013,Luong2013}, but lacked the
probabilistic \mbox{basis} necessary for use with a machine translation
decoder.

The method we propose strikes a balance between probabilistic language
modelling and morphology-based representation learning.  Word vectors are
composed as a linear function of arbitrary sub-elements of the word, 
\eg surface form, stem, affixes, or other latent information.  The effect is to tie
together the representations of morphologically related words, directly
combating data sparsity.  This is executed in the context of a log-bilinear
(LBL) LM \cite{Mnih2007}, which is sped up sufficiently by the use of word
classing so that we can integrate the model into an open source machine
translation decoder and evaluate
its impact on translation into 6 languages, including the morphologically
complex Czech, German and Russian.

\section{Additive word representations}
\label{sec:additive-wordreps-idea}
As introduced in \autoref{ch:bg}, a DLM associates with each word type $v$ in the vocabulary $\V$ a
$d$-dimensional feature vector $\mathbf{r}_v \in \mathbb{R}^d$.  Regularities
among words are captured in an opaque way through the interaction of these
feature values and a set of transformation weights.  This leverages linguistic
intuitions only in an extremely rudimentary way, in contrast to hand-engineered
linguistic features that target very specific phenomena, as often used in
supervised-learning settings.

We seek a compromise that retains the unsupervised nature of DLM
feature vectors, but also incorporates pre-existing morphological information
into the model in a flexible and efficient manner.  In particular, morphologically related
words should share statistical strength in spite of differences in word form.

As a way of integrating additional information about a word type, we define a mapping $\mu:\V \mapsto \F^+$ of a surface word into a
variable-length sequence of \emph{factors}, \ie $\mu(v) = (f_1, \dots, f_{K(v)})$,
where $v\in \V, f_i \in \F$ and $1\leq i \leq {K(v)}$.  A factor $f$, which could be
a word's stem or suffix, or some other relevant sub-word element, has an
associated \emph{factor feature vector} $\mathbf{r}_f \in \mathbb{R}^d$. 
Under this view, the vector representation
$\tilde{\mb{r}}_v$ of a word $v$ is reformulated as a function $\omega_\mu(v)$ of its factor
vectors, adding them together: %
\begin{equation} %
\tilde{\mb{r}}_v =\omega_\mu(v)= \sum_{f \in \mu(v)} \mathbf{r}_f 
\end{equation} %
The vectors of morphologically related words become explicitly linked through
shared factor vectors, for example
\begin{center}%
\begin{tabular}{rcl}%
$\overrightarrow{\mathrm{imperfection}}$ 
&$=$& $\overrightarrow{\mathit{im}} + \overrightarrow{\mathit{perfect}} + \overrightarrow{\mathit{ion}}$ \\
$\overrightarrow{\mathrm{perfectly}}$ 
&$=$& $\overrightarrow{\mathit{perfect}} + \overrightarrow{\mathit{ly}}$,
\end{tabular} %
\end{center} %
where we use the notation $\overrightarrow{\mathrm{word}}$ and $\overrightarrow{\mathit{factor}}$.

A valuable by-product of this approach is that it gives a well-defined method
for constructing representations for out-of-vocabulary (OOV) words using the
available morpheme vectors.

In practice, we set up the factorisation $\mu$ to also include the surface form of a word as a
factor.
This is done to account for non-compositional constructions such as
\[
\overrightarrow{\mathrm{greenhouse}} \neq
\overrightarrow{\mathit{green\vphantom{k}}} + \overrightarrow{\mathit{house}}
\]
and also makes the approach more
robust to noisy morphological segmentations. 
The word representation for the preceding example is therefore 
\[
\overrightarrow{\mathrm{greenhouse}} = \overrightarrow{greenhouse} +
\overrightarrow{\mathit{green\vphantom{k}}} + \overrightarrow{\mathit{house}}.
\]

An obvious shortcoming in the choice of addition as composition function $\omega$ is
its commutativity. Inclusion of the surface factor also overcomes the associated
invariance in the order of morpheme factors, with the desired outcome that
 $\overrightarrow{\mathrm{houseboat}} \neq \overrightarrow{\mathrm{boathouse}}$, for example.

The number of factors per word is free to vary across the vocabulary, making
the approach applicable across the spectrum of more fusional languages (\eg 
Czech, Russian) to more agglutinative languages (\eg Turkish).  This is in
contrast to factored language models \cite{Alexandrescu2006}, which
assume a fixed number of factors per word.  Their method of concatenating
factor vectors to obtain a single representation vector for a word can be seen
as enforcing a partition on the feature space.  Our method of addition avoids
such a partitioning and better reflects the absence of a strong intuition about
what an appropriate partitioning might be.  A limitation of our method compared
to theirs is that the deterministic mapping $\mu$ currently enforces a single
factorisation per word type, which sacrifices information obtainable from
context-disambiguated morphological analyses.

The additive composition function $\omega$ can be regarded as an 
instantiation of the weighted addition strategy \emph{wadd} that performed well
in the distributional semantics approach to derivational morphology reported by
\citet{Lazaridou2013}.
Their weighted addition involves using a global pair of weights $\alpha$ and
$\beta$ to compose a $\overrightarrow{\mathrm{word}}$ as
$\alpha \times \overrightarrow{\mathit{stem\vphantom{k}}} 
+ \beta \times \overrightarrow{\mathit{suffix\vphantom{k}}}$. (Their focus is
limited to the segmentation of a word into a stem and suffix, and evaluated only
on English.) 
Importantly, they fit the two weights while keeping the vector representations
fixed. As described shortly, we situate our composition method within a distributed language model.
Our feature representations can therefore adjust jointly with the effect of the
composition. To a first approximation, individual feature values can thereby absorb a
 weighting component specific to a morpheme type or a feature dimension, rather
 than using a single pair of global weights.

Another aspect of our composition method is how it models derivational ambiguity.
Unlike the recursive neural-network method of \citet{Luong2013}, we do not
impose a single tree structure over a word, which ignores the semantic ambiguity
inherent in words like un[[lock]able] vs.\ [un[lock]]able.\footnote{Example due
  to \citet{deAlmeidaUnXable}.}
The consideration is that a model should either account for such ambiguity
directly, or operate in a way that treats alternative derivations uniformly so
that it makes minimal assumptions about the language in question. Our approach is
to do the latter.

In contrast to these two recent approaches to vector-based morphological modelling, our
additive representations are readily implementable in a probabilistic language
model suitable for use in a decoder.\footnote{Our focus is on surface
  morphology, but the method could equally be used to integrate other
  information, \eg PoS, lemma.
}

\section{Log-bilinear language models}
\label{sec:dist:LBLbackground}
In this section we formalise the distributed language model into which we
subsequently integrate our compositional approach to morphology.

Log-bilinear (LBL) models \cite{Mnih2007} are an instance of DLMs that make
the same Markov assumption as \ngram language models, as introduced in
\autoref{ch:bg}.
The conditional distribution $P(w_i \mid w_{i-n+1}^{i-1})$ of a word given its
context is modelled by a smooth scoring function $\nu(\cdot)$ over vector representations of words. 

The LBL predicts the vector $\mb{p}$ for the next word as a function of the
context vectors $\mb{q}_j \in \mathbb{R}^d$ of the preceding \mbox{$n$--1}
 words, %
\begin{equation} 
\mb{p} = \bm{\eta}\left(\sum_{j=1}^{n-1}  \mb{q}_j C_j\right),
\label{eq:dist:LBL-predict-p}
\end{equation} 
where $C_j \in \mathbb{R}^{d\times d}$ are position-specific weight matrices,
and $\bm{\eta}$ is the identity function \mbox{$\bm{\eta}(\mb{x})=\mb{x}$},
written this way for a subsequent variation.

The continuous scoring function $\nu(w)$ measures how well a word $w$ fits the predicted
representation $\mb{p}$ and is
defined as 
\begin{equation}
  \nu(w)=\mb{p} \cdot \mb{r}_w + b_w, \label{eq:dist:lbl-nu}
\end{equation}
where 
$\mb{r}_w \in \mathbb{R}^d$ is the representation vector of the target word $w$
and
$b_w$ is a bias term
encoding the prior probability of the word type.
The conditional probability of the word is calculated through exponentiation and
normalisation of the scoring function:
\begin{equation} %
P\left(w_i | w_{i-n+1}^{i-1}\right) = 
\frac{\exp\left(  \nu(w_i) \right) }
{\sum_{v \in \V} \exp\left( \nu(v)\right)}. \label{eq:dist:softmax}
\end{equation} %
This model is subsequently denoted as {\bf \lbl} with parameters
\mbox{$\Theta_{\lbl}=\left(C_j,Q,R,\mb{b} \right)$}, 
where $Q, R \in \mathbb{R}^{|\V|\times d}$ contain the word representation
vectors as rows, and \mbox{$\mb{b} \in \mathbb{R}^{|\V|}$}.  $Q$ and $R$ imply
that separate representations are used for conditioning and output.
See \autoref{fig:dist:modelviz-lbl} for an illustrative diagram.

\section{Variants of the log-bilinear model}
\subsection{Additive log-bilinear model}
\label{sec:additive-lbl}
\begin{figure}[tb]
  \centering
  \begin{minipage}[b][5cm][t]{0.4\textwidth}
    \centering
    \includegraphics[height=3.8cm]{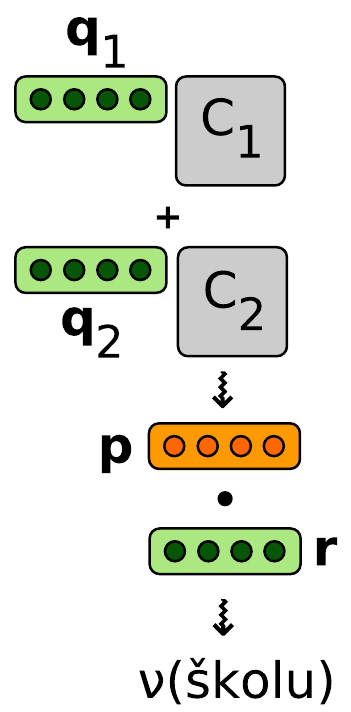}
    \subcaption{\lbl}
    \label{fig:dist:modelviz-lbl}
  \end{minipage}
  \hspace{0.05\textwidth}
  \begin{minipage}[b][5cm][t]{0.4\textwidth}
    \centering
    \includegraphics[height=3.8cm]{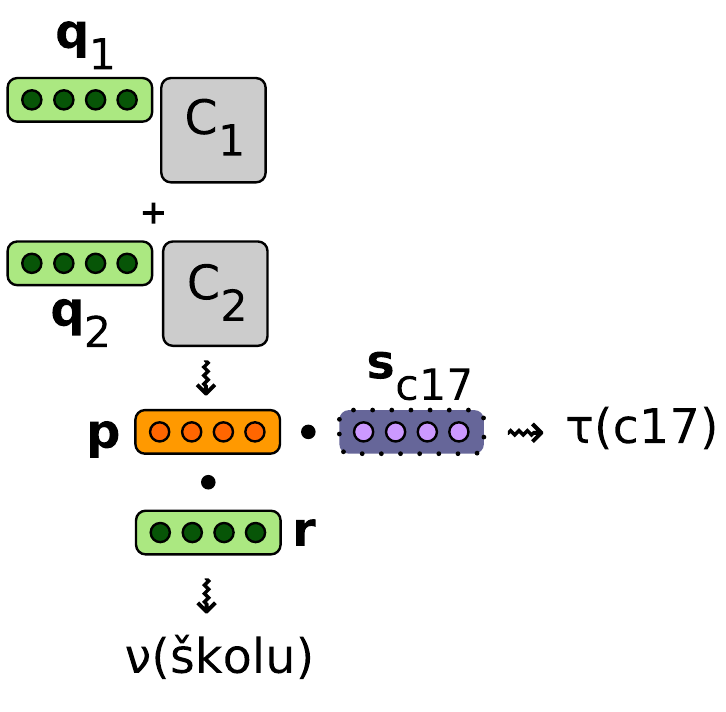}
    \subcaption{\clbl}
    \label{fig:dist:modelviz-clbl}
  \end{minipage}

  \vspace{2ex}
  \begin{minipage}[b][5cm][t]{0.35\textwidth}
    \includegraphics[height=3.8cm]{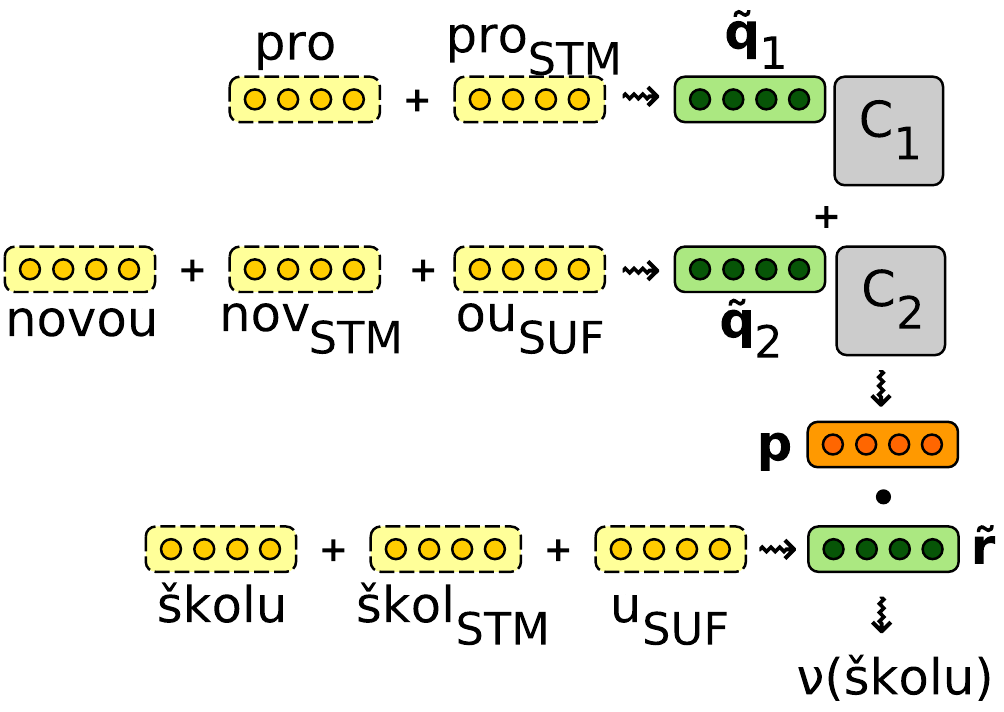}
    \subcaption{\lblpp}
    \label{fig:dist:modelviz-lblpp}
  \end{minipage}
  \hspace{0.05\textwidth}
  \begin{minipage}[b][5cm][t]{0.55\textwidth}
    \includegraphics[height=3.8cm]{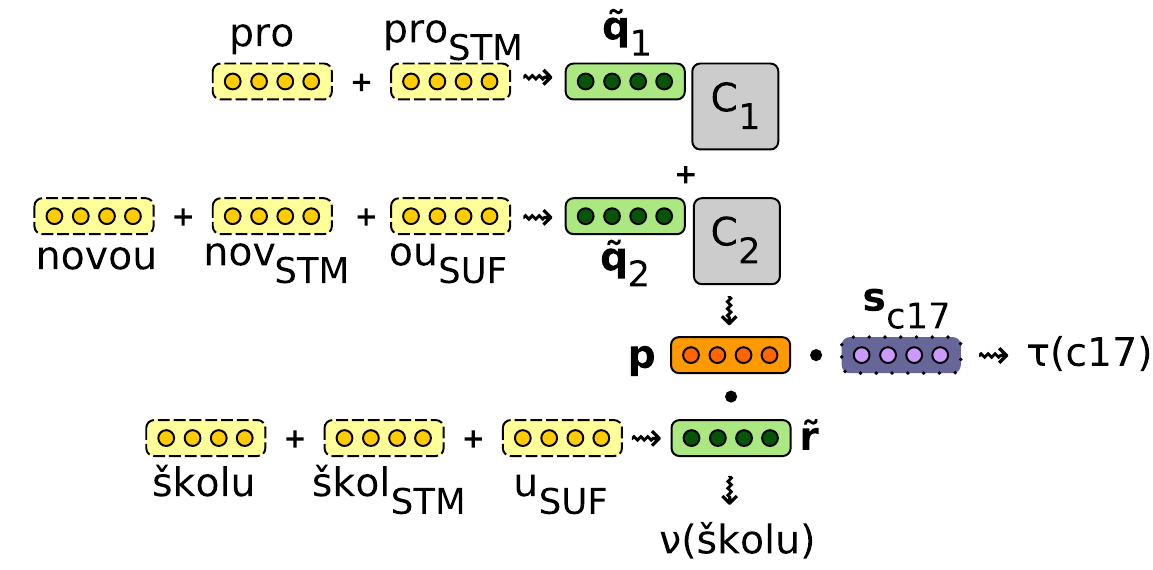}
    \subcaption{\clblpp}
    \label{fig:dist:modelviz-clblpp}
  \end{minipage}
  \caption[Model diagrams for LBL, LBL++ and CLBL++]%
{ Illustration of how three different \mbox{\kgram{3}}
      model variants treat the Czech phrase \iteg{pro novou školu} \iten{for [the] new
        school}, where the preposition causes accusative case markings on the
      adjective and noun.
      In (b) and (d), we assume the vocabulary clustering (\S\ref{sec:class-based}) assigned class~17 to the target word.
    }
    \label{fig:instance}
\end{figure}

We introduce a variant of the LBL that makes use of additive representations
(\S\ref{sec:additive-wordreps-idea}) by associating the \emph{composed word
vectors} $\tilde{\mb{r}}$ and $\tilde{\mb{q}}_j$ with the target and context
words, respectively.
We redefine the representation matrices to be
$Q^{(f)}, R^{(f)}\in \mathbb{R}^{|\F|\times d}$ 
thus containing a vector for each factor type in the global vocabulary of
factors $\F$.
This model is designated {\bf \lblpp} and has parameters
\mbox{$\Theta_{\lblpp}=\left(C_j,Q^{(f)}, R^{(f)},\mb{b} \right)$}. 
See the example in \autoref{fig:dist:modelviz-lblpp}.

Words sharing factors are tied together, which is expected to improve
performance on rare word forms.
Note that the composed word vectors are not model parameters in their own right, but are
strictly a function of the factor vectors according to the composition function
$\omega$ and mapping $\mu$. The scoring function for this model is an expansion
of Equations~\ref{eq:dist:LBL-predict-p}--\ref{eq:dist:lbl-nu}
to include the additive representations:
\begin{align}
\nu_{\lblpp}(w)
  &= \bm{\eta}\Biggl(\sum_{j=1}^{n-1} \tilde{\mb{q}}_j C_j \Biggr)
    \cdot \tilde{\mb{r}}_w + b_w \\
  &= \bm{\eta}\Biggl(\sum_{j=1}^{n-1} \,%
    \Bigl(\!\!\sum_{f \in \mu(w_j)} \mb{q}_f \Bigr) C_j  \Biggr)
    \cdot \Bigl(\!\!\sum_{f \in \mu(w)} \mb{r}_f \Bigr) + b_w.
\end{align}
Similar to \autoref{eq:dist:softmax}, the probability model is
\begin{equation} %
P\left(w_i | w_{i-n+1}^{i-1}\right) = 
\frac{\exp\left(  \nu_{\lblpp}(w_i) \right) }
{\sum_{v \in \V} \exp\left( \nu_{\lblpp}(v)\right)}. \label{eq:dist:softmax-lblpp}
\end{equation} %

Representing the mapping $\mu$ with a sparse transformation matrix $M\in
\mathbb{Z}_{+}^{\V \times |\F|}$ establishes the relationship between composed word and factor
representation matrices as $R = MR^{(f)}$ and $Q = MQ^{(f)}$. 
A row in $M$ selects the factor vectors for a given word type from
the $R^{(f)}$ (or $Q^{(f)}$) matrix. These row-vectors
are a generalisation of one-hot vectors: they have multiple non-zero entries,
corresponding to the multiple factors that make up a word,
and integer values greater than one would encode multiple occurrences of the
same factor within a given word.
In practice, we
exploit this correspondence between composed word representations and factor
representations for test-time efficiency---the composed word vectors are compiled offline so
that the computational cost of \lblpp probability lookups is equivalent to
that of \lbl.

We consider two obvious variations of the \lblpp to evaluate the extent to
which interactions between context and target factors affect the model:
{\bf \lblpo} only factorises {\bf o}utput words and retains simple word vectors
for the context (\ie $Q\equiv Q^{(f)}$), while {\bf \lblpc} does the reverse,
only factorising {\bf c}ontext words.\footnote{The +c, +o and ++ naming
suffixes denote these same distinctions when used with the \clbl model
introduced later.} Both reduce to the \lbl when setting $\mu$ to be the
identity function, such that $\V\equiv\F$.

The factorisation permits an approach to handling unknown context words that is less
harsh than the standard method of replacing them with a global unknown
symbol---instead, a vector can be constructed from the known factors of the word
(\eg the observed stem of an unobserved inflected form).  A similar scheme
can be used for scoring unknown target words, but requires changing the event
space of the probabilistic model.  We use this vocabulary stretching capability
in our word similarity experiments, but leave the extensions for test-time
language model predictions as future work.

A consequence of the additional free parameters introduced by the additive representations
is that a maximum-likelihood
training criterion would favour the degenerate solution where the non-surface
vectors in $\F \setminus \V$ are pushed to zero, rendering the additive
extension ineffective.   Including
a regularisation term in the training objective is crucial and should
yield non-trivial values for the parameters tied through our additive
representations. (Figure~\ref{fig:lambda} in the experimental section
demonstrates the sensitivity of performance to the regularisation strength.)

\subsection{Class-based model decomposition}
\label{sec:class-based}
DLMs are known to achieve much lower perplexities than conventional \ngram
models, but their application to decoder-based tasks like translation and speech
recognition have largely been limited to rescoring of lattices or n-best
lists of hypotheses. However, they could exercise a greater influence in those
tasks if integrated into the decoders.

The key obstacle to using DLMs in a decoder is the expensive normalisation
over the vocabulary (\autoref{eq:dist:softmax}).  Our approach to reducing the computational cost of
normalisation is to use a class-based decomposition of the probabilistic model
\cite{Goodman2001a,Mikolov2011}.  Using Brown-clustering
\cite{Brown1992},\footnote{In preliminary experiments, Brown clusters gave much
  better perplexities than frequency-binning \cite{Mikolov2011}.} we partition
the vocabulary into $|\Set{C}|$ classes, denoting as $\Set{C}_c$ the set of
vocabulary items in class $c$, such that $\V =
\Set{C}_1\cup\dots\cup\Set{C}_{|\Set{C}|}$. This is a hard clustering, which we
treat as fixed and observed for a given vocabulary and dataset.
The class $c(w)$ of a word $w \in \V$ can therefore be looked up. We
emphasise that we are working in a discriminative setting.

\label{correction:hard-cluster}
In this model, the probability of a word conditioned on the history $h$ of
$n-1$ preceding words\footnote{This is therefore not a class-based language
  model in the sense of predicting the next class conditioned on the preceding
  \emph{classes} \cite{Brown1992}.}
is decomposed as
\begin{align} %
P(w|h) &= \sum_{c' \in \Set{C}} P(c'|h) P(w|h,c') \\
       &= P(c(w)|h) P(w|h,c(w)) 
          +\sum_{c' \in \Set{C}, c'\neq c(w)} \left(P(c'|h) \cdot 0 \right) \\
       &= P(c(w)|h) P(w|h,c(w)).
\end{align} %
This class-based model, {\bf \clbl}, extends the \lbl by associating a
representation vector $\mb{s}_c$ and bias parameter $t_c$ with each class $c$,
such that
\mbox{$\Theta_{\clbl}=\left(C_j,Q,R,S,\mb{b},\mb{t} \right)$}.
The same prediction vector $\mb{p}$ is used to compute both class score
$\tau(c)=\mb{p}\cdot\mb{s}_c+t_c$ and word score $\nu(w)$, which are normalised
separately:
\begin{align} %
P(c | h) &= 
\frac{\exp\left(\tau(c)\right) }
{\sum\limits_{c'=1}^{|\Set{C}|} \exp\left(\tau(c')\right)}
\\[3pt]
P(w | h,c) &= 
\frac{\exp\left( \nu(w) \right) }
{\sum\limits_{v'\in \Set{C}_c} \exp\left(\nu(v')\right)}.
\end{align} %
\autoref{fig:dist:modelviz-clbl} provides an illustration.

We favour this flat vocabulary partitioning for its computational adequacy,
simplicity and robustness.  Computational adequacy is obtained by using
$|\Set{C}| \approx |\V|^{0.5}$, thereby reducing the $\BigO{|\V|}$
normalisation operation of the \lbl to two $\BigO{|\V|^{0.5}}$ operations in
the \clbl, as Brown clustering mostly yields balanced class sizes.

Other methods for achieving more drastic complexity reductions exist in the
form of frequency-based truncation, shortlists \cite{Schwenk2004}, or casting
the vocabulary as a full hierarchy \cite{Mnih2008} or partial hierarchy
\cite{Le2011}.  We expect these approaches could have adverse effects in the
rich morphology setting, where much of the vocabulary is in the long tail of
the word distribution.

\subsection{Non-linearity}
The use of non-linear activation functions in DLMs
are known to improve performance over strictly linear ones
\cite{Mnih2009,Pachitariu2013}, but comes at a cost. Typical sigmoidal and
$\tanh$ non-linearities shrink the gradients, which causes slower convergence
during training
than the linear models (\mytilde{}20 vs.\ \mytilde{}4 iterations in our
experience with \clbl, \S\ref{sec:class-based}).  Non-linearities can create
difficulties during training \cite{Pascanu2012} and incur a computational cost
at test-time that one would wish to avoid when using language models in a
decoder. Rectified linear units \cite{NairHinton2010rectlin} have been used as
a workaround \cite{Vaswani2013}, but a pertinent question is
whether a simple technique like our additive representations might serve as a
substitute for non-linearities altogether.\footnote{This consideration is
  limited to single layer networks---in multi-layer networks, non-linearities
  take on a defining role.} 

For comparison, we define a non-linear variant {\bf \nlbl} that uses a
sigmoidal activation function on the prediction vector,
\ie $\bm{\eta} \equiv \bm{\sigma}$ (cf.\ \autoref{eq:dist:LBL-predict-p}), 
which operates element-wise as \mbox{$\sigma(x) = \left(1+e^{-x}\right)^{-1}$}.
Our definition
here is simpler than the various options explored by \citet{Mnih2009} in that
it only applies to the final prediction vector.

\section{Experiments}
\label{sec:dist:experiments}
The overarching aim of our evaluation is to investigate the effect of using the
proposed additive representations across languages with a range of
morphological complexity.

Our intrinsic language model evaluation has two parts.  We first perform a
model selection experiment on small data to consider the relative merits of
using additive representations for context words, target words, or both, and to
validate the use of the class-based decomposition.

Then we consider class-based additive models trained on tens of millions of
tokens and large vocabularies.  These larger language models are applied in two
extrinsic tasks: i)~a word-similarity rating experiment on multiple languages,
aiming to gauge the quality of the induced word and morpheme representation
vectors; ii)~a machine translation experiment, where we are specifically
interested in testing the impact of an LBL LM feature when translating into
morphologically rich languages.

\subsection{Training and initialisation}
Model parameters $\Theta$ are estimated by optimising an L2-regularised log
likelihood objective.  Training the \clbl and its additive variants directly
against this objective is fast because normalisation of model scores, which is
required in computing gradients, is over a small number of events, namely
classes, and word types in a given class.

For the classless LBLs we use noise-contrastive estimation (NCE)
\cite{Gutmann2012,Mnih2012} to avoid normalisation during training. 
NCE achieves this by casting the problem as one of
discriminating between real training instances and synthetically generated
``noise'' instances. For each unigram in the data, we generated 100~noise instances from the empirical unigram distribution.
While NCE speeds up training, it leaves the expensive test-time normalisation
of LBLs unchanged, precluding their usage during decoding.

Bias terms $\mb{b}$ (resp.\ $\mb{t}$) are initialised to the log unigram
probabilities of words (resp.\ classes) in the training corpus, with Laplace
smoothing, while all other parameters are initialised randomly according to
sharp, zero-mean Gaussians.  Representations are thus learnt from scratch and
not based on publicly available embeddings, meaning our approach can easily be
applied to many languages.

Optimisation is performed by stochastic gradient descent with updates after
each mini-batch of $L$ training examples.  We apply AdaGrad \cite{Duchi2011},
which effectively adjusts the learning rate individually for each model
parameter as a function of a global step-size hyperparameter $\xi$, which we tune
on held-out development data. The number of training iterations $I$
is also treated as a hyperparameter, and training is halted once the
perplexity on the development
data starts to increase.\footnote{%
    $L$=10k--40k, $\xi$=0.05--0.08, dependent on $|\V|$ and data size;
    $I$=2--4 passes through the data for the linear models.
}%

\subsection{Data and methods}
\label{sec:methods}

\begin{table}[tb]%
    \centering
        \begin{tabular}{l*{2}{r}|*{2}{r}c} \toprule
            & \multicolumn{2}{c}{\textsc{Data-1m}}
            & \multicolumn{3}{c}{\textsc{Data-Main}} 
            \\
            & \multicolumn{1}{c}{Tokens}  & \multicolumn{1}{c}{$|\V|$}  
            & \multicolumn{1}{c}{Tokens}  & \multicolumn{1}{c}{$|\V|$} 
            & Sentence Pairs
            \\ \midrule
            \Lcs & 1m & 46k & 16.8m & 206k & 0.7m \\
            \Lde & 1m & 36k & 50.9m & 339k & 1.9m \\
            \Len & 1m & 17k & 19.5m & 60k &  0.7m \\
            \Les & 1m & 27k & 56.2m & 152k & 2.0m   \\
            \Lfr & 1m & 25k  & 57.4m & 137k & 2.0m   \\
            \Lru & 1m & 62k  & 25.1m & 497k & 1.5m \\ \bottomrule 
        \end{tabular}
    \caption[Statistics of monolingual and parallel training data]%
      {Corpus statistics. The number of sentence pairs for a
      row \data{X} refers to the English$\rightarrow$X parallel data (but row
      \Len has Czech as source language).}
    \label{tbl:corpus_stats}
\end{table}

We make use of data from the 2013 ACL Workshop on Machine
Translation.\footnote{http://www.statmt.org/wmt13/translation-task.html}\ %
We first describe data used for translation experiments, since the monolingual
datasets used for language model training were derived from that. The language
pairs are English$\rightarrow$\{German, French, Spanish, Russian\} and
English$\leftrightarrow$Czech.  Our parallel data comprised the
\corpus{Europarl-v7} and \corpus{news-commentary} corpora, except for
English--Russian where we used \corpus{news-commentary} and the \corpus{Yandex}
parallel corpus.\footnote{https://translate.yandex.ru/corpus?lang=en}\ %
Pre-processing involved lowercasing, tokenising and filtering to exclude
sentences of more than 80 tokens or substantially different lengths.

\kgram{4} language models were trained on the target data in two batches:
\textsc{Data-1m} consists of the first million tokens only, while
\textsc{Data-Main} uses the full target-side of the corpora listed above.  Statistics are given in
Table~\ref{tbl:corpus_stats}.  \corpus{newstest2011} was used as development
data\footnote{The \corpus{\scriptsize{newstest2011}} data set does not feature Russian, so
  for that language some training data was held out for tuning.} for tuning
language model hyperparameters, while intrinsic LM evaluation was done on
\corpus{newstest2012}, measuring perplexity. In addition to contrasting the LBL
variants, we also compare against the  MKN baseline.

\subsubsection{Language model vocabulary}
Additive representations that link morphologically related words specifically
aim to improve modelling of the long tail of the lexicon, so we do not want to
prune away all rare words, as is common practice in language modelling and word
embedding learning.  We define a \emph{singleton pruning rate} $\kappa$, and
randomly replace a fraction $\kappa$ of words occurring only once in the training
data with a global \unk symbol.
We use low pruning rates\footnote{%
    \textsc{Data-1m}: $\kappa=0.2$;
    \textsc{Data-Main}: $\kappa=0.05$
}\ %
and thus model large vocabularies.\footnote{ %
    We also mapped digits to~0, (\ie 15.1\%$\Rightarrow$00.0\%),
    and cleaned the Russian data by replacing tokens having \textless80\%
    Cyrillic characters 
    with \unk.
}

\subsubsection{Word factorisation $\mu$}
\label{sec:dist:wordfactorisation}
We obtain morphological segmentations from the \mbox{unsupervised}
segmentor \tool{Morfessor Cat-MAP} \cite{Creutz2007}, which takes as input a
list of word forms and their frequencies in a corpus. The output consists of the
segmentation points along with labels that identify each segment as a prefix,
stem or suffix. For example, the word \emph{unsegmented} is analysed as \emph{un\subpre segment\substem
  ed\subsuf}.

The mapping $\mu$ of a
word is taken as its surface form and the morphemes identified by Morfessor.
Keeping the morpheme labels 
allows the model to learn separate vectors for, say, in\substem
the preposition and in\subpre occurring as ``in-appropriate''.
By not post-processing segmentations in a more sophisticated way (cf.\ \citet{Luong2013}) 
we keep the overall method language independent. \autoref{fig:wordmaps} provides
an example from the word map derived for our Czech data set.

The choice of an unsupervised method of segmentation in this chapter stands in
contrast to the supervised compound segmentor used in \autoref{ch:hpylm}.
The evaluation setup there was simple given the focus on one language, so that
there was no downside in relying on a supervised segmentor. In this
chapter, we aim to have a broader evaluation on multiple languages. The use of
an unsupervised tool such a Morfessor simplifies the empirical work by not
requiring special models or calibration for each new language, even though this approach
may affect segmentation quality adversely. The \lblpp models are
specifically devised to be robust against noisy segmentation.
In the experiments that follow, we set the single \emph{perplexity threshold} parameter of Morfessor to 50 and
400 for \textsc{Data-1m} and \textsc{Data-Main}, respectively, based on prior
experience. Segmentation quality was gauged informally by manual inspection of
the output. \autoref{tbl:dist:avg-morphemes} provides statistics that summarise
the degree of segmentation obtained in each language.

\begin{table}[tb]
\centering 
\begin{tabular}{l>{\textit\bgroup}l<{\egroup}} \toprule 
\textbf{Word} $v$ & \textbf{Factors} $\left(f_1,\dots,f_{|\mu(v)|}\right)$                     \\ \midrule
aktivovanými      & aktivovanými aktiv\substem ova\subsuf ný\subsuf mi\subsuf \\
čerpány           & čerpány čerpá\substem ny\subsuf                                           \\
dorozumívat       & dorozumívat do\subpre rozum\substem í\subsuf vat\substem   \\
modlitba          & modlitba modlitba\substem                                                         \\
nejvážnější       & nejvážnější nej\subpre váž\substem nější\subsuf         \\
nepodkopal        & nepodkopal ne\subpre podkopal\subsuf                                     \\
škola             & škola škol\substem a\subsuf                                               \\
školou            & školou škol\substem ou\subsuf                                             \\
školu             & školu škol\substem u\subsuf                                               \\
školy             & školy škol\substem y\subsuf                                               \\
\bottomrule 
\end{tabular}
\caption[Word map snippet for Czech]%
{A fragment of the word map $\mu(v)$ obtained for Czech.}
\label{fig:wordmaps}
\end{table}

\begin{table}[tb]
  \centering
  \begin{tabular}{lcc} \toprule
    & \multicolumn{1}{c}{\textsc{Data-1m}} & \multicolumn{1}{c}{\textsc{Data-Main}} \\ \midrule
    \Lcs & 2.27 & 1.92 \\
    \Lde & 2.81 & 2.98 \\
    \Len & 1.72 & 1.50 \\ 
    \Les & 2.03 & 1.71 \\
    \Lfr & 2.26 & 1.78 \\
    \Lru & 2.32 & 1.84 \\ \bottomrule
  \end{tabular}
  \caption[Average number of morphemes via Morfessor segmentation]%
  {Average number of morphemes per word type obtained with Morfessor
    for each language (see \S\ref{sec:dist:wordfactorisation}). These numbers reflect the
    segmentation only and do not include the surface factors.}
  \label{tbl:dist:avg-morphemes}
\end{table}

\subsection{Model selection}
\label{sec:dist:selection}

\begin{figure}[tb]
\centering 
\includegraphics[width=0.6\textwidth]{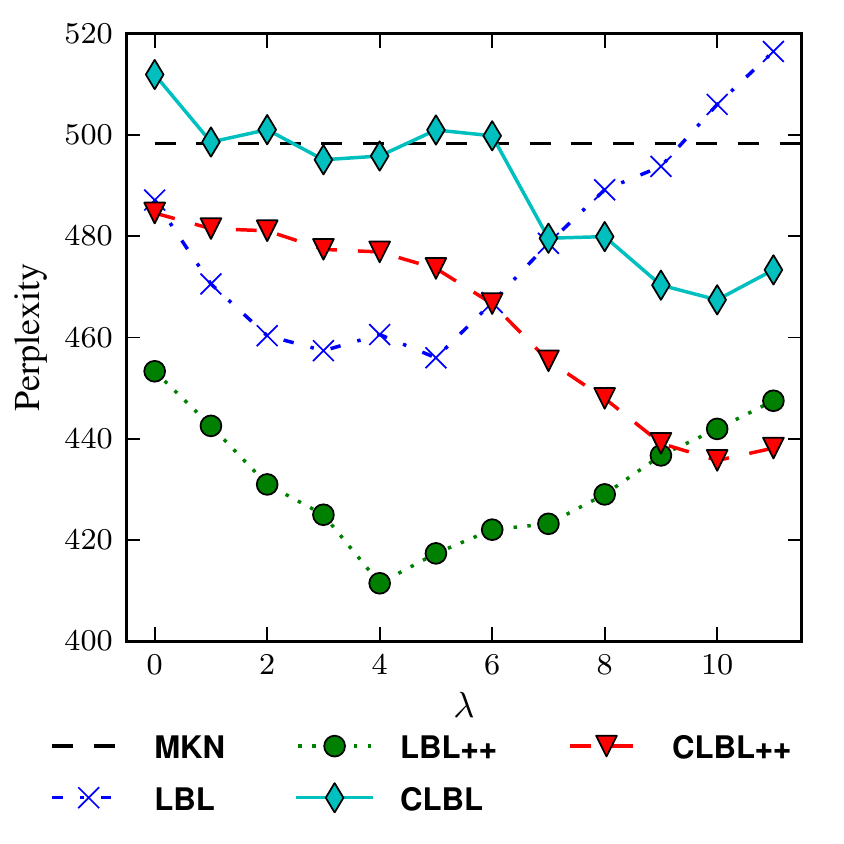}
\caption[Tuning regularisation strength]%
{Relative performance of models is sensitive to regularisation
  strength~$\lambda$ when using 1m training tokens. Perplexities are measured
  on the devset \corpus{newstest2011}. 
}
\label{fig:lambda}
\end{figure}

\paragraph{Results on \textsc{Data-1m}.}
The LBL training objectives include a weight penalty term
$0.5\lambda\|\boldsymbol{\theta}\|^2$, which we tune against the
\corpus{newstest2011} dataset.  For the 1m-word corpus size, we find tuning
$\lambda$ to be vital for good model performance (Figure~\ref{fig:lambda}).

\begin{figure}[tb]
    \centering 
    \includegraphics[ width=0.7\textwidth]{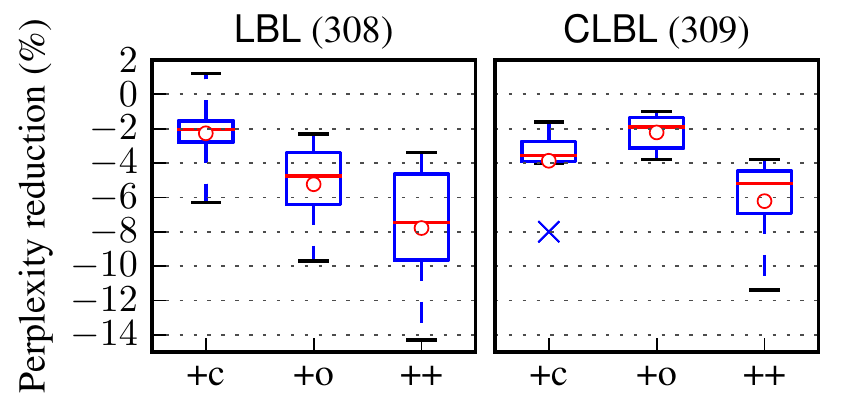}
    \caption[Model selection results]%
    { Model selection results. Box-plots show the spread,
      across 6~languages, of relative perplexity reductions obtained by each
      type of additive model against its non-additive baseline, for which
      median absolute perplexity is given in parentheses; for \mkn, this is~348.
      Each box-plot summarises the behaviour of a model across languages.
      Circles give sample means, while crosses show outliers beyond~3$\times$
      the inter-quartile range.
    }
    \label{fig:boxplots-ppl-1m}
\end{figure}

The use of morphology-based, additive representations for both context and
output words (models++) yielded perplexity reductions on all 6 languages when
using 1m training tokens (Table~\ref{tbl:ppl_1m}). Furthermore, these double-additive models
consistently outperform the ones that factorise only context (+c) or only
output (+o) words, indicating that context and output contribute complementary
information and supporting our hypothesis that is it beneficial to model
morphological dependencies \emph{across} words. The results are given by language in
Table~\ref{tbl:ppl_1m} and summarised visually in
Figure~\ref{fig:boxplots-ppl-1m}.

The impact of our \clblpp method varies by language, correlating with
vocabulary size: Russian benefited most, followed by Czech and German.  Even on
English, often regarded as having simple morphology, the relative improvement
is 4\%.

The relative merits of the +c and +o schemes depend on which model is used as
starting point.  With \lbl, the output-additive scheme (\lblpo) gives larger
improvements than the context-additive scheme (\lblpc) (Figure~\ref{fig:boxplots-ppl-1m}). The reverse is true for
\clbl, indicating the class decomposition dampens the effectiveness of using
morphological information in output words. 

The non-linear class-based model \nclbl obtains improvements up to 4\% over its
linear counterpart, except for Russian where it performs worse.  Significantly,
we observe that our purely linear \clblpp outperforms the non-linear \nclbl
model, establishing an alternative to the non-linear model.  Using
additive representations with the non-linear model further decreases perplexity
slightly, though the relative improvements obtained when using them for output
words only are very small.

The use of classes increases perplexity slightly compared to the \lbl{}s (see
Table~\ref{tbl:ppl-1m-avg-x-langs}), but this is in exchange for much faster
computation of language model probabilities, allowing the \clbl{}s to be used
in a machine translation decoder (\S\ref{sec:eval-mt}). On average, the \clbl{}s
obtain a 35-fold speed-up over the \lbl{}s. The running time of a model was calculated by measuring
the total wall-time it took to compute the test-set perplexity, and then
averaging across languages. For example, the \lbl{} computed the test-set
perplexity in 209~seconds, while the \clbl{} did so in 6~seconds. The test sets
contain 72650~tokens on average.
\label{correction:timing}

\begin{table}[tb]
    \centering 
    \begin{betweenfootnoteandscriptsize}
    \input{tbldata_ppl_1m}

    \caption[Test-set perplexities by model variant on small data]%
    {   Test-set perplexities for \textsc{Data-1m},
        measuring the separate effects of context-only~(+c), output-only~(+b)
        and double~(++) additive representations when applied to the linear and
        non-linear class-based LBL.
        Additive model results are also given in terms of percentage reduction
        in perplexity relative to the corresponding non-additive model.
    }
    \label{tbl:ppl_1m}
  \end{betweenfootnoteandscriptsize}
\end{table}

\begin{table}[tb]
    \centering
    \begin{tabular}{llll} \toprule
        \multicolumn{2}{r}{\mkn}    & \multicolumn{2}{l}{360} \\ \midrule
        \lbl    & 317 & \clbl   & 320 \\
        \lblpc  & 309 & \clblpc & 307 \\
        \lblpo  & 299 & \clblpo & 313 \\
        \lblpp  & \textbf{290} & \clblpp & 299 \\
        \bottomrule
    \end{tabular}
  \caption[The effect of classes vs.\ no classes]%
    {The role of classes. Test-set perplexity averaged across
    languages, using \textsc{Data-1m} for training.}
    \label{tbl:ppl-1m-avg-x-langs}
\end{table}

\subsection{Intrinsic evaluation}
\label{subsec:dist:xxx}

\paragraph{Results on \textsc{Data-Main}.}
Based on the outcomes of the small-scale evaluation, we focus our main language
model evaluation on the additive class-based model \clblpp in comparison to
\clbl and \mkn baselines, using the larger training dataset, with vocabularies
of up to 500k word types.

The overall trend that morphology-based additive representations yield lower
perplexity carries over to this larger data setting, again with the biggest
impact being on Czech and Russian (Table~\ref{tbl:ppl_mono}, top). Perplexity improvements
are in the \mbox{2\%--6\%} range, slightly lower than the corresponding
differences on the small data.

Our hypothesis is that much of the improvement is due to the additive
representations being especially beneficial for modelling rare words. We test
this by repeating the experiment under the condition where all word types
occurring only once are excluded from the vocabulary \mbox{($\kappa$=1)}. If
the additive representations were not beneficial to rare words, the outcome
should remain the same. Instead, we find the relative improvements become a lot
smaller (Table~\ref{tbl:ppl_mono}, \mbox{bottom}) than when only excluding some
singletons (\mbox{$\kappa$=0.05}), which supports that hypothesis.

\begin{table}[tb]
    \centering     %
    \input{tbldata_ppl_mono}

    \caption[Test-set perplexities for main dataset]%
    {   Test-set perplexities for \textsc{Data-Main} using two
        vocabulary pruning settings.  Percentage reductions are relative to the
        preceding model, \eg the first Czech \Dmodelx{CLBL} improves over
        \Dmodelx{MKN} by 20.8\% (rel1); the \Dmodelx{CLBL++} improves over that
        \Dmodelx{CLBL} by a \emph{further} 5.9\% (rel2).
    }
    \label{tbl:ppl_mono}
\end{table}

Model perplexity on a whole dataset is a convenient summary of its intrinsic
performance, but such a global view does not give much insight into \emph{how}
one model outperforms another.  To focus more directly on the source of
improvements, we partition the test data into subsets of interest and measure
perplexity over these subsets.

\subsubsection{Analysis by frequency}
\begin{figure}[t]
    \centering 
    \includegraphics[trim=0 0mm 0 0,clip, width=0.7\textwidth]{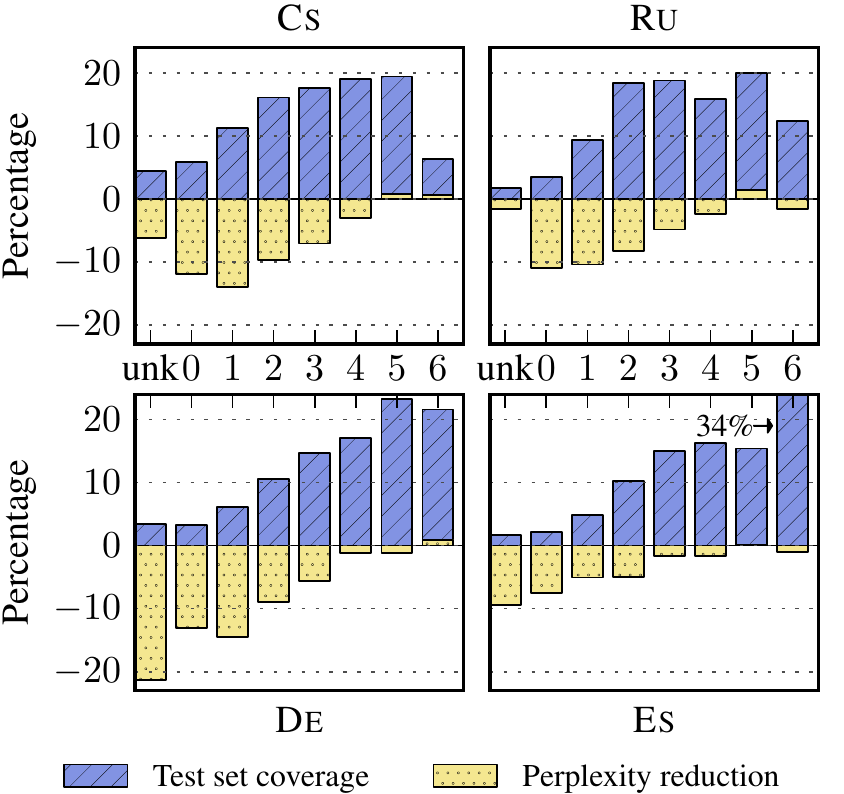}
    \caption[Results broken down by token frequency]%
    {   Perplexity reductions by token frequency, \Dmodelx{CLBL++} relative to \Dmodelx{CLBL}.
        Dotted bars extending further down are better.
        A bin labelled with a number~$x$ contains those test tokens that occur 
        $y \in [10^x,10^{x+1})$ times in the training data.
        Striped bars give percentage test-set coverage for each bin.
                }
    \label{fig:ppl-by-freq}
\end{figure}

We first partition on token frequency, as computed on the training data.
Figure~\ref{fig:ppl-by-freq} provides further evidence that the additive models
have most impact on rare words generally, and not only on singletons.  Czech,
German and Russian see relative perplexity reductions of 8\%--21\% for words occurring
fewer than 100 times in the training data.  Reductions become negligible for
the high-frequency tokens. These tend to be punctuation and closed-class words,
where any putative relevance of morphology is overwhelmed by the fact that the
predictive uncertainty is very low to begin with (absolute PPL\textless10 for
the highest frequency subset).  For the morphologically simpler Spanish case,
perplexity reductions are generally smaller across frequency scales.

\subsubsection{Analysis by part of speech}
\begin{figure}[t]
    \centering 
    \includegraphics[ 
     width=0.65\textwidth]{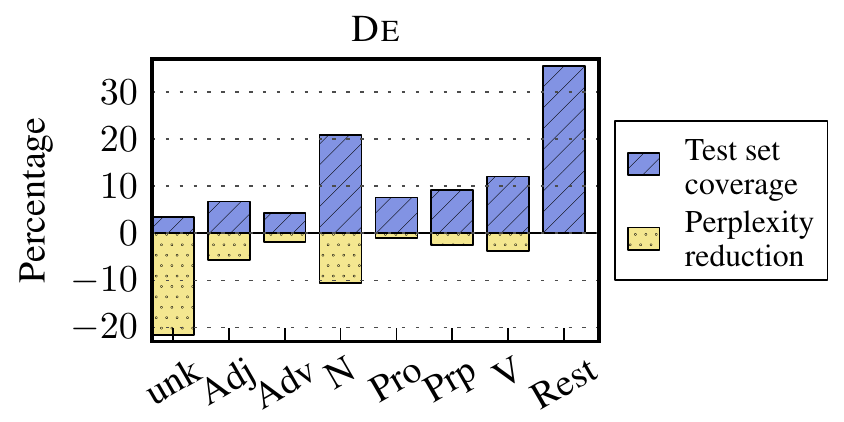}
    \caption[Results broken down by part of speech]%
     {Perplexity reductions by part of speech, \clblpp
      relative to \clbl on German.
        Dotted bars extending further down are better.
        Tokens tagged as foreign words or other opaque symbols resort under ``Rest''.
        Striped bars are as in Figure~\ref{fig:ppl-by-freq}.
              }
    \label{fig:ppl-de-by-pos}
\end{figure}

We also break down perplexity reductions by part of speech tags, focusing on German.
We used the decision tree-based tagger of \citet{Schmid2008}, which reportedly
has a tagging accuracy of 91\%.  Aside from unseen tokens, the biggest
improvements are on nouns and adjectives (Figure~\ref{fig:ppl-de-by-pos}),
suggesting our segmentation-based representations help abstract over German's
productive compounding.

German noun phrases require agreement in \mbox{gender}, case and number, which
are marked overtly with fusional morphemes, and we see large gains on such test
\ngrams: 15\% improvement on adjective-noun sequences,
and 21\% when considering the more specific case of adjective-adjective-noun
sequences.  An example of the latter kind is 
``der 
ehemalig$\bm{\cdot}$e 
sozial$\bm{\cdot}$ist$\bm{\cdot}$isch$\bm{\cdot}$e
bildung$\bm{\cdot}$s$\bm{\cdot}$minister'' 
(\emph{the former socialist minister of education}), where the morphological agreement surfaces in the \mbox{repeated e-suffix}.

\begin{figure}[tb]
    \centering 
    \includegraphics[width=0.65\textwidth]{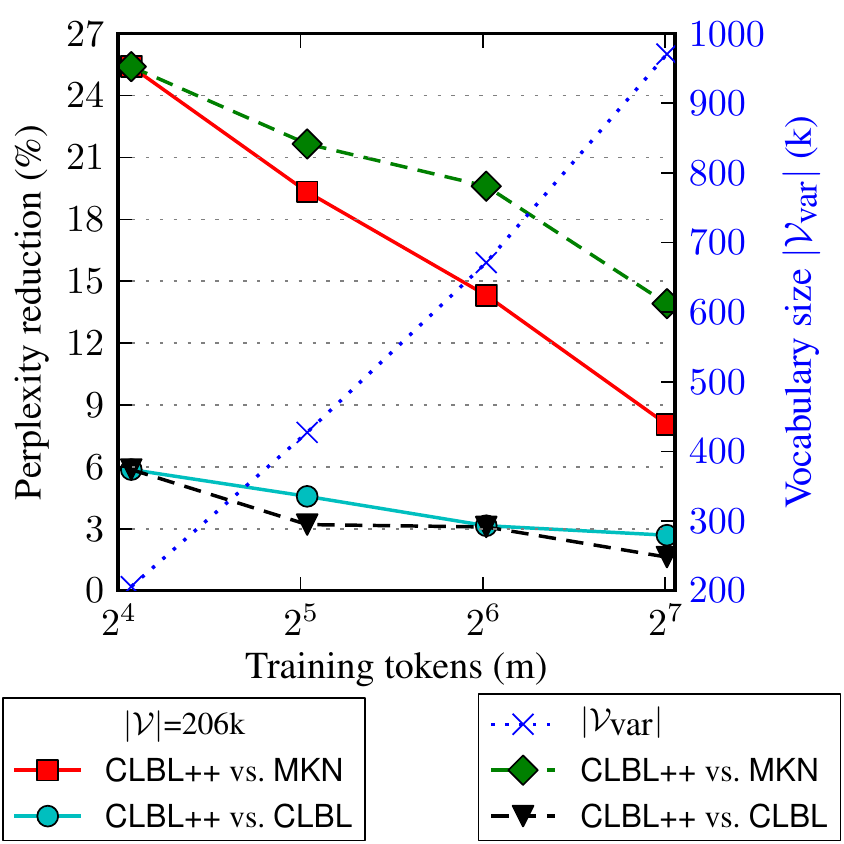}
    \caption[Effect of using more training data]%
     {Relative perplexity reductions
      obtained when varying the Czech training data size (16m--128m).
        In the first batch, the vocabulary was first held fixed to the set~$\V$
        as obtained for \textsc{Data-Main} (squares \& circles).
        In the second batch, the vocabulary~$\V_{\text{var}}$ varied according
        to the data set of the given size (diamonds \& triangles).
    }
    \label{fig:datascale}
\end{figure}

\subsubsection{Scaling}
We conducted a final scaling experiment on Czech by training models on
increasing amounts of data from the monolingual news corpora.  Improvements
over the \mkn baseline decrease, but remain substantial at 14\% for the largest
setting when allowing the vocabulary to grow with the data.  Maintaining a
constant advantage over \mkn requires also increasing the dimensionality $d$ of
representations \cite{Mikolov2013}, but this was outside the scope of our
experiment.  Although gains from the additive representations over the \clbl
diminish down to \mbox{2\%--3\%} at the scale of 128m training tokens
(Table~\ref{fig:datascale}), these results demonstrate the tractability of our
approach on large vocabularies of nearly 1m types.

\subsection{Application to word similarity rating}
\label{sec:dist:task_wordsim}
In the previous section, we established the positive role that morphological
awareness played in building DLMs that better
predict unseen text.  Here we focus on the quality of the word representations
learnt in the process.  We evaluate on a standard word similarity rating task,
where one measures the correlation between cosine-similarity scores for pairs
of word vectors and a set of human similarity ratings.  An important aspect of
our evaluation is to measure performance on multiple languages using a single
unsupervised, model-based approach.

Morpheme vectors from the \clblpp enable handling OOV test words in a more
nuanced way than using the global unknown word vector.
In general, we compose a vector 
$\tilde{\mb{u}}_v=[\tilde{\mb{q}}_v;\tilde{\mb{r}}_v]$ 
for a word $v$ according to a \emph{post hoc} word map $\mu'$ by summing and 
concatenating the factor vectors $\mb{r}_f$ and $\mb{q}_f$,
where \mbox{$f \in \mu'(v) \cap \F$}.
This ignores unknown morphemes occurring in OOV words, and uses 
$[\mb{q}_{\text{\unk}};\mb{r}_{\text{\unk}}]$ for
$\tilde{\mb{u}}_{\text{\unk}}$ only if all morphemes are
unknown.\footnote{Concatenation of context and target vectors performed better
  overall than using either type individually, suggesting the two
  representation spaces capture complementary information.}

To see whether the morphological representations improve the quality of vectors
for known words, we also report the correlations obtained when using the
\clblpp word vectors directly, resorting to $\tilde{\mb{u}}_{\text{\unk}}$ for
all OOV words \mbox{$v\notin \V$} (denoted ``$-$\compose'' in the results).
This is also the strategy that the baseline \clbl model is forced to follow for
OOVs.

We evaluate first using the English rare-word dataset (\data{Rw}) created by
\citet{Luong2013}.  Its 2034 word pairs contain more morphological complexity
than other well-established word similarity datasets,
\eg crudeness---impoliteness.  We compare against their context-sensitive
morphological recursive neural network (csmRNN), using Spearman's rank
correlation coefficient, $\rho$.  Table~\ref{tbl:wordsim_rw} shows our model
obtaining a $\rho$-value between the two csmRNN results, depending on which baseline word embeddings they used to initialise the csmRNN.

This is a strong result given that our vectors come from a simple linear
probabilistic model that is also suitable for integration directly into a
decoder for translation (\S\ref{sec:eval-mt}) or speech recognition, which is
not the case for csmRNNs.  Moreover, the csmRNNs were initialised with
high-quality, publicly available word embeddings trained over weeks on much
larger corpora of 630m--990m words \cite{CollobertWeston2008,HSMN2012}, in
contrast to ours that are trained from scratch on much less data.  This renders
our method directly applicable to languages which may not yet have those
resources.

A minor difference in configuration is that our vectors are 100-dimensional as
opposed to the csmRNNs' 50 dimensional ones.
Using embeddings\footnote{Source: http://metaoptimize.com/projects/wordreprs}\ %
from the hierarchical LBL \cite{Mnih2008}, we verified that the doubling in
dimensionality increased $\rho$ by less than 0.02 absolute, so this difference
should be negligible in our evaluation.

Relative to the \clbl baseline, our method performs well on datasets across
four languages.  For the English \data{Rw}, which was designed with morphology
in mind, the gain is 64\%.  But also on the standard English WS353 dataset
\cite{Finkelstein2002}, we get a 26\% better correlation with the human
ratings.  On German, the \clblpp obtains correlations up to three times
stronger than the baseline, and 39\% better for French
(Table~\ref{tbl:wordsim_multilang}).

\begin{table}[tb]
    \centering
    \begin{tabular}{lrlr} \toprule
        \multicolumn{2}{c}{\emph{\cite{Luong2013}}} &\multicolumn{2}{c}{\emph{Our models}} \\ \midrule
        HSMN        & 2  & \clbl       & 18 \\ 
        HSMN+csmRNN & 22 & \clblpp     & \textbf{30} \\ 
        C\&W        & 27 & $-$\compose & 20   \\ 
        C\&W+csmRNN & \textbf{34}      &   \\  
        \bottomrule
    \end{tabular}
    \caption[Word-pair similarity rating (rare word dataset)]%
      {Word-pair similarity task (English), showing Spearman's
      \rhotimes for the correlation between model scores and human ratings on
      the English {\sc Rw} dataset.  The \mbox{csmRNNs} benefit from
      initialisation with high quality pre-existing word embeddings, while our
      models used random initialisation. HSMN refers to the embeddings from \citep{HSMN2012}; C\&W to those from \citep{CollobertWeston2008}.
    }
    \label{tbl:wordsim_rw}
\end{table}

\begin{savenotes}
    \begin{table}[t]
        \centering
        \begin{tabular}{l *{6}{@{\hskip 0.1em}c} } \toprule
Datasets:\footnote{
    \Les WS353 \cite{Hassan2009}; 
    Gur350 \cite{Gurevych2005};
    RG65 \cite{RubensteinGoodenough65} with \Lfr \cite{Joubarne2011};
    ZG222 \cite{Zesch2006}.
}
            & \multicolumn{2}{c}{WS}
            & \multicolumn{1}{c}{Gur}
            & \multicolumn{2}{c}{RG}
            & \multicolumn{1}{c}{ZG}            
            \\
            & \multicolumn{1}{c}{\Len}  & \multicolumn{1}{c}{\Les}
            & \multicolumn{1}{c}{\Lde}
            & \multicolumn{1}{c}{\Len}  &  \multicolumn{1}{c}{\Lfr}  &  \multicolumn{1}{c}{\Lde} \\ \midrule
            HSMN          & 63        & --       & --     & 63        & --   & -- \\ 
            +csmRNN       & 65        & --       & --     & 65        & --   & -- \\ \midrule
            \clbl       & 32        & 26       & 36     & {\bf 47}  & 33   & 6 \\
            \clblpp     & 39  & {\bf 28}    & {\bf 56}  & 41        & {\bf 45}   & {\bf 25} \\  
            $-$\compose & {\bf 40}  & 27       & 44     & 41        & 41   & 23 \\ \midrule
            \multicolumn{1}{l}{\# pairs} & \multicolumn{2}{c}{353} 
            &  \multicolumn{1}{c}{350} 
            &  \multicolumn{2}{c}{65}
            &  \multicolumn{1}{c}{222}  
            \\ \bottomrule
        \end{tabular}
        \caption[Word-pair similarity rating in multiple languages]%
          {Word-pair similarity task in multiple languages,
             showing Spearman's \rhotimes and the number of word pairs in each data set.
            As benchmarks, we include the best results from \citet{Luong2013}, who relied on more data and pre-existing embeddings not available in all languages. In the penultimate row our model's ability to compose vectors for OOV words is suppressed.
        }
        \label{tbl:wordsim_multilang}
    \end{table}
\end{savenotes}

\subsection{Qualitative analysis of morpheme vectors}
\label{sub:dist:morpheme_vector_viz}
\begin{figure}[bt]
    \centering 
    \includegraphics[trim=0 1.5mm 0 1mm,clip,width=0.48\textwidth]{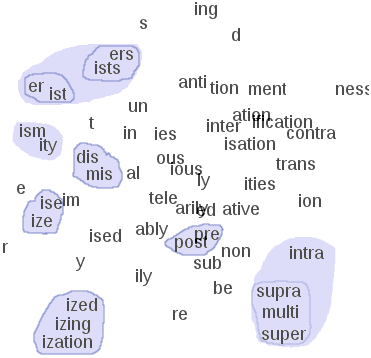}
    \caption[Visualisation of morpheme vectors]%
    {Visualisation of the morpheme vectors learnt by \clblpp
      on English data, using t-SNE \cite{MaatenHintonTSNE2008} for
      dimensionality reduction.  Shading added for emphasis.
    }
    \label{fig:tsne}
\end{figure}

A visualisation of the English morpheme vectors (Figure~\ref{fig:tsne})
suggests the model captured non-trivial morphological regularities: noun
suffixes relating to persons (writ\emph{er}, human\emph{ists}) lie close
together, while being separated according to number; negation prefixes share a
region 
(\mbox{un-}, \mbox{in-}, \mbox{mis-}, \mbox{dis-});
and relational prefixes are grouped 
(\mbox{surpa-}, \mbox{super-}, \mbox{multi-}, \mbox{intra-}),
with a potential explanation for their separation from \mbox{inter-} being that
the latter is more strongly bound up in lexicalisations (\emph{inter}national,
\emph{inter}section).
Some clusters abstract over spelling variation (\mbox{-ise} vs.\ \mbox{-ize}),
while others share similarity in terms of the stems they would attach to,
\eg global(\mbox{-ize},\mbox{-izing},\mbox{-ization}).

\subsection{Application to machine translation}
\label{sec:eval-mt}
The final aspect of our evaluation focuses on the integration of
class-decomposed log-bilinear models into a machine translation system.  To the
best of our knowledge, this is the first study to investigate large vocabulary
normalised DLMs inside a decoder when translating into a range of
morphologically rich languages.  We consider 5 language pairs, translating from
English into Czech, German, Russian, Spanish and French.

Aside from the choice of language pairs, this evaluation diverges from
\citet{Vaswani2013} by using normalised probabilities, a process made tractable
by the class-based decomposition and caching of context-specific normaliser
terms.  \citet{Vaswani2013} relied on unnormalised model scores for efficiency,
but do not report on the performance impact of this assumption.  It is not
clear how reliable an unnormalised language model score is as a feature that
must help the translation model discriminate between alternative hypotheses, as
frequency-effects would give rise to scores of very different scales.

We use \tool{cdec} \cite{Dyer2010, DyerFastAlign2013} to build symmetric
word-alignments and extract rules for hierarchical phrase-based translation
\cite{Chiang2007}.  Our baseline system uses a standard set of features in a
log-linear translation model. This includes a baseline 4-gram \mkn language
model, trained with \tool{SRILM} \cite{Stolcke02srilm} and queried efficiently
using \tool{KenLM} \cite{Heafield2011}.  The DLMs are integrated directly into
the decoder as an \emph{additional}
feature function, thus exercising a stronger influence on the search than in
n-best list rescoring.  Translation model feature weights are tuned with MERT
\cite{Och2003a} on \corpus{newstest2012}.

\begin{table}[tb]
    \centering
    \begin{tabular}{l 
            c @{}>{\small\bgroup \;(}c<{)\egroup}  
            c @{}>{\small\bgroup \;(}c<{)\egroup}
            c @{}>{\small\bgroup \;(}c<{)\egroup}
        } \toprule
        & \multicolumn{2}{c}{\mkn}   & \multicolumn{2}{c}{\clbl} & \multicolumn{2}{c}{\clblpp}  \\   
        \midrule 
        \data{En}$\rightarrow$\Lcs           & 12.6     & 0.2 & 13.2 & 0.1 & \textbf{13.6} & 0.04   \\
        \phantom{\data{En}$\rightarrow$}\Lde & 15.7     & 0.1 & \textbf{15.9} & 0.2 & 15.8 & 0.4    \\
        \phantom{\data{En}$\rightarrow$}\Les & 24.7     & 0.4 & 25.5 & 0.5 & \textbf{25.7} & 0.3    \\
        \phantom{\data{En}$\rightarrow$}\Lfr & 24.1     & 0.2 & 24.6 & 0.2 & \textbf{24.8} & 0.5    \\
        \phantom{\data{En}$\rightarrow$}\Lru & 15.9     & 0.2 & 16.9 & 0.3 & \textbf{17.1} & 0.1    \\
        \data{Cs}$\rightarrow$\Len           & 19.8     & 0.4 & \textbf{20.4} & 0.4 & \textbf{20.4} & 0.5 \\
        \bottomrule
    \end{tabular}
    \caption[Translation results using BLEU metric]%
    {\bleu scores (case-insensitive) obtained on
      \corpusx{newstest2013}, with standard deviation over 3 MERT runs given in
      parentheses.  The two right-most columns use the listed DLM as a feature
      in addition to the MKN feature, \ie these MT systems have at most 2
      LMs. Language models are from Table~\ref{tbl:ppl_mono} (top).
    }
    \label{tbl:bleu}
\end{table}

Table~\ref{tbl:bleu} summarises our translation results.
Inclusion of the \clblpp language model feature outperforms the \mkn-only
baseline systems by 1.2~\bleu points for translation into Russian, and by 1~point
into Czech and Spanish.  The \lp{En}{De} system benefits least from the
additional DLM feature, despite the perplexity reductions achieved in the
intrinsic evaluation.  In light of German's productive compounding, it is
conceivable that the bilingual coverage of that system is more of a limitation
than the performance of the language models.

On the other languages, the \clbl adds 0.5--1~\bleu points over the baseline,
whereas additional improvement from the additive representations lies within
MERT variance except for \lp{En}{Cs}.

\section{Related work}
\label{sec:dist:related}
Factored language models (FLMs) have been used to integrate morphological
information into both discrete \ngram LMs \cite{Bilmes2003FLM} and DLMs
\cite{Alexandrescu2006} by viewing a word as a set of factors.
\citet{Alexandrescu2006} demonstrated how factorising the representations of
context-words can help deal with out-of-vocabulary words, but they did not
evaluate the effect of factorising output words and did not conduct an
extrinsic evaluation.
The FLM notion has since been used to integrate part-of-speech information into
recurrent neural-network language models for ASR \cite{Wu2012} and
code-switching \cite{Adel2013a}.

A variety of strategies have been explored for bringing DLMs to bear on
machine translation.  Rescoring lattices with a DLM proved to be beneficial
for ASR \cite{Schwenk2004} and was subsequently applied to translation
\cite{Schwenk2006,SchwenkKoehn2008}, reaching training sizes of up to 500m
words \cite{SchwenkGPUs2012}.  For efficiency, this line of work relied heavily
on small ``shortlists'' of common words, by-passing the DLM and using a
back-off \ngram model for the remainder of the vocabulary.  Using unnormalised
DLMs during first-pass decoding has generated improvements in \bleu score for
translation into English \cite{Vaswani2013}.

More indirect approaches have approximated DLMs by sampling text from them and
then training conventional \ngram models on that \cite{Allauzen2013}, or by
casting the DLM directly as a back-off \ngram model through pruning
\cite{WangUtiyama2013}, while \citet{DuhNeubig2013} used them for selecting
relevant in-domain training data.

Vector representations have also been studied extensively within the area of
distributional semantics \citep{TurneyPantel2010}. There, vectors are typically
constructed directly from the occurrence counts of words in a corpus, with each
dimension of the vector space designating a specific context, for example. A variety of
techniques have been considered for how to compose word vectors into representations of
phrases or sentences
\citep[][\emph{inter alia}]{MitchellLapata2010,BaroniZamp2010}, which have been
extended upon to address the question of composing morpheme vectors into words
by \citet{Lazaridou2013}.

Other recent work has moved beyond monolingual vector-space modelling, incorporating
phrase similarity ratings based on bilingual word embeddings as a translation
model feature \cite{ZouSocher2013}, or formulating translation purely in terms
of distributed models \cite{NalRCTMs}.

Accounting for linguistically derived information such as morphology
\cite{Luong2013,Lazaridou2013} or syntax \cite{Hermann2013} has proved
beneficial to learning vector representations of words.  Our contribution is to
create morphological awareness in a \emph{probabilistic} language model. 

Parameter tying within an LBL has also been used in domain adaptation
\cite{Xiao2013}, where the representation space is subdivided into source,
target and background domains.

\section{Summary}
\label{sec:dist:summary}
We introduced a method for integrating morphology into probabilistic distributed language models.
Our method has the flexibility to be used for morphologically rich languages (MRLs) across a range of linguistic typologies.
Our empirical evaluation focused on multiple MRLs and different tasks.
The primary outcomes are that 
(i)~our morphology-guided DLMs improve intrinsic language model performance when
compared to baseline DLMs and \ngram MKN models;
(ii)~word and morpheme representations learnt in the process compare favourably in terms of a word similarity task to a recent more complex model that used more data, while obtaining large gains on some languages;
(iii)~machine translation quality as measured by \bleu was improved consistently
across six language pairs when using DLMs during decoding, although the morphology-based representations led to further improvements beyond the level of optimiser variance only for English$\rightarrow$Czech.
By demonstrating that the class decomposition enables full integration of a
normalised DLM into a decoder, we open up many other possibilities in this active modelling space.

%% file: ch-dist/tbldata_ppl_1m.tex
\centering
\begin{tabular}{l c |cccc|c || cccc|c } \toprule
& \Dmodelx{MKN} & \Dmodelx{CLBL} & +c & +o & \multicolumn{1}{c}{++} & ++  & \Dmodelx{nCLBL} & +c & +o & \multicolumn{1}{c}{++} & ++  \\    \midrule
\Lcs & 550 & 508 & -4.0\% & -3.4\% & -7.4\%  & 470 & 487 & -3.0\% & 0.7\%  & -4.6\% & \textbf{465}  \\
\Lde & 366 & 316 & -3.5\% & -1.0\% & -4.3\%  & 303 & 309 & -2.8\% & -0.6\% & -4.2\% & \textbf{296}  \\
\Len & 330 & 303 & -1.6\% & -1.3\% & -3.8\%  & \textbf{291} & 301 & -2.6\% & 0.1\%  & -3.0\% & 292  \\
\Les & 241 & 212 & -3.6\% & -2.3\% & -4.9\%  & 202 & 209 & -2.1\% & -0.2\% & -4.4\% & \textbf{200}  \\
\Lfr & 274 & 239 & -2.5\% & -1.5\% & -5.5\%  & 226 & 230 & -1.6\% & -0.7\% & -2.4\% & \textbf{225}  \\
\Lru & 396 & 343 & -8.0\% & -3.8\% & -11.4\% & \textbf{304} & 353 & -5.4\% & -0.6\% & -8.3\% & 324  \\
\midrule
avg & 360 & 320 & -3.9\% & -2.2\% & -6.2\% & \textbf{299} & 315 & -2.9\% & -0.2\% & -4.5\% & 300 \\
\bottomrule
\end{tabular}

%% file: ch-dist/tbldata_ppl_mono.tex
\begin{tabular}{l@{}>{\quad}  lr r r<{\%}  r r<{\%}  } \toprule
& &  & \multicolumn{2}{c}{\clbl} &\multicolumn{2}{c}{\clblpp} \\ 
& & \mkn & \multicolumn{1}{c}{abs} & \multicolumn{1}{c}{rel1}
       & \multicolumn{1}{c}{abs} & \multicolumn{1}{c}{rel2} \\
\midrule
\underline{$\kappa$=0.05}
& \Lcs & 862 & 683 & -20.8 & 643 & -5.9 \\ 
& \Lde & 463 & 422 & -8.9  & 404 & -4.2 \\ 
& \Len & 291 & 281 & -3.4  & 273 & -2.8 \\ 
& \Les & 219 & 207 & -5.7  & 203 & -1.9 \\ 
& \Lfr & 243 & 232 & -4.9  & 227 & -1.9 \\ 
& \Lru & 390 & 313 & -19.7 & 300 & -4.2 \\ 
\cmidrule{3-7}
& avg  & 411 & 356 & -10.6 & 342 & -3.5 \\
\midrule
\underline{$\kappa$=1.0}
& \Lcs & 634 & 477 & -24.8 & 462 & -3.1 \\ 
& \Lde & 379 & 331 & -12.6 & 329 & -0.9 \\ 
& \Len & 254 & 234 & -7.6  & 233 & -0.7     \\ 
& \Les & 195 & 180 & -7.7  & 180 & 0.02 \\ 
& \Lfr & 218 & 201 & -7.7  & 198 & -1.3 \\ 
& \Lru & 347 & 271 & -21.8 & 262 & -3.4 \\ 
\cmidrule{3-7}
& avg  & 338 & 282 & -13.7 & 277 & -1.6 \\
\bottomrule   
\end{tabular}

%% file: ch-pysrcag/ch-pysrcag.tex
\chapter{\label{ch:pymcfg}Unsupervised learning of complex morphology} 
\chapoverview{%
The previous two chapters considered methods for integrating morphological
information into two different types of LMs on the assumption that
an external source of information is available.
In this chapter, we go beyond sequence-based language modelling 
to consider how morphological information can be obtained in the first place
using unsupervised methods.
We introduce a novel approach to modelling concatenative and non-concatenative
morphology jointly.
This chapter is an extension of material originally published as
  \citep{Botha2013}.
}  

Past work on unsupervised morphology learning has overwhelmingly considered
\emph{concatenative morphology}, where surface words can be analysed into a
sequence of continuous morphemes or string segments.
Consecutive instances of the \emph{Morpho\-Challenge}
shared-task boosted contributions to unsupervised learning in this
regard \cite{kurimo10clef_lncs}.
But another important aspect of unsupervised learning of surface-level
morphology concerns \emph{non-concatenative} processes, where the morphemes of a
word do not necessarily coincide with contiguous strings but are instead
dispersed throughout the word.
This type of morphology has received much less attention within the realm of unsupervised
learning---the survey by
\citet{Hammarstrom2011} identifies only 10 of the 200+ studies within its
purview as targeting non-concatenative morphology.

Concatenative and non-concatenative morphological processes do not function in isolation. The
classical example is Arabic, which uses non-concatenative processes to derive
noun and verb stems, onto which various clitics and affixes attach in a
concatenative fashion. As another example, Tagalog verb conjugation makes extensive use of
infixing and circumfixing in its conjugation of verbs.\footnote{Tagalog is spoken in
  the Philippines.} 
In both these examples, a segmentational approach recovers some of the building
blocks of words, but naturally misses out on certain regularities, \eg the notion that the
Arabic stems \iteg{katabat} \iten{she wrote} and \iteg{>akotubu} \iten{I write} are bound together by
the shared, discontiguous root \iteg{ktb}; or that two Tagalog words
\iteg{\textbf{s}um\textbf{ulat}} and \iteg{\textbf{s}in\textbf{ulat}an} (different
conjugations of `wrote') share the same root, \iteg{sulat}.
An unsupervised model that does not have the expressivity to capture these kind
of relations therefore cannot benefit from the strong signals they encode.
Modelling non-concatenative morphology is thus a way of extracting more value
from the unannotated text by reducing sparsity.

The primary contribution of this chapter is the formulation of a probabilistic
model that jointly captures concatenative and non-concatenative morphology, as
exemplified above. We extend adaptor grammars \citep{Johnson2007a} to a mildly context-sensitive
grammar formalism that easily captures both kinds of surface-level morphology.
We argue that the general approach is suitable to a variety of settings, but
focus the application on Semitic morphology.

The structure of the chapter is as follows: The next section provides more
background on Semitic morphology. The subsequent section reviews existing
computational treatments of Semitic morphology and explains the relative merits
of an unsupervised approach.
The modelling work follows in three parts: First, we
cover background material for unsupervised learning with adaptor grammars. Secondly, we 
present the more expressive grammar formalism and demonstrate its application to
Semitic morphology. Finally, we report empirical evaluation results on Arabic and Hebrew.

\section{Basic overview of Semitic morphology}
\label{sec:pymcfg:semitic-overview}
The Semitic language family is a suitable test-bed for the modelling approach we
propose in light of the fundamentally non-concatenative root-templatic
morphology.\footnote{Many different linguistic terms apply to this discussion.
  For simplicity's sake we do not dwell on their subtle distinctions and use
  whichever term is most convenient and sufficiently precise in a given context.
  ``Root-templatic'' morphology is also referred to as root-and-pattern morphology. The
  particular sense in which it is non-concatenative is also variously denoted as
  transfixation, intercalation and interdigitisation.}
Member languages that are widely spoken today include Arabic, Hebrew, Amharic
and Tigrinya.

In contrast to many languages where the word stem is the most elementary
semantic building block, Semitic stems are derived from semantic root morphemes and
grammatical templates which combine in a non-concatenative way.\footnote{Linguistic morphology distinguishes between the consonant-vowel
  skeleton (\eg CVCVC) and the vocalism (\eg i\dots{}a) \cite{McCarthy1981}. We
  use ``templates'' to refer to the combination.}

The roots consist of between two and five \emph{radicals} which are exclusively
consonants. Roots are often referred to by the number of radicals they have.
\emph{Triliteral} roots featuring three radicals are the dominant kind. Because of
the dominance of triliteral roots, existing modelling approaches are often
limited to this class.
An aim of the model we introduce is that it should easily encode roots of different lengths.

Templates encode grammatical information, such as number or
person. To a first approximation, verb and noun stems are derived by interspersing
root radicals into templates.  In Arabic, for example,
\iteg{kitAb} \iten{book}, \iteg{kutub} \iten{books}, \iteg{katab} \iten{write}
and \iteg{takAtab} \iten{correspond} all derive from the root \iteg{ktb}. The root
cannot in general be identified as a continuous substring of the stem.

Such stems undergo further transformation to give rise to surface
word forms. Various inflectional affixes and clitics attach to stems by
concatenation to give rise to multi-segment surface words. 
A clitic embodies the syntactic properties of an independent word but nonetheless
surfaces as a bound morpheme.
Pronouns, prepositions, conjunctions and the definite
article are all cliticised in Arabic, which means segmentation is challenging also when
one ignores the non-concatenative element of the morphology.
Indeed, one hypothesis we investigate is that an account of the underlying root
has a disambiguating effect on segmentation (\S\ref{sec:mcfg:exp}).

The following list shows typical examples of the aforementioned morphological
processes,\footnote{See \cite{HabashArabNLPbook} for a more complete introduction to Arabic morphology.}
for surface words arising from stems based on the root \iteg{ktb}:
\begin{itemize}
\item verb person: \iteg{katab} \iten{write} $\Rightarrow$ \iteg{kataba} \iten{he wrote},
\iteg{katabat} \iten{she wrote}, \iteg{katabona} \iten{we wrote} 
\item pronominal enclitics: \iteg{katabato\textbf{hu}} \iten{she wrote \textbf{it}} 
\item tense proclitic: \iteg{>akotubu} \iten{I write} $\Rightarrow$ 
\iteg{\textbf{sa}>akotubu} \iten{I \textbf{will} write}
\item determiner:  \iteg{\textbf{Al}kitAbi} \iten{\textbf{the} book}
\item prepositional proclitics: \iteg{\textbf{bi}AlokitAbi} \iten{\textbf{for} the book}
\item conjunction: \iteg{\textbf{wa}biAlokitAbi} \iten{\textbf{and} for the book}
\end{itemize}

\subsection{Orthographic variety}\label{sec:pymcfg:orthography}
Another central property of Semitic languages is that their orthographic systems
are so-called \emph{abjads}, which means the explicit writing of vowels is partly or
completely optional.
The examples given so far use the less common \emph{vocalised} form, as it
illustrates the morphological processes better. In standard orthography, a
system of diacritic marks is used to indicate that vocalisation, especially in
literary and religious texts.
But the more dominant case is \emph{unvocalised} text, which suppresses some
vocalisation information. This can render individual words highly ambiguous so
that their correct interpretation relies on sentential context.
For example, the unvocalised word token \iteg{ktb} has at least 15
readings.\footnote{This example was obtained using the Xerox Arabic
  Morphological Analyzer, available at \url{https://open.xerox.com/Services/arabic-morphology}.}
As the focus in this chapter is on type-based morphological analysis it does not
include solving this disambiguation problem.\footnote{For studies that
  make direct use of sentential context to address this ambiguity problem, see
  \citep{HabashRambowMADA2005,Poon2009,Lee2011}.}
However, the relevance of the orthographic variation is that it complicates
computational approaches to dealing with Semitic text.
A key test for an unsupervised model is whether it can effectively
handle such variation without requiring manual modification. Aside from a
usability standpoint, the advantage of such a model is that it could learn
directly from text that is heterogeneous in its vocalisation.

\subsection{Complex non-concatenative processes outside scope}
The preceding characterisation of root-templatic derivation is a simplification.
It excludes cases where the correspondence between the
characters in the surface word and those in the constituent morphemes is not
one-to-one. Such cases occur where deletion and insertion of characters happens
during word formation. The first instance of this is weak roots, where one or
more of the radicals do not appear in the derived stem. The second instance
concerns gemination, where there is a lengthening of a consonant, \eg
the stem \iteg{kuttAb} \iten{writers} as derived from root \iteg{ktb}.

These phenomena are present in the data used for our empirical evaluation though
not accounted for by the model we introduce. The reason for limiting the scope
this way is that the probabilistic modelling would otherwise be significantly
more complex. Instead, we focus on giving a unified probabilistic account
of concatenative and non-concatenative processes purely at the level of surface
strings.

\section{Existing approaches}
This section surveys the dominant existing computational treatments of Semitic morphology.
 The aim is to situate our unsupervised probabilistic
approach among alternative approaches while considering their relative merits.

\subsection{Rule-based methods}
There is an extensive literature on encoding linguistic rules explicitly in
formal computational frameworks to perform automatic morphological analysis.
The dominant strand of work uses finite-state transducers (FSTs).
FSTs are naturally suited to concatenative morphology, given that concatenation
is one of the fundamental operations in regular languages.  To deal with
non-concatenative phenomena, FSTs require careful augmentation
\cite{BeesleyKarttunenFSM2003}.  Early efforts on both Arabic
\citep{KatajaKosk1988} and Hebrew \citep{Lavie1988} made extensions to the
foundational two-level morphological framework \cite{Koskenniemi1984} to cater
for root-templatic morphology.  More complete approaches have directly
operationalised the influential theory of non-concatenative morphology proposed
by \citet{McCarthy1981}, using multi-tape transducers that are able to process
the root, template and vocalism of the Semitic stem separately
\cite{Kay1987,Kiraz2000}. The \tool{MAGAED} system extends this approach further and
its creators argue that the non-concatenative handling of stems is essential to
deal efficiently with the different Arabic dialects
\cite{HabashRambowKiraz2005,Habash2006a}. 
Other augmentations make use of registers to give FSTs some
limited memory to handle short-term dependencies \cite{Cohen-Sygal2006}, whereas
\citet{Gasser2009}
used weighted FSTs with a unification operation over attribute-value pairs.

The FST-based approaches are broadly lexical in their view of inflection and
derivation---the morphosyntactic properties of words are carried by lexical
morphemes. A separate strand of work in Semitic computational morphology is
based instead on \emph{inferential-realisational} theories of morphology
\cite{Stump2001}, where morphosyntactic properties are specified purely by rules.
\citet{Finkel2002} use this to formulate an account of Hebrew verb morphology
using hierarchies of inherited rules. 
\citet{SmrzThesis2007} developed a complete account of Arabic morphology that
falls within this theoretical class. His ElixirFM system makes use of a
functional morphology framework \cite{Forsberg2004} to formulate a
domain-specific language embedded in a functional programming language.

The aforementioned approaches in both strands of work share the property of
encoding morphological analysis and generation.
A more ad-hoc approach that is limited to analysis is that of
the Buckwalter Arabic Morphological Analyser \citep[BAMA][]{Buckwalter2002}. BAMA relies on
lists of stems and affixes together with consistency constraints on the co-occurrence of
different morphemes within a word. It ignores the non-concatenative derivation of
stems, but its lexica, which also indicate root morphemes, have been influential
in the development or evaluation of other methods \citep[][\emph{inter
  alia}]{Darwish2002,SmrzThesis2007,Rodrigues2007}.

These rule-based approaches typically perform exceptionally well at their task,
given the level of targeted design involved.  But the limitations are
substantial: A significant amount of manual effort and expert linguistic
knowledge is required to develop the methods described in this section. This
includes having a precise linguistic framework to operationalise, or a
sufficient amount of lexical information as in the case of BAMA.  Consequently,
transferring a given method or system to other dialects or languages within the
Semitic family can be a big undertaking. Finally, even within their target
language, the rule-based approaches that have a lexical component invariably
offer limited coverage and require manual addition for application to new
domains.

\subsection{Supervised learning}
Supervised learning of morphology allows some of the limitations of a rule-based
approach to be overcome, and shifts the resource requirement away from
linguistic expertise toward data. Common sources of supervision include pairs
of inflected words, words labelled with their roots, and lexica of affixes.  A
learning component then has to infer parametrisations or parameter values (or
both) from these labelled resources. The existing literature on Semitic
morphology that falls into this category is limited and fragmented.

One relevant strand of work considers the simpler problem of predicting the
Arabic broken plural (\eg \iteg{kutub}) given a base form (\eg \iteg{katab}).
\citet{Clark2001PartialSup,Clark2002} applies pair hidden Markov models (HMMs), which have two emission
streams, to train stochastic FSTs from such word-pair data. The approach is
well-motivated for non-concatenative phenomena such as Semitic stem derivation,
but it does not perform well in evaluation.

A more common theme has been to leverage the aforementioned sources of
supervision to construct methods for identifying the root morpheme of a given
word. \citet{Darwish2002} combines pre-existing Arabic resources with various
heuristics and rudimentary statistical models to devise an analyser that outputs
a ranked list of candidate roots for a word. 

\citet{Daya2008} approached root identification from the perspective of
character-based classification. They train off-the-shelf classifiers for each
radical and consider various ways of combining their predictions. Their
investigation demonstrates the feasibility of the supervised approach by
obtaining F-scores in the eighties, while transferring it from Hebrew to Arabic
required relatively little effort. The methods are however specific to
triliteral roots and do not contain a segmentational component.

\citet{Boudlal2009} employed a first-order HMM with roots as hidden
states to identify the root of a given word with an accuracy of
94\%; they relied on greedy methods and dictionaries of templates,
roots and affixes to limit the hidden state space.

\subsection{Unsupervised learning}
As stated in the introduction of this chapter, the existing work on unsupervised learning of
non-concatenative morphology is much more limited than concatenative morphology,
and this extends to Semitic morphology.  We take ``unsupervised'' to be
situations where instances of the expected output of a method are not supplied
during its training.

Probabilistic generative models constitute an attractive approach in this
category as they allow high-level intuitions to be encoded in a coherent
fashion. The only work we are aware of that directly targets root-templatic
morphology in this sense is the non-parametric Bayesian model of
\citet{Fullwood2013}.  Their model generates a word from a root, template, and
``residue'', which captures the non-root characters of the word.  For example,
\iteg{kitAbi} would arise through the combination of root(ktb), template(r-r-r-)
and residue(iAi). Each of these categories is modelled by an independent
PYP-distributed lexicon, where the base distributions support arbitrary length
sequences.  Their evaluation is limited to template identification accuracy,
where it performs very well.  Two main shortcomings of the model are the lack of
affix modelling and the atomicity of the lexicon elements:
\begin{enumerate}
  \item The model only targets stem-formation. Given a word and its inferred
    template and residue, one could in principle extract a segmentation into
    prefix, stem and suffix, but segmentation of multiple adjacent affixes is
    nonetheless precluded. 
  \item Lexicon elements are atomic in the sense that their characters are
    sampled independently from a uniform distribution over the alphabet. This
    means there is no aggregation of probability mass around the recurrence of,
    say, consonants in the root lexicon---the PYP only caches the full strings.
\end{enumerate}
The approach we introduce in \S\ref{sec:pymcfg:AG-for-root-template-morphology} was developed contemporaneously with
\citet{Fullwood2013}, but specifically overcomes these two shortcomings, while
aiming to be more easily extensible.

Other probabilistic approaches targeting Semitic segmentation have exploited
regularities beyond word types, such as using sentential context
\citep{Poon2009,Lee2011} or multi-lingual phrases \citep{Snyder2008}.

There are unsupervised methods that indirectly address root-templatic morphology
without necessarily attempting to model the word formation process.  Words
sharing the same root can be clustered based on heuristic scoring of character
tuples \cite{DeRoeck2000}, which is useful for information retrieval.
\citet{BaroniMatiasekTrost2002} similarly acquire morphologically related word
pairs (\eg \iteg{Verkauf}, \iteg{Verkäufe}) from raw text using string-edit distances and
distributional cues.\footnote{This is a variation of the knowledge-free method
  by \citet{SchoneJurafsky2000}, which was affix-based and thus limited to
  concatenative morphology.} Their approach is in principle applicable to
Semitic morphology since the string-edit operations suit root-templatic
morphology, but they frame the method as a potentially useful precursor to
morphological analysis and did not apply it to Semitic languages.

\section{Unsupervised learning with CFGs}
\label{sec:pymcfg:ag-main}
Our overall approach to the problem of learning morphology from
unannotated text is to construct a model based on rewrite rules. This type of
approach has been applied successfully to morphological segmentation
\citep{Johnson2007a,JohnsonGrifGold2007PCFGs,Johnson2008} and offers well-defined mechanisms
for encoding prior intuitions about the morphology.

The following two sections develop the relevant background material for learning
with context-free grammars, which we subsequently build on for learning
non-concatenative phenomena.

\subsection{Probabilistic context-free grammars}
\label{sec:pymcfg:pcfgs}
A context-free grammar $\mathcal{G}_{\text{CFG}}=(S,N,T,P)$ is specified by a set of non-terminal
symbols $N$, terminal symbols $T$ and rewrite rules $P$ of the form $A\arrowg
\gamma$, where $A\in N$ and $\gamma \in (N\cup T)^+$. $S\in N$ is the start
symbol.
A string $w \in T^+$ in the implied context-free language is derived from the
start symbol $S$ through a sequence of rule applications $r_1,\dots,r_k \in P$
that expand non-terminal symbols until no further expansions are possible.

This symbolic grammar can be cast as a generative probabilistic model, known as
a probabilistic context-free grammar (PCFG), by
defining a probability distribution over each subset $P_A \subset P$ that contains
the rules rewriting a symbol $A\in N$. That is, each rule $r \in P_A$ has an
associated real value $\theta_r$ such that $0 \leq \theta_r \leq 1$ and $\sum_{r\in
  P_A}\theta_r=1$.
The stochastic procedure for generating a string from the model follows the same
derivation pattern outlined above, but samples the rule $r$ to use for expanding
a given non-terminal $A$ from a categorical distribution parametrised by
$\theta$. The key point is that this assumes conditional independence among
expansions given a fixed $\mathcal{G}_{\text{CFG}}$, so that the
joint probability of a string $w$ and a derivation 
$R=r_1,\dots,r_k$ is $P(w,R|\theta) = \prod_{r\in R} \theta_r$.

Two learning problems thus become apparent---what are the rules in $P$, and what
are their probabilities $\theta$? The former problem is 
challenging to pursue in a completely unsupervised fashion
\citep[cf.][]{StolckeOmohundro1994,Clark2001}.
If one assumes a set of rules is given, the second issue of learning their
probability distributions can be approached through
maximum likelihood estimation \cite{LariYoung1990,GoodmanThesis1998} or in a
parametric Bayesian fashion by introducing a Dirichlet prior over $\theta$
\cite{JohnsonGrifGold2007PCFGs}.

\subsection{Context-free adaptor grammars}
\label{sec:pymcfg:agcfg}
Adaptor grammars \cite{Johnson2007a} provide an elegant solution to both
learning problems mentioned above.
The core idea of adaptor grammars is that whole sub-trees
are memorised and can be reused when generating strings from a
probabilistic grammar.
This relaxes the independence assumptions governing non-terminal expansion in
the generative process of a PCFG to better
match the pattern that non-trivial
constructions in language re-occur across different instances. For example, 
the same stem would occur in many different surface word forms. Likewise, a
given discontiguous Hebrew root would occur across different stems, which is the kind of
regularity we intend to capture with our extension of adaptor grammars beyond
context-free rules (\S\ref{sec:pymcfg:inference-ag}).

The tendency for a sub-tree fragment to be reused is governed by the choice of
adaptor function. We follow earlier applications
\citep[\eg][]{Johnson2007a,Huang2011} and use the Pitman-Yor process (PYP) as adaptor
function \cite{Pitman1995,Pitman1997}. As with the language models discussed in
\autoref{ch:hpylm}, the PYP's extensibility via its base distribution and its
power-law behaviour make it suitable for modelling distributions over trees.

\subsubsection{Formalism}
\label{sec:mcfg:cfg-ag-formalism}
A Pitman-Yor context-free adaptor grammar (\agcfg) is specified as a tuple
{$\gram{G}=(\gram{G_{\text{PCFG}}}, M, \bm{a}, \bm{b}, \bm{\alpha})$},
where \gram{G_{\text{PCFG}}}
is a PCFG as defined before and \mbox{$M \subseteq N$} is a set of
\emph{adapted non-terminals}.  The vectors~$\bm{a}$ and~$\bm{b}$, indexed by the
elements of~$M$, are the discount and concentration parameters for each adapted
non-terminal, with $a\in[0,1]$ and $b\geq0$.  $\bm{\alpha}$ are parameters to
Dirichlet priors on the rule probabilities~$\theta$.

\agcfg defines a generative process over a set of trees~$\bm{T}$.
Unadapted non-terminals $A' \in N \setminus M$ are expanded in the standard way
of PCFGs, as described in the previous section.

For each adapted non-terminal $A \in M$ there is a cache~$C_A$ which stores
terminating tree fragments having $A$ as their root.
We denote the tree fragment in $C_A$ that corresponds to the~$i$\th expansion of~$A$ as~$z_i$.
A sequence of indices \mbox{$\mb{z}_i=z_1,\dots,z_i$} therefore assigns
each individual expansions of~$A$ to some tree fragment in the cache.

Given a cache~$C_A$ that has~$n$ previously generated trees comprising~$m$
unique trees each used $n_1,\dots,n_m$ times (where $n=\sum_k n_k$), the tree
fragment for the next expansion of~$A$, $z_{n+1}$, is sampled conditional on the
previous assignments~$\bm{z}_n$ according to
\begin{equation}\label{eq:mcfg:seatconditional}
  z_{n+1} \mid \bm{z}_n,a,b \sim
  \begin{cases}
    \dfrac{n_k-a}{n+b} & \text{if } 1\leq z_{n+1} \leq m \\[3ex]
    \dfrac{am+b}{n+b} & \text{if }z_{n+1}=m+1,
  \end{cases}
\end{equation}
where~$a$ and~$b$ are the elements of~$\bm{a}$ and~$\bm{b}$ corresponding to~$A$.
\begin{enumerate}
\item The first case denotes the situation where a previously cached tree is reused
for this~$n+1$\th expansion of~$A$.
This expands~$A$ with a fully terminating tree fragment in a single step, meaning
that none of the nodes descending from~$A$ in the tree being generated are
subject to further expansion.
\item The second case by-passes the cache and expands~$A$ according to the rules~$P_A$
and rule probabilities~$\theta_A$ of the base grammar \gram{G_\text{PCFG}}.
This samples a sub-tree with root $A$ from the PYP base distribution, a process
during which other
caches~$C_B$ (s.t. $B\in M, B \neq A$) may come into play when expanding 
descendants of~$A$; thus a \agcfg can define a hierarchical stochastic process.
\end{enumerate}
Both cases eventually result in a terminating tree-fragment for~$A$, which is
then added to the cache, updating the counts~$n$, $n_{z_n+1}$ and
potentially~$m$.

The adaptation does not affect the string language of~\gram{G_S}, but it maps
the distribution over trees to one that is distributed according to the PYP.

\subsubsection{Recursion}
\label{sec:mcfg:recursion-comments}
Context-free grammars may, in general, include recursion in their rewrite rules,
either directly, \eg $\nt{X} \arrowg \nt{X}\; \nt{Y}$,
or indirectly, \eg $\nt{X} \arrowg \nt{Y}\; \nt{Z}$
and $\nt{Z} \arrowg \nt{X} \; \nt{Q}$.

A \agcfg that features this type of recursion among its adapted
non-terminals $M$ complicates sampler-based inference, since it can involve
repeated sampling from the same CRP without being able to update its counts
correctly. In theory, a Metropolis-Hastings accept/reject step could be used as
remedy but it is not obvious what proposal distribution would lead to acceptable
rejection rates. Thus we follow \citet{Johnson2007a} and restrict the adaptor
grammars we use to exclude recursive rewrite rules among adapted non-terminals.
(This is what the ``$B\neq A$'' condition in step~2 of the previous section
alludes to.)

\section{An idealised grammar for learning morphology}
\label{sec:pymcfg:idealised}
A unifying theme in this thesis is to show that specific though high-level
intuitions about morphology can be encoded in probabilistic models to improve
their quality and capabilities.
The high-level intuition we want to employ here is that morphological
derivation and inflection are often non-concatenative.

Context-free adaptor grammars (\S\ref{sec:pymcfg:agcfg}) are effective at
modelling concatenative morphology \citep{Johnson2007a,Johnson2008}.
They offer the convenience of being able to capture the essence of phenomena
ranging from single-slot inflection to agglutinative derivation by writing down
a few context-free rewrite rules:
\begin{align}
\nt{Word} &\arrowg ( \nt{Pre}^*\; \nt{Stem}\; \nt{Suf}^*)^+ \\
\nt{Stem} \,|\,\nt{Pre}\,|\,\nt{Suf} &\arrowg \nt{Morph} \\
\nt{Morph} &\arrowg \mathit{(terminal\;strings)} 
\end{align}

Our objective is to be able to specify additional rewrite rules of the
following kind that would capture non-concatenative phenomena as exemplified:
\begin{align}
\label{eq:intercal}
\nt{Stem} &\arrowg \mathbf{intercal}\left(\text{Root}, \text{Template}\right) \\
&\quad\text{e.g. } \text{Arabic derivation \emph{\roottri{k}{t}{b} + i$\cdot$a $\Rightarrow$ kitAb} \iten{book}} \notag
\\
\nt{Stem} &\arrowg \mathbf{infix}\left(\text{Stem}, \text{Infix}\right) \\
&\quad \text{e.g. } \text{Tagalog \emph{sulat} \iten{write} $\Rightarrow$ \emph{s\textbf{um}ulat} \iten{wrote}} \notag
\\
\nt{Stem} &\arrowg \mathbf{circfix}\left(\text{Stem}, \text{Circumfix}\right) \\
&\quad \text{e.g. } \text{Indonesian \emph{percaya} \iten{to trust}}  
       \Rightarrow \text{ \emph{\textbf{ke}percaya\textbf{an}} \iten{belief}} \notag
\end{align}

The bold-faced ``functions'' combine the
potentially discontiguous yields of the argument symbols into single contiguous
strings, e.g. $\mathbf{infix}$(s$\cdot$ulat, um) produces stem \emph{sumulat}.

The question is how to express the multi-argument functions appearing in
the idealisation above in terms of a formal rewrite grammar that can function as
an adaptor grammar.
We propose the use of a \emph{mildly context-sensitive grammar} formalism \citep{joshi1985} that will
allow for a concrete definition of \textbf{intercal} that is a formal rewrite
rule. The more powerful grammar formalism retains the modelling convenience, and
is notably consistent with the universal {\em morphological rule constraint} proposed by
\citet[][p.\ 405]{McCarthy1981}, whereby ``morphological rules must be
context-sensitive rewrite rules affecting no more than one segment at a time,
and no richer type of rule is permitted in the morphology''.
It will allow a discontiguous string like \roottri{k}{t}{b} to be dominated by a
single non-terminal node in a parse tree, which is very useful for probabilistic
modelling.

\section{Simple range concatenating grammars}
\label{sec:pymcfg:srcgs}
We apply \emph{simple Range Concatenating Grammars} (SRCGs)
\cite{Boullier2000journal} to parse contiguous and discontiguous morphemes from
an input string.  These grammars are mildly context-sensitive, forming a superset of context-free grammars that retains 
parsing time-complexity that is polynomial in the input length.\footnote{Our
  formulation is in terms of SRCGs, which are equivalent in power to linear
  context-free rewrite systems \cite{Vijay-ShankerLCFRS} and multiple
  context-free grammars \cite{Seki1991}, all of which are weaker than
  (non-simple) range concatenating grammars \cite{Boullier2000journal}. A good
  overview of these formalisms and their parsing complexity is given by
  \citet{Kallmeyer2010}. }
 An SRCG-rule operates on vectors of ranges over the input, in contrast to the
 way a CFG-rule operates on spans (single ranges).  In this way a non-terminal
 symbol in an SRCG (CFG) derivation can dominate a subset (substring) of
 terminals in an input string.

\subsection{Formalism}
\label{sec:pymcfg:srcg-formalism}
An SRCG \gram{G} is a tuple $(N, T, V, P, S)$, with finite sets of
non-terminals~($N$), terminals~($T$) and variables~($V$), with a start symbol
$S\in N$.  A rewrite rule $p \in P$ of rank $r=\rank(p)\geq0$ has the form
\[
A(\alpha_1,\dots,\alpha_{\fan(A)}) \arrowg \, B_1(\beta_{1,1},\dots,\beta_{1,\fan(B_1)}) \, \dots\, B_r(\beta_{r,1},\dots,\beta_{r,\fan(B_r)}),
\]
where each $\alpha \in (T \cup V)^*$, each $\beta \in V$, and~$\fan(A)$ is the number of
arguments a non-terminal~$A$ has, called its \emph{arity}. By definition, the
start symbol has arity~1.  Any variable $v \in V$ appearing in a given rule must
be used exactly once on each side of the rule.  Terminating rules are written
with~$\epsilon$ as the right-hand side and thus have rank~0.%

A \emph{range} is a pair of integers $(i,j)$ denoting the substring
$w_{i+1}\dots w_{j}$ of a string $w=w_1\dots w_n$.  A non-terminal becomes
\emph{instantiated} when its variables are bound to ranges through substitution.
Variables within an argument imply concatenation and therefore have to bind to
adjacent ranges.

An instantiated non-terminal~$A'$ is said to derive~$\epsilon$ if the
consecutive application of a sequence of instantiated rules rewrite it as~
$\epsilon$.  A string~$w$ is within the language defined by a particular SRCG
iff the start symbol~$S$, instantiated with the exhaustive range $(0,w_n)$,
derives~$\epsilon$.

An important distinction with regard to CFGs is that, due to the instantiation
mechanism, the ordering of non-terminals on the right-hand side of an SRCG rule
is irrelevant, \ie $A(ab) \arrowg B(a) C(b)$ and $A(ab) \arrowg C(b) B(a)$ are
the same rule.\footnote{Certain ordering restrictions over the variables
  \emph{within} an argument  need to hold for an SRCG to indeed be a
  \emph{simple} RCG \cite{Boullier2000journal}.} Consequently, the isomorphisms
of any given SRCG derivation tree all encode the same string, which is uniquely
defined through the instantiation process.

\subsection{Application to root-templatic morphology}
\label{sec:pymcfg:appliedSRCG}
A fragment of the idealised grammar schema from the previous section
(\S\ref{sec:pymcfg:idealised}) can be rephrased as an SRCG by writing the rules
in the newly introduced notation, and supplying a definition of the {\bf
  intercal} function as simply another rule of the grammar:
\begin{align*}
    \nt{Word}(abc) &\arrowg \nt{Pre}(a)\; \nt{Stem}(b)\; \nt{Suf}(c) \\
    \nt{Stem}(abcde) &\arrowg \nt{Root}(a,c,e)\; \nt{Template}(b,d),
\end{align*} 
where $a,b,c,d,e \in V$.
An instantiation of the latter rule with $w=\text{kitAb}$ is
\begin{align*}
    \nt{Stem}(\range{0}{1},\range{1}{2},\range{2}{3},\range{3}{4},\range{4}{5})
        \arrowg%
      & \;     \nt{Root}(\range{0}{1},\range{2}{3},\range{4}{5})\; \\
      & \qquad \nt{Template}(\range{1}{2},\range{3}{4})
\end{align*}

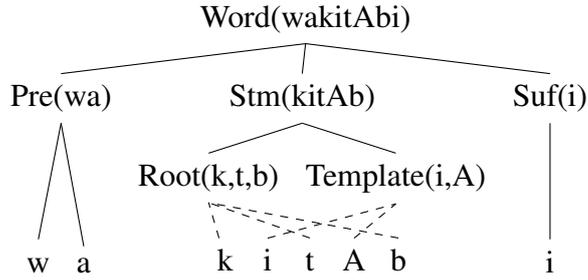
\begin{figure}
\centering
    \input{tex/example_tree}
\caption[SRCG derivation example]%
{Example derivation for \word{wakitAbi}{and my book} using the SRCG
  fragment from~\S\ref{sec:pymcfg:appliedSRCG}. CFGs cannot capture such crossing
  branches.}
\label{fig:mcfg:deriv}
\end{figure}

Given an appropriate set of grammar rules (as we present in
\S\ref{sec:mcfg:concreterules}), we can parse an input string to obtain a tree
as shown in Figure~\ref{fig:mcfg:deriv}.  The overlapping branches of the tree
demonstrate that this grammar captures something a CFG could not.  From the
parse tree one can read off the word's root morpheme and the template used.

\subsection{Time complexity and tractability}
\label{sec:pymcfg:time-complexity-theory}
Although SRCGs specify mildly context-sensitive grammars, each step in a
derivation is context-free---a node's expansion does not depend on other parts
of the tree.  This property implies that a recognition or parsing algorithm can
have a worst-case time complexity that is polynomial in the input length~$n$.
The asymptotic expression is
$\BigO{n^{(\rho+1)\psi}}$ for arity~$\psi$ and rank~$\rho$, which reduces to
$\BigO{n^{3\psi}}$ for a binarised grammar
(\citealp{RodriguezSatta2009a}; \citealp[ch.\ 7]{Kallmeyer2010}); 
the formal basis for this result traces back to analyses by
\citet{Seki1991,Boullier2000journal}, \emph{inter alia}.
We briefly state the intuition. For a binarised grammar, the worst-case occurs
for a rule
 $A(\alpha_1,\dots,\alpha_\psi) \arrowg B(\beta_1,\dots,\beta_\psi)\;
 C(\gamma_1,\dots,\gamma_\psi)$, with each $\alpha_i$ consisting of two variables
 $u_iv_i$. In a bottom-up parser of SRCG \citep{Kallmeyer2010}, $\psi$ ranges,
 hence $2\psi$ indices into the input string, have to be tracked for each
 right-hand side item. The adjacency in the arguments of $A$ implies that the
 end point of the range bound to $u_i$ is the same as the starting point of the
 range bound to $v_i$, so that the total number of \emph{independent} indices
 are $3\psi$, each ranging up to $n$.

To capture the maximal case 
of a stem comprising $k$~discontiguous templatic character segments 
and $k-1$~interspersed root characters 
would require a grammar that has arity $\psi=k$.  For Arabic, which has up to
quadriliteral roots, hence $k\leq5$, the time complexity would be at most
$\BigO{n^{15}}$.  This is a daunting proposition for parsing, but we are careful
to set up our application of SRCGs in such a way that this is not too big an
obstacle:

Firstly, our grammars are defined over the characters that make up a word, and
not over words that make up a sentence.  As such, the input length~$n$ would
tend to be shorter than when parsing full sentences from a corpus.

Secondly, we do type-based morphological analysis, a view supported by evidence
from \citet{Goldwater2006NIPS}, so each unique word in a dataset is only 
parsed once with a given grammar.  The set of word types attested in the data
sources of interest here is fairly limited, typically in the tens of thousands.
For these reasons, our parsing and inference tasks turn out to be tractable
despite the high time-complexity.

As a side remark, note that by refactoring rules the arity and rank can be
balanced in a way that is optimal for parsing time-complexity, as characterised
by \citet{Gildea2010}. It is thus conceivable that the appropriate refactoring
of our grammars could bring down the complexity from what is given above.

\section{Simple range concatenating adaptor grammars}
\label{sec:pymcfg:probSRCG}
We have argued that generative grammars are convenient for
expressing linguistic intuitions of word formation
(\S\ref{sec:pymcfg:idealised}) and demonstrated how SRCGs
can be used to encode root-templatic morphology
(\S\ref{sec:pymcfg:appliedSRCG}). In this section, we
tie together SRCGs and adaptor grammars to obtain a framework capable of
learning non-concatenative morphology in an unsupervised fashion.

\subsection{Probabilistic SRCG}
The probabilistic extension of SRCGs is similar to the probabilistic extension
of CFGs to PCFGs, and has been used in other guises \cite{Kato2006,Maier2010}.  Each rule
$r \in P$ of the SRCG $\mathcal{G_S}$ (as defined in
\S\ref{sec:pymcfg:srcg-formalism}) has an associated probability~$\theta_r$ such that 
$\sum_{r \in P_A} \theta_r=1$.  A random string in the language of the grammar can then be
obtained through a generative procedure that begins with the start symbol~$S$
and iteratively expands it until deriving~$\epsilon$: At each step for some
current symbol~$A$, a rewrite rule~$r$ is sampled randomly from~$P_A$ in
accordance with the distribution over rules and used to expand~$A$.  This
procedure terminates when no further expansions are possible.  Of course,
expansions need to respect the range concatenating and ordering constraints
imposed by the variables in rules.  The expansions imply a chain of variable
bindings going down the tree, and instantiation happens only when rewriting into
$\epsilon$s but then propagates back up the tree.

The probability~$P(w,t)$ of the resulting tree~$t$ and terminal string~$w$ is
the product $\prod_{r}{\theta_r}$ over the sequence of rewrite rules used.  

The invariance of SRCGs trees under isomorphism does not make the probabilistic
model deficient. It adds a source of spurious ambiguity that we pre-empt in
practice by requiring that grammar rules are
specified in a canonical way which ensures a one-to-one correspondence between
the order of nodes in a tree and of terminals in the yield.

\subsection{\ag}
\label{sec:pymcfg:inference-ag}
A Pitman Yor simple Range Concatenating Adaptor Grammar (\ag) is
specified as a tuple $\gram{G} = (\gram{G_S}, M, \mb{a}, \mb{b}, \bm{\alpha})$,
where \gram{G_S} is a probabilistic SRCG, $M$ is the set of adapted
non-terminals and the hyperparameters are analogous to those of the \agcfg
(\S\ref{sec:pymcfg:agcfg}).

The inference procedure under our model is very similar to that of \agcfg{}s,
so we restate the central aspects here but refer the reader
to the original article by \citet{Johnson2007a} for further details.  First,
one may integrate out the adaptors to obtain a single distribution over the set
of trees generated from a particular non-terminal.
Thus, the joint probability of a particular sequence~$\bm{z}$ for the adapted
non-terminal~$A$ with cached counts $(n_1,\dots,n_m)$ is
\begin{equation}
PY(\bm{z}|a,b) = \frac
    {  \prod_{k=1}^m \left(a(k-1) + b\right) \; \prod_{j=1}^{n_k-1}(j-a)  }
    {  \prod_{i=0}^{n-1}(i+b)  }.
\end{equation}
Taking all the adapted non-terminals into account, the joint probability of a
set of full trees~$\bm{T}$ under the grammar~\gram{G} is 
\begin{equation}
P(\bm{T} | \bm{a}, \bm{b}, \bm{\alpha})
 = 
\prod_{A \in M} 
\frac{ \mathrm{B}(\bm{\alpha}_A + \bm{f}_A)} {\mathrm{B}(\bm{\alpha}_A)}
PY(\bm{z(T)} | \bm{a}, \bm{b}), \label{eq:joint_T}
\end{equation}
where~$\bm{f}_A$ is a vector of the usage counts of rules $r\in P_A$
across~$\bm{T}$, and~$\mathrm{B}$ is the Euler beta function.

The posterior distribution over a set of \emph{strings}~$\bm{w}$ is obtained by
marginalising \eqref{eq:joint_T} over all trees that have~$\bm{w}$ as their
yields.  This is intractable to compute directly, so instead we use MCMC
techniques to obtain samples from that posterior using a component-wise
Metropolis-Hastings sampler.\footnote{An alternative to MCMC is to do
  variational inference on top of the stick-breaking representation of the PYP \cite{CohenBleiSmith2010}.}
The sampler works by visiting each string~$w$ in turn and drawing a new tree for
it under a proposal grammar~\gram{G_Q} and randomly accepting that as the new
analysis for~$w$ according to the Metropolis-Hastings accept-reject probability.
As proposal grammar, we use the analogous approximation of our~\gram{G} as
Johnson et al.\ used for PCFGs, namely by taking a static snapshot~\gram{G_Q} of
the adaptor grammar where additional rules rewrite adapted non-terminals as the
terminal strings of their cached trees.  Drawing a sample from the proposal
distribution is then a matter of drawing a random tree from the parse chart of
~$w$ under~$G_Q$. The main distinction of the \ag therefore lies in the
use of an SRCG parser to produce the chart.\footnote{We acknowledge the use of Mark Johnson's publicly available implementation of PY-CFAGs, which we extended as necessary for SRCGs.}

\section{PYSRCAG model for root-templatic morphology}
\label{sec:pymcfg:AG-for-root-template-morphology}
The previous sections have laid the ground-work for learning adaptor grammars
that can cover discontiguous strings. This section
provides the detail for how we apply PYSRCAGs to model Semitic morphology,
covering both its concatenative and root-templatic aspects.

\label{sec:mcfg:concreterules}
\subsection{Words as concatenated morphemes}
We start with a CFG-based adaptor grammar that models words as consisting of a
stem and any number of prefixes and suffixes:\footnote{
    Adapted non-terminals are indicated by underlining and we use the following
    abbreviations: $\nt{X} \arrowg \nt{Y}^+$ means one or more instances of
    \nt{Y} and encodes the rules 
    \mbox{$\nt{X} \arrowg \nt{Ys}$} and
    \mbox{$\nt{Ys} \arrowg \nt{Ys Y} \;\;| \;\; \nt{Y}$}.
    Similarly, \mbox{$\nt{X} \arrowg \nt{Y}^*\; \nt{Z}$}
    allows zero or more instances of \nt{Y} and encodes the rules  \mbox{$\nt{X} \arrowg \nt{Z}$} 
    and  \mbox{$\nt{X} \arrowg \nt{Y}^+ \; \nt{Z}$}.
    Further relabelling is added as necessary to avoid recursion among adapted non-terminals.
}%
\begin{flalign}%
    \nt{Word} &\arrowg \nta{Pre}^*\; \nta{Stem}\; \nta{Suf}^*       \label{eq:RuleCfgWord}  \\
    \nta{Pre} \;|\; \nta{Stem} \;|\; \nta{Suf} &\arrowg \nt{Char}^+ \label{eq:RuleCfgMorphs}
\end{flalign}
This fragment can be seen as building on the stem-and-affix adaptor grammar
presented in \cite{Johnson2007a} for morphological analysis of English, of which
a later version also covers multiple affixes \cite{Sirts2013}.  In the
particular case of Arabic, multiple affixes are required to handle the
attachment of particles and clitics onto base words.

\subsection{Complex stems}
\label{sec:mcfg:complex-stem-rules}
In the preceding grammar, \nt{Stem} is purely a sequence of characters without
further cached sub-structure.  Although there are certain word forms for which
this is an appropriate view, this rule is obviously blind to the reuseable
sub-structures inherent in root morphemes and templates. 

Here, we make the
logical extension of \nt{Stem} to \emph{complex stems}. To establish the
necessary conventions, we start with a rule-set that is appropriate for
modelling triliteral roots:%
\begin{flalign}%
\nta{Stem}(abcdefg) &\arrowg \nta{R3}(b,d,f)\;\nt{T4}(a,c,e,g)\label{eq:RuleT4} \\ 
\nta{Stem}(abcdef)  &\arrowg \nta{R3}(a,c,e)\;\nt{T3}(b,d,f) \label{eq:RuleT3}  \\
\nta{Stem}(abcde)   &\arrowg \nta{R3}(a,c,e)\;\nt{T2}(b,d)   \label{eq:RuleT2}  \\
\nta{Stem}(abcd)    &\arrowg \nta{R3}(a,c,d)\;\nt{T1}(b)     \label{eq:RuleT1}  \\
\nta{Stem}(abc)     &\arrowg \nta{R3}(a,b,c)                 \label{eq:RuleR3collapse}
\end{flalign}
Some of these cases call for additional rules that permute the variables. For
example,
for handling other possible ways of forming a stem from a triliteral
root and a single templatic character, rule~\eqref{eq:RuleT1} also has a
variant $\nta{Stem}(abcd)\arrowg\nta{R3}(a,b,d)\;\nt{T1}(c)$. 
In these and other rule-sets given subsequently, such permuted variants are
suppressed but implied. 

The justification for rule~\eqref{eq:RuleR3collapse}, which may look
trivial, is that it is common in unvocalised text for the stem to
be nothing more than the concatenation of the root characters. 
The inclusion of that rule is thus important for sharing statistical strength of
a given root across all stems it may occur in, including ones that feature no
templatic characters.

Note that although we provide the model with two sets of discontiguous
non-terminals, \nt{R} and \nt{T}, we do not pre-specify their mapping onto 
terminal strings. No subdivision of the alphabet into vowels and consonants is
hard-wired.

To complete the ruleset, a discontiguous non-terminal $\nt{A}n$ is rewritten through recursion on its
arity,\footnote{
Including the arity as part of the non-terminal symbol names forms part of our
convention here to ensure that the grammar contains no cycles, a situation which
would complicate inference under \ag.}%
\begin{align*}
\nt{A}n(v_1,\dots,v_n)  &\arrowg \nt{A}l(v_1,\dots,v_{n-1}) \; \nt{Char}(v_n)
\quad&\text{\em(recursive case; $1<l=n-1$)} \\
\nt{A1}(v) &\arrowg \nt{Char}(v) \quad&\text{\em(base case)}%
\end{align*}
where $v_i \in V$ are variables and \nt{Char} rewrites 
  all terminals $t \in T$ as $\nt{Char}(t) \arrowg \epsilon$.

\section{Experiments}
\label{sec:mcfg:exp}
We evaluate our model on Modern Standard Arabic, Quranic Arabic and Hebrew. These
languages are closely related in their morphology, and feature lexical cognates.
But they are
sufficiently different so that the transferral of rule-based morphological
analysers from one to the other typically require manual intervention.
A key question in this
evaluation is therefore whether an appropriate instantiation of our adaptor
grammars successfully generalises across these related languages.

We evaluate on three tasks: segmentation of a word into a linear sequence of
morphemes, identification of a word's root, and the acquisition of lexica of
stems, affixes and roots.

\subsection{Evaluation data}
\label{sec:mcfg:datagen}
Our models are unsupervised and therefore learn from raw text, but their
evaluation requires annotated data as a gold-standard and these were
derived as follows:
\subsubsection{Arabic (MSA)}
We created the dataset \bw by synthesising 50k morphotactically correct word
types from the morpheme lexicons and consistency rules supplied with the 
Buckwalter Arabic Morphological Analyser \citep[BAMA,][]{Buckwalter2002}.\footnote{
    We used BAMA version 2.0, LDC2004L02,
    and sampled word types having a single stem and at most one prefix, suffix
    or both, according to the following random procedure: Sample a shape
    according to (stem:~0.1, pre+stem:~0.25, stem+suf:~0.25, pre+stem+suf:~0.4).
    Sample uniformly at random (with replacement) a stem from the BAMA stem
    lexicon, and affix(es) from the ones consistent with the chosen stem.  The
    BAMA lexicons contain elementary and compound affixes, so some of the
    generated words would permit a linguistic segmentation into multiple
    prefixes/suffixes.  Nonetheless, we take as gold-standard segmentation
    precisely the items used by our generation procedure.
}
This allowed control over the word shapes, which is important to focus the
evaluation, while yielding reliable segmentation and root annotations.  \bw has
no vocalisation; we denote the corresponding \emph{vocalised} dataset as \bwvoc.
Example words are presented in Table~\ref{tbl:mcfg:example_genwords}.

\begin{table}[tb]
\centering
 \begin{tabular}{l l l}\toprule
   \multicolumn{1}{c}{\bw} & \multicolumn{1}{c}{\bwvoc} & Root
   \\ \midrule
   wlAt\subpre fHm\substem                 &  walAta\subpre foHam\substem   & \roottri{f}{H}{m} \\
   EArDy\substem                           & EAriDiyy\substem & \roottri{E}{r}{D} \\
   w\subpre jbA\substem hm\subsuf          & wa\subpre jabbA\substem hum\subsuf & \roottri{j}{b}{n} \\
   fll\subpre mkAtb\substem & falil\subpre mukAtib\substem & \roottri{k}{t}{b} \\
     \bottomrule
\end{tabular}
\caption[Fragment of simulated Arabic dataset]%
{Example of words in our generated dataset \bw, shown in Buckwalter transliteration.}
\label{tbl:mcfg:example_genwords}
\end{table}

\subsubsection{Quranic Arabic}
We extracted the roughly~18k word types from a morphologically analysed version
of the Quran \cite{DukesHabashQuran2010}.  As an additional challenge, we left
all given diacritics intact for this dataset, \qu.

\subsubsection{Hebrew}
\label{sec:mcfg:hebrew-data}
We leveraged the Hebrew \emph{CHILDES} database as an annotated resource
\cite{HebChildes}, specifically using both the adult and child utterances from the
\emph{Berman} and \emph{Ravid} longitudinal corpora.
5k word types featuring at least one affix each were extracted to form our dataset \heb. 
For words marked as non-standard child language, we used the corrected
forms provided by the database.
Stressed and unstressed vowels were conflated to overcome some inconsistencies in the source data.

\begin{table}[tb]
\centering
\begin{tabular}{l*5{r}} \toprule
     \multicolumn{1}{c}{\textbf{Dataset}}
    & \multicolumn{1}{c}{\textbf{Words}}
    & \multicolumn{1}{c}{\textbf{Stems}}
    & \multicolumn{1}{c}{\textbf{Roots}} 
    & \multicolumn{1}{c}{\textbf{m/w}}
    & \multicolumn{1}{c}{\textbf{c/w}}  \\ \midrule
\bw    & 48428   & 24197 & 4717  & 2.3 & 6.4   \\
\bwvoc & 48428   & 30891 & 4707  & 2.3 & 10.7  \\
\qu    & 18808   & 12021 & 1270  & 1.9 & 9.9 \\
\heb   & 5231    & 3164  & 492   & 2.1 & 6.7 \\ \bottomrule
\end{tabular}
\caption[Corpus statistics]%
{Corpus statistics in number of types, including average number of
  morphemes (m/w) and characters (c/w) per word type. The Roots column gives the
  number of distinct surface-realised roots of length~3 or~4. }
\label{tbl:mcfg:corpus-stats}
\end{table}

\subsection{Model instantiations}
\label{sec:pymcfg:modelspecs}
We use as baselines adaptor grammars the context-free models that express
morphemes as sequences of characters. We consider two variants that match
the known characteristics of our datasets: \cfgone allows only up to one of each
morpheme type in a word, while \cfgmulti allows multiple prefixes and suffixes.

Using these concatenative grammars as starting points, we formulate more
expressive grammars by adding rules that feature discontinuity. The canonical
complex stem grammars aimed at triliteral roots are denoted as \tplonetrilit and
\tplmultitrilit, depending on the starting point grammar.

Informal inspection of the data revealed that cases exist where multiple
characters intervene between root characters. We thus also experiment with a
variant that allows the non-terminal \nt{T1} to be rewritten as up to three
\nt{Char} symbols. This model is denoted \tplonetrilitTtoThree. 

For efficiency reasons, the aforementioned models are limited to rules of arity
at most~3. The grammars \tplonetrilitTFour and \tplquad relax this constraint
to include non-terminal categories \nt{T4} and \nta{R4} \& \nt{T4},
respectively, in order to model quadriliteral roots as well.

These model definitions are given in more detail in \autoref{tbl:pymcfg:modelrules}.

\subsubsection{Sampling details}
Our \ag{}s have various hyperparameters. The PYP hyperparameters $\bm{a}$ and~$\bm{b}$ are modelled by
placing flat $\text{Beta}(1,1)$ and vague $\text{Gamma}(10,0.1)$ priors on
them, respectively.
Their values are then inferred using slice sampling \cite{Neal2003,Johnson2009}.
The hyperparameters $\mb{\alpha}$ are set to give symmetric Dirichlet prior
distributions over the rule probabilities $\theta$.

We start the sampler using batch initialisation, meaning the initial parse tree
of each word is generated purely from the base distribution grammar and is not
influenced by the PYP priors.
\citet{Johnson2009} found this approach to lead to higher posterior probabilities
than initialising incrementally according to the model itself.
Another option explored by that work is to include a type-based element into the
sampler by occasionally resampling the cached trees themselves. This can improve
mixing by updating the analyses of multiple words in a single step.
Experimentation with these variations in initialisation and sampling was not
central to our aim of applying SRCG-based adaptor grammars to root-templatic
morphology, but it is conceivable that they may enable small improvements over our
results for the same grammars we propose.

For each adaptor grammar, we collected 100~posterior samples after allowing
900~iterations of burn-in of the MCMC sampler described in
\S\ref{sec:pymcfg:inference-ag}.
Morphological analyses were obtained from the
collected samples in various ways, as set out per task in the following sections.
To facilitate exposition, let $\mathcal{S}$ denote the $J=100$ collected samples,
where the $j$\th sample 
$\mathcal{S}^{(j)}=\{t_i^{(j)}\}$ 
consists of a parse tree $t_i^{(j)}$ for each word type $w_i$ in the dataset,
  $1 \leq i \leq N$ and $1 \leq j \leq J$.

\begin{table}[tb]
\input{tex/modelrules}

\caption[Model names and definitions]%
{Summary of model identifiers in terms of canonical grammar rules.
  Certain rule shapes involve a permutation in variables, but these are not
  shown explicitly.}
\label{tbl:pymcfg:modelrules}
\end{table}

\subsection{Task 1: Segmentation}
\subsubsection{Method}
\label{sec:pymcfg:method-segmentation}
The segmentation of a word under our adaptor grammars is determined
unambiguously by traversing the parse tree assigned to it.  We split a word into
morphemes along the boundaries of the substrings dominated by the non-terminals
\nt{Stem}, \nt{Pre}, and \nt{Suf}.

We use the maximum \emph{a posteriori} (MAP) parse tree $\tilde{t}(w)$ of a word $w$, which is simply the tree that
is assigned to that word most frequently in the collected posterior
samples.\footnote{An alternative method for obtaining a segmentation from the
  collected posterior samples is to use \emph{maximum marginal decoding}, which
  marginalises out sub-structures which are irrelevant to segmentation
  \cite{Johnson2009}. In our experiments, this approach yielded very similar segmentation results 
  as the MAP-based method, and we thus focus on the latter.}
The joint probability of a word and tree can be factored as
\begin{equation}
  P(w,t\mid \mb{a},\mb{b},\mb{w}) = P(w \mid t)\; P(t \mid \mb{a},\mb{b},
  \mb{w}),
\end{equation}
where the first term is 1 if $t$ yields $w$, and 0 otherwise;
 $\mb{a}$ and $\mb{b}$ are the hyperparameters, and $\mb{w}$ the observed data.
The second term is the posterior distribution over trees
(\cf \autoref{eq:joint_T}) and we can obtain an estimate of the probability of a
tree under this distribution by aggregating over the collected samples
$\mathcal{S}^{(1)}_i, \dots, \mathcal{S}^{(J)}_i$, where $i$ is such that
$w_i=w$.
That is, 
\begin{equation}
  P(t \mid \mb{a},\mb{b}, \mb{w}) \approx \frac{1}{J}\sum\limits_{j=1}^J \delta_{t}(t_i^{j}),
\end{equation}
where $\delta$ is Kronecker-delta.
Thus for a fixed word $w$, the MAP parse tree $\tilde{t}(w)$ is
\begin{align}
 \tilde{t}(w) &= \argmax_t P(w,t \mid \mb{a},\mb{b}, \mb{w}) \\
              &= \argmax_t P(t \mid \mb{a},\mb{b}, \mb{w}) \quad \text{(fixed
                $w$)}\\
              &\approx \argmax_t\sum\limits_{j=1}^J \delta_{t}(t_i^{j}).
              \label{eq:pymcfg:MAP-tree}
\end{align}

Segmentation quality is measured using the \emph{segment border F1-score}
\citep[SBF,][]{Sirts2013}. This is a micro-averaged metric comparing the
word-internal segmentation points of the predicted segmentation against the gold
standard segmentation.\footnote{We acknowledge the use of the evaluation script
  by Sharon Goldwater, obtainable from \\
  \url{http://homepages.inf.ed.ac.uk/sgwater/resources.html}.}

As external baseline model we used \morfessor \cite{Creutz2007}, which performs
decently in unsupervised morphological segmentation of a variety of languages, but only
handles concatenation. 

\subsubsection{Results}
The introduction of complex-stem forming rules consistently brings large gains
in segmentation quality compared to the purely concatenative baseline adaptor
grammars (\autoref{tbl:mcfg:segmentation_bwgen}).

Across our data sets, the simplest templatic grammar featuring triliteral roots
obtains an average relative increase in SBF of~21\% (average
SBF~47.8$\rightarrow$58.0). These gains are not distributed uniformly across
data sets and range from~12\% to~29\%. 
But regarding the question of whether a particular templatic adaptor grammar can
deliver benefits across both Arabic and Hebrew, we conclude that the answer is
in the positive for the segmentation task.

Variation in SBF among the different variants of the triliteral grammars is
small and mixed across data sets. 

The vocalised Arabic data set \bwvoc is most sensitive to changes in grammar
expressivity. Introduction of the~\nt{T4} non-terminal to the basic
triliteral grammar increases SBF on \bwvoc by
4~points, on top of which the most expressive grammar, \tplquad, adds
another 10~points. This amounts to a substantial total improvement of~57\% by
\tplquad over the baseline \cfgone.

On the unvocalised \bw, grammar expressivity beyond \tplonetrilit brings smaller
gains.  This is consistent with the fact that \bw represents an easier
segmentation task, since its unvocalised words are by definition shorter, on
average, than those in \bwvoc, and cover fewer distinct contiguous morphemes.

When repeating the experiment for a
    version of the Hebrew training data that
    \emph{includes} affixless words, SBF drops to the fifties. The overall trend
    that the templatic grammars improve segmentation remains, and the relative
    improvement is slightly larger in this setting.

A salient aspect of these results, when considered in conjunction with the root
identification results to be presented in
\S\ref{sec:pymcfg:results-root-ident}, is that templatic rules aid
segmentation quality independent of root identification accuracy. 
In other words, our attempt at capturing discontiguous roots and templates
benefits segmentation quality by allowing the model to capture discontiguous
regularities at the sub-stem level, which is advantageous regardless of whether
the captured elements actually correspond to the linguistically motivated
analyses.

\begin{table}[bt]%
\centering
\input{tex/segmentation-results}

\caption[Evaluation of segmentation quality]%
{
  Segmentation quality in segment-border F-score (SBF). 
}
\label{tbl:mcfg:segmentation_bwgen}
\end{table}

\subsubsection{Further analysis of Quran segmentation}
\label{sec:pymcfg:analysis-quran-segmentation}
The trend that templatic rules improved segmentation quality holds for the Quran
data set (improvements of 15\%~to 31\% relative to the baseline adaptor grammar),
but absolute SBF for the adaptor grammars is disappointingly low for this data
set.

\begin{figure}[tb]
\centering
\includegraphics[width=0.95\textwidth]{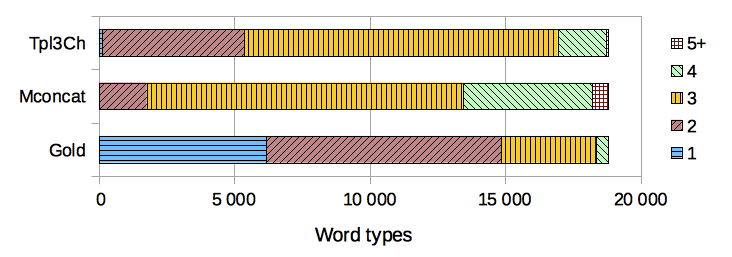}
\caption[Word length distributions]%
{Distribution of number of morphemes per word for data set \qu, showing
  the large discrepancy between the gold standard and the induced
  segmentations.}
\label{fig:pymcfg:word-length-distribution}
\end{figure}

Our analysis shows that all the evaluated adaptor grammars over-segmented
heavily. As an illustration,
the gold standard has on average 1.9~morphemes per
word~(m/w, \autoref{tbl:mcfg:corpus-stats}), while the models hypothesise 2.8--3.2.
The distribution of~m/w is very different. The gold standard is dominated by
words having one or two morphemes, giving a skewed distribution of~m/w,
while the adaptor grammars induce a more symmetric distribution where length~three
words dominate (\autoref{fig:pymcfg:word-length-distribution}).
If we evaluate only on the subset of words that have 3~or more morphemes
according to the gold standard, segmentation quality is higher (\cfgmulti:~32.7,
\tplmultitrilitTtoThree:~38.6).

We investigated two hypotheses that could explain the low SBF scores on this
data set. The first hypothesis is that the outcome is an artefact of our
methodology for extracting segmentations from parse trees. As detailled in
\S\ref{sec:pymcfg:method-segmentation}, we split words maximally along the
elementary affix categories, \nt{Pre} and~\nt{Suf} even though the grammars used
for \qu allow composite affixes. In \autoref{tbl:pymcfg:coarse-by-fine-grid}, we
designate this as the fine-grained grammar evaluated at the fine-grained level.
An alternative is to split at the boundaries of the \nt{Stem} category only and
to ignore the internal structure of composite affixes.
That is, evaluate the fine-grained grammar in a coarse-grained fashion.
Finally, we can train a coarse-grained grammar~(\cfgone) instead of a
fine-grained one~(\cfgmulti), and evaluate coarsely.

The outcome of this two-way comparison, using only the concatenative baseline
adaptor grammars for efficiency, is that all three valid combinations give
low SBF scores (\autoref{tbl:pymcfg:coarse-by-fine-grid}). We thus reject the hypothesis that these methodological
choices by themselves explain the below-baseline performance of the adaptor grammars on~\qu.

\begin{table}[bt]
  \centering
\begin{tabular}{lrr}
  & \multicolumn{2}{c}{Evaluation } \\ 
                           & \multicolumn{1}{c}{Coarse} & \multicolumn{1}{c}{Fine} \\ \cmidrule{2-3}
  \cfgone (coarse grammar) & $13.2$                     & n/a                      \\
  \cfgmulti (fine grammar) & $9.6$                      & $^\dagger19.6$   
\end{tabular}
\caption[Detailled evaluation of segmentation quality on Quranic data]%
{Effect of granularity choices in grammar design and evaluation when
  measuring segmentation quality on data set \qu.
  These results are for the strictly concatenative adaptor grammars. See
  \S\ref{sec:pymcfg:analysis-quran-segmentation} for details on ``coarse''
  versus ``fine'' evaluation. $^\dagger19.6$ is highlighted as being the
  combination presented in \autoref{tbl:mcfg:segmentation_bwgen}.}
\label{tbl:pymcfg:coarse-by-fine-grid}
\end{table}

The second hypothesis we considered is that performance suffered
because of a lack of hierarchical adaptation in the multi-affix adaptor
grammars.\footnote{\citet{Sirts2013} demonstrated that hierarchical adaptation can play
  a crucial role.}
From \autoref{tbl:pymcfg:modelrules}, the fundamental word-forming rule for the multi-affix grammars is
\[\nt{Word}\arrowg\nt{Pres}\;\nta{Stem}\;\nt{Sufs},\] where \nt{Pres} and \nt{Sufs}
are \emph{not} adapted (although \nta{Pre} and \nta{Suf} were adapted).
We added adaptation to these categories, also introducing intermediate dummy rules to
avoid ill-defined recursive adaptor grammar rules like $\nta{Pres} \arrowg
\nta{Pres}\;\nta{Pre}$. The segmentations induced by the resulting
hierarchically adapted, concatenative grammar obtain an SBF of~22.2,
marginally above the \cfgmulti score of~19.6. 
The initial lack of hierarchical adaptation therefore affected segmentation
quality adversely, but not sufficiently to fully explain 
the low segmentation quality on the Quran data set.\footnote{The adaptation of
  affix non-terminals is nonetheless crucial. In preliminary experiments on \bw,
  segmentation quality dropped by about~50\% when using unadapted non-terminal
  categories \nt{Pre} and \nt{Suf}.}

Given the low segmentation quality on the Quran, it is not included in the
subsequent tasks.

\subsection{Task 2: Root identification}
\label{sec:pymcfg:root-ident-task-intro}
The next evaluation task is that of obtaining the root morpheme, if any, given a
surface word type as input. We describe two methods for root
identification, followed by details of the baselines we compare
to.

\subsubsection{Root extraction method}
\label{sec:pymcfg:root-ident-method}
Our root-templatic adaptor grammars are fundamentally unsupervised. The only
built-in bias they have for capturing roots under the \nt{R}~non-terminals
instead of the \nt{T}~non-terminals is that we only adapted the former, in
particular \nta{R3} and \nta{R4}, as set out in \autoref{tbl:pymcfg:modelrules}.
Root extraction \emph{Method~A} thus hypothesises as root morpheme
the characters governed by \nta{R3} or \nta{R4} in the parse tree, depending on
whether the evaluation is on triliterals or quadriliterals.
A shortcoming of this method is that it would miss cases where the root happens
to be governed by a \nt{T} non-terminal, which is also licensed. The more
lenient extraction \emph{Method~B} thus considers both types of non-terminals,
possibly hypothesising two root candidates for a word.
As with the segmentation task, we use the MAP~parse tree $\tilde{t}(w)$ (cf.\ 
\autoref{eq:pymcfg:MAP-tree}) as input to these root
extraction methods.\footnote{The posterior distribution over
  roots conditioned on a word form, as estimated from the collected samples, is sharply
  peaked and in most cases uni-modal, so that a MAP estimate is reasonable.}
\autoref{tbl:mcfg:extract-example} provides an example of these root extraction
methods, and illustrates the evaluation calculation we define below.

Given an input word $w$, we denote the set of $k$-literal roots hypothesised
as 
$\tilde{h}_{k}\!\left(\tilde{t}(w)\right)
\in
\left\{ \emptyset, \{r^h_1\}, \{r^h_1, r^h_2\}\right\}$.
We focus on $k\in\{3,4\}$.

\input{tex/extraction-example}

Similarly, we write the set of correct roots of the word $w$ as $g_{k}(w)=\{r^g_j\}$.
Roots that contain weak radicals are filtered out, since our models deal purely with
surface-realised phenomena and thus cannot express such roots. Owing to this
filtering and the fact that some word forms legitimately do not derive from a
root, $g_{k}(w)$ can be empty. Our evaluation skips such words.

The evaluation metric is exact match accuracy.
A hypothesised root is counted as a match if it appears among
the word's gold roots. When multiple roots are hypothesised we check if any one
matches a gold root.
To be precise, define the exact match score between hypothesised
and gold standard roots for a word~$w$ as 
\begin{equation}
\mathrm{exact\_match}_k(w) = 
\begin{cases}
  1 & \text{if } \tilde{h}_{k}\!\left(\tilde{t}(w)\right) \cap g_{k}(w) \neq \emptyset \\
  0 & \text{otherwise.}
\end{cases}
\end{equation}
The accuracy across the $N'$ filtered test word types is
$\frac{1}{N'}\sum_{i=1}^{N'} \mathrm{exact\_match}_k(w_i)$.\footnote{These definitions are aimed at precise exposition,
  belying the fact that the ambiguity observed in our evaluation is very low;
  the vast majority of words have a single gold root, if any, and for
  hypothesised roots it is only Method~B that occasionally posits more than one
  candidate.}

This metric is therefore recall-based. 
A precision-based evaluation would be overly
harsh, since our root identification methods posit a root for most words.
That is, the templatic adaptor grammars tend to
rely heavily on the complex stem-forming rules, despite the presence of the
simple rule, $\nta{Stem}\arrowg\nt{Char}^+$.

\subsubsection{Baselines}
\label{sec:pymcfg:root-ident-baseline-methods}
We devised three straightforward randomised baseline methods to
compare our model-based prediction accuracies.
They all involve incremental deletion of characters from a word, chosen
uniformly at random, until a string of $k$ characters remain.
The {\bf uniform guesser} baseline applies this reduction method directly to the surface word
form, while three further baselines operate on a constrained guessing space:
\begin{description}
  \item[Stem-constrained:] The gold standard stem of a word is taken as input,
    thereby establishing the difficulty of identifying a root under a perfect
    segmentation model.
  \item[Character-constrained:] Starts with the surface word form, but strips
    out characters never appearing as roots in the data set. This baseline thus
    has external knowledge not encoded in our models, but has no sense of word
    structure.
  \item[Doubly-constrained:] Uses both aforementioned constraints. This would
    often leave little ambiguity as to the identity of the root, and could be
    regarded as a benchmark rather than baseline.
\end{description}
Given the non-determinism of these baselines, we report the average of five
repetitions of each.

\subsubsection{Results}
\label{sec:pymcfg:results-root-ident}
The results for root identification are given in
\autoref{tbl:pymcfg:root-ident-tri} and \autoref{tbl:pymcfg:root-ident-quad}.

The performance of the baseline methods confirms the intuition that root
identification is harder in the presence of vocalisation.\footnote{Hebrew
  morphology and orthography are less complicated than Arabic
  \cite{Daya2008}, which
  is why the baseline accuracies for the vocalised \heb are generally higher
  than for \bwvoc.}
It is consequently also for vocalised Arabic (\bwvoc) that identification Method~A performs poorly.
The more liberal Method~B succeeds in correctly identifying the triliteral root
for up to one in five word types in \bwvoc that meet the evaluation conditions
specified in \S\ref{sec:pymcfg:root-ident-method}.
This is still low accuracy, but exceeds the character-constrained baseline.
Since that baseline gauges the extent to which vocalisation
complicates root identification, we conclude from this outcome
that our models go beyond merely overcoming the presence of
vocalisation. The difference is explained by the model's segmentational
component. Nonetheless, the regularities leveraged in the process do not
necessarily coincide with the linguistically correct roots.

\begin{table}[tb]
\centering
\input{tex/root-ident-results}
\caption[Triliteral root identification accuracy]%
{Triliteral root identification accuracy. Exact matches of extracted
  and gold roots expressed as a percentage of the number of word types $N'$ for which the
  true analysis contains a strong triliteral root.
}
\label{tbl:pymcfg:root-ident-tri}
\end{table}

\begin{table}[tb]
\centering
\input{tex/root-ident-results-quad}
\caption[Quadriliteral root identification accuracy]%
{Quadriliteral root identification accuracy. Exact matches of extracted
  and gold roots expressed as a percentage of the number of word types $N'$ for which the
  true analysis contains a strong quadriliteral root.
}
\label{tbl:pymcfg:root-ident-quad}
\end{table}

On Hebrew and unvocalised Arabic, there is no meaningful difference between
Methods~A and~B.

For triliteral root identification, the templatic grammar \tplonetrilitTFour
that includes four-character templates performs best, with accuracies of 74\% on
Hebrew and 67\% on unvocalised Arabic. In the latter data set there are about~
9000 words for which the stem and root are equivalent; excluding
those words from the evaluation decreases the accuracy from 67\% to 57\%
accuracy. This is still substantially above the competitive Doubly constrained
baseline.
The fact that \tplonetrilitTFour outperforms \tplonetrilitTtoThree shows that, for
these two data sets, it is more important to capture templatic structures of the
form VCVCVCV than to handle forms like CVVCVC.
On Hebrew and unvocalised Arabic, this adaptor grammar jointly solves
the task of segmentation and root identification to a fairly high standard.

Our model capable of quadrilteral root identification, \tplquad, obtained
accuracies in the region of the Character-constrained baseline for the two
Arabic data sets. Closer analysis shows that this low accuracy is not due to worse
segmentation performance on those words containing quadriliteral roots. Instead,
we find that even if the model segments a word correctly, it often accounts for the
stem sufficiently using the R3 or T3 categories, so that our method extracts no
quadriliteral root. 
For cases where a quadriliteral root is hypothesised by our
extraction methods, we find that they often match the gold root partially. In
this sense, the exact match metric is relatively harsher than for evaluating
triliteral root extraction.

In summary, these results show that a given instantiation of our root-templatic
adaptor grammars can correctly identify triliteral and quadriliteral roots of
a small portion of words, but
that more refined grammar engineering would be needed to increase the
accuracy.
Example analyses are provided in \autoref{fig:mcfg:tree} on
p.~\pageref{fig:mcfg:tree}.

\begin{figure}[p!] 
\centering %
\fbox{%
\begin{minipage}[b]{.9\textwidth}
    \input{tex/trees}
    \subcaption{Unvocalised Arabic input word ``wl$>$stArkm''. \\ Reference analysis:
      wl\segTP{}$>$\HGRT{s}\HGRT{t}A\HGRT{r}\segTP{}km
    }
    \label{fig:mcfg:tree-bw}
\end{minipage}%
}%
\vspace{1ex}

\fbox{%
\begin{minipage}[b]{.90\textwidth}
    \input{tex/trees_bwvoc}
    \subcaption{Vocalised Arabic input word ``lidanotillA''. Reference analysis:
    li\segTP{}\HGRT{d}a\HGRT{n}o\HGRT{t}i\HGRT{l}lA
    }

    \label{fig:mcfg:tree-bwvoc}    
\end{minipage}%
}%
\caption[Examples of induced parse trees]%
{Parse tree examples for a word from the Arabic data sets \bw and \bwvoc
  where the concatenative adaptor grammar segmented incorrectly (left).  \\
The templatic grammars (right)
  correctly identified the triliteral and quadriliteral roots, and fixed the
  segmentation of (a).  In (b), the templatic grammar improved over the baseline
  by finding the correct prefix but falsely posited a suffix. \\
  (Unimportant
  subtrees are elided for space, while the yields of discontiguous constituents
  are indicated next to their symbols, with dots marking gaps.  Crossing
  branches are not drawn but should be inferrable. Root characters are
  bold-faced in the reference analyses.  The non-terminal \nt{X2} in
  (a) is part of a number of  implementation-specific helper rules that ensure
  the appropriate handling of partly contiguous roots.)  }
\label{fig:mcfg:tree}
\end{figure}

\subsection{Task 3: Morpheme lexicon induction}
\label{sec:pymcfg:lexicon-task}
The previous two tasks considered the challenging setting of predicting a single
correct segmentation and root for a given word type.
In this section, we turn to another dimension of morphology learning by
evaluating the morpheme lexica inferred by the adaptor grammars.  By morpheme
lexica we mean the sets of strings identified by a model as being members of a
given morphological category.\footnote{For simplicity, we use the term
  ``string'' in this section to mean both contiguous character sequences as well
  as non-contiguous tuples of characters.}  The ability to derive such lexica
from unannotated
data can be useful for the construction of linguistic resources in low-resource
or undocumented languages, and is relevant in computational studies of language
acquisition \cite{DeMarckenThesis1996,GoldwaterThesis2007}.

Purely concatenative models of morphology can only infer lexica of prefixes,
suffixes and stems. Our approach to modelling explicitly the transfixal
formation of Semitic stems thus have two potential benefits: it can improve the
quality of the aforementioned lexica, but it also enables the acquisition of the
lexicon of root morphemes. The aim here is to test the extent to which these two
potential benefits are borne out by the SRCG-based adaptor grammars on our
Hebrew and Arabic data sets.

We evaluate two methods of obtaining morpheme lexica from the adaptor
grammar posterior samples:
\begin{enumerate}
  \item use the MAP~parse of each word type as before
    (\S\ref{sec:pymcfg:method-segmentation}), but aggregate morphemes across
    word types; 
  \item marginalise directly over both word types and posterior samples to induce
    probabilistic lexica.
\end{enumerate} 

\subsubsection{MAP-based morpheme lexica}
\label{sec:pymcfg:lexicon-task-MAP}
We take the MAP~parse trees under a given adaptor grammar as starting point, and
predict the lexicon of a particular morphological category (stem, affix, root,
etc.) as the unique set of strings appearing under that category across the word
types in the data set.\footnote{A given string can therefore appear in multiple
  lexica, which fits the linguistic reality.} Prefixes, stems and suffixes are
identified using the same method of traversing the parse trees as in the
segmentation task, and we discuss the root lexicon shortly.

The evaluation metric is the F-score when comparing a hypothesised lexicon
against the corresponding gold lexicon.

The primary result is that, consistent with our expectations, the models capable
of forming complex stems obtain a marked improvement in prefix, stem and suffix
lexicon F-score compared to the baseline concatenative adaptor grammar
(\autoref{tbl:mcfg:lexscores}).
The amount of improvement grows with the expressivity of the templatic
grammar, most pronouncedly for the vocalised Arabic data set.

For unvocalised Arabic, the templatic grammars improve over the baseline by
recovering more of the correct stems---recall goes from 32\%~under \cfgone to
47\%~under \tplquad while precision remains at 74\%. This equates to an
additional 3673~correctly inferred stems.
On vocalised Arabic and Hebrew, the stem F-score improvement is much more substantial
(roughly 20\%~to~60\%) and arises from increases in both precision and recall;
precision reaches about 70\% and recall 54\% for these two vocalised data sets
when using \tplquad.
This grammar thus induces stem lexica at quite high precision for three different
data sets.

In contrast to stems, the affix lexica are very noisy. The concatenative models
tend to over-generate to obtain very high recall but low precision. The
templatic grammars balance out that trade-off to some extent.

To obtain a root lexicon based on the MAP derivations for each data set we
leveraged the outcome of the earlier root identification experiment
(\S\ref{sec:pymcfg:results-root-ident}).
For \bw and \heb, we take the R3 yields, while for \bwvoc we take the T3 yields.
Variation among the different templatic grammars was negligible, so we report
the results only for \tplonetrilit in \autoref{tbl:pymcfg:root-lex-prf}.
The lexica of roots identified for the two Arabic data sets in this fashion contained just over
50\% legitimate roots. The high recall of 80\% on the unvocalised
version equates to recovering 2602 of the 3254 surface-realised triliteral roots
present in that data set.

\begin{table}[tb]
\centering
\begin{tabular}{l rrr r@{\hspace{1em}}    rrr r@{\hspace{1em}}    rrr} \toprule
& \multicolumn{3}{c}{\bwvoc} &  & 
  \multicolumn{3}{c}{\bw}    &  & 
  \multicolumn{3}{c}{\heb}      \\ 
\cmidrule{3-3} \cmidrule{7-7} \cmidrule{11-11} 
& \multicolumn{1}{c}{Pre} & \multicolumn{1}{c}{Stem} & \multicolumn{1}{c}{Suf} &  &
  \multicolumn{1}{c}{Pre} & \multicolumn{1}{c}{Stem} & \multicolumn{1}{c}{Suf} &  &
  \multicolumn{1}{c}{Pre} & \multicolumn{1}{c}{Stem} & \multicolumn{1}{c}{Suf}  \\
\cmidrule(lr){2-4} \cmidrule(lr){6-8} \cmidrule(lr){10-12} 
\cfgone               & 15.0 & 20.2 & 25.4 &  & 32.8 & 44.1 & 40.3 &  & 18.7 & 20.9 & 29.2 \\
\tplonetrilit         & 24.7 & 39.4 & 35.2 &  & 45.9 & 54.7 & 47.9 &  & 35.1 & 59.6 & 52.9 \\
\tplquad              & 37.8 & 60.3 & 47.0 &  & 53.0 & 57.7 & 51.9 &  & 38.0 & 62.4 & 55.2 \\
\bottomrule
\end{tabular}
\caption[Morpheme lexicon induction quality]%
{Morpheme lexicon induction quality. F1-scores for lexicons induced from
  the maximum \emph{a posteriori} parse of each different dataset.
}
\label{tbl:mcfg:lexscores}
\end{table}

\begin{table}[tb]
  \centering
  \begin{tabular}{lccc} \toprule
      & P    & R    & F \\ \cmidrule{2-4}
\bwvoc (T3) & 54.9 & 47.9 & 51.2 \\
\bw    (R3) & 51.6 & 80.0 & 62.7 \\
\heb   (R3) & 25.3 & 55.3 & 34.8 \\ \bottomrule
  \end{tabular}
  \caption[Evaluation of induced root lexica]%
{Precision, recall and F-score for root lexicon induction using the
    grammar \tplonetrilit.}
  \label{tbl:pymcfg:root-lex-prf}
\end{table}

\subsubsection{Probabilistic morpheme lexica}
\label{sec:pymcfg:lexicon-task-prob}
The preceding set-based evaluation of the morpheme lexica imposes hard decisions
about category membership and does not make use of the fact that
the adaptor grammars induce a probability distribution over strings.
In this section we measure morpheme lexicon induction performance by considering
the probability with which a string falls under a particular non-terminal
category in an adaptor grammar.

From our posterior samples we can estimate the joint probability of a string $s$
appearing under grammar category $C$ by summing over all word types and
posterior samples:
\[
  P(s, C)
  \approx 
  \frac{1}{JN}
    \sum\limits_{j=1}^J\sum\limits_{i=1}^N 
    \begin{cases}%
      1 & \text{ if non-terminal \nt{C} dominates $s$ in analysis } t_i^{(j)} \\ 
      0 & \text{ otherwise.}
    \end{cases}
\]
We can thus produce a list of roots ranked by their joint probability values
under a given adaptor grammar. As an example, we present in
\autoref{fig:mcfg:heb_samples} the top five triliteral roots obtained from the
Hebrew data set in this fashion. The first four of these correspond to correct
Hebrew roots, demonstrating that the SRCG-based adaptor grammars can
learn linguistically relevant, discontiguous entities from unannotated data.
In the same table, we present some example usages of the roots. As a concrete
example of how the model successfully encodes the original intuition that
discontiguous morphemes can be shared across word forms, consider that it
correctly associates the root \iteg{lbS} with such orthograpically distinct word forms
as \iteg{labaSt}, \iteg{tilbeSi} and \iteg{lehalbiS}, which are not linked by a shared stem.

\paragraph{ROC analysis}
A snapshot of the top-ranked items does not quantify the quality of a morpheme
lexicon very broadly. But it illustrates a simple sense in which we can combine
a decision rule with an adaptor grammar to obtain a discrete classifier of
strings.  Instead of the decision rule ``top $n$ items from ranked list'', we
can also base the binary decision of whether a particular string $s$ is part of
a morphological category $C$ directly on the probability $P(s|C)$ using a
threshold $\tau$ between 0 and 1.  Let a predicted lexicon $L_\tau^C$ be the set
of strings for which $P(s|C) \geq \tau$.

This allows using receiver operating characteristics (ROC) analysis
\cite{Egan75,Fawcett2006} to characterise the trade-off between true positive
and false positive predictions across different decision thresholds $\tau$.  A
ROC plot is obtained by plotting the number of true positives (\eg true stems)
versus the number of false positives (\eg false stems) for $L_\tau^C$ as $\tau$
is varied from 0 to 1.  A coin-tossing random baseline classifier traces a
diagonal line ($y=x$) through the origin. Better classifiers give rise to points
in ROC space toward the upper left of the plot. Finally, the area under a curve
(AUC) on the ROC plot gives a summary of the general predictive performance of a
model. Because the ROC space is the unit square, AUC is a number between 0 and 1.
AUC is equivalent to the probability that a model would rank a randomly chosen
positive instance higher than a randomly chosen negative instance
\cite{Fawcett2006}.  ROC plots and AUC are non-parametric methods, and therefore
give robust results insensitive to internal differences between models or skews
in the number of true versus negative instances.

\begin{figure}[bt]%
\centering
\begin{minipage}[b]{0.42\textwidth}
\centering
\includegraphics[width=\textwidth]{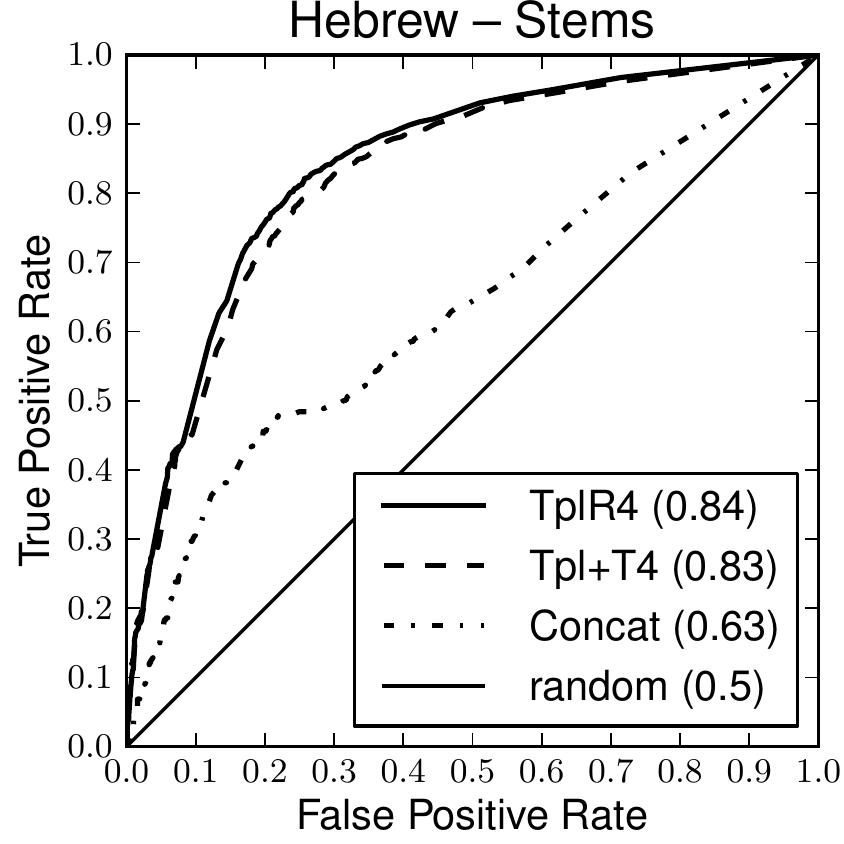}
\vspace{1ex}
\end{minipage} %
\hspace{0.1\textwidth}
\begin{minipage}[b]{0.42\textwidth}
\centering
\includegraphics[width=\textwidth]{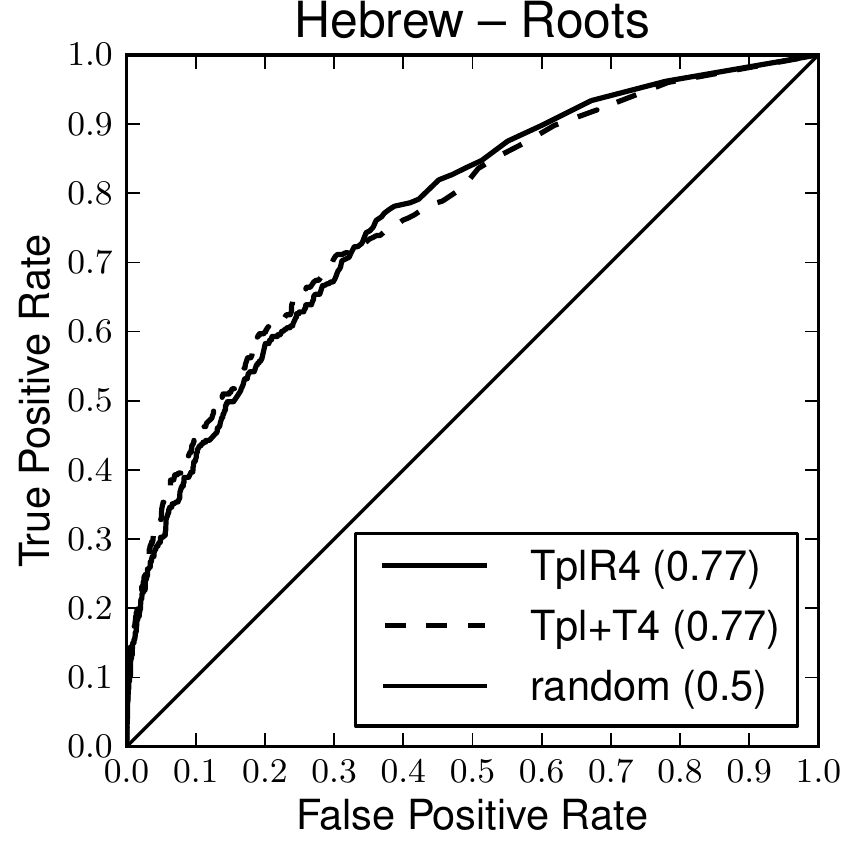}
\vspace{1ex}
\end{minipage} %
\begin{minipage}[b]{0.42\textwidth}
\centering
\includegraphics[width=\textwidth]{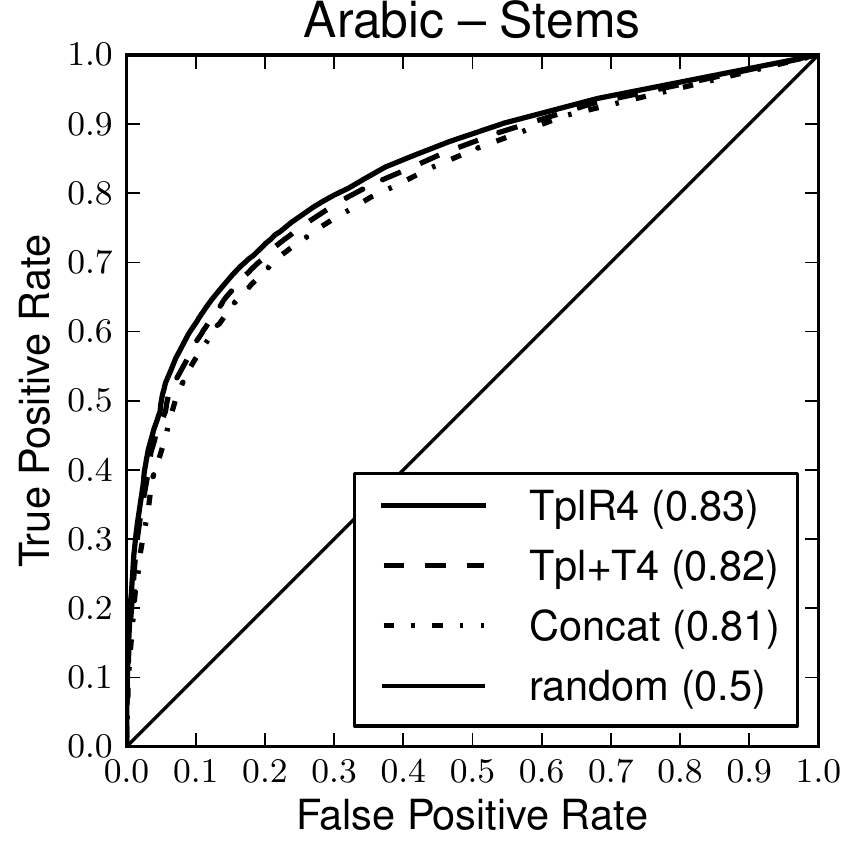}
\vspace{1ex}
\end{minipage} %
\hspace{0.1\textwidth}
\begin{minipage}[b]{0.42\textwidth}
\centering
\includegraphics[width=\textwidth]{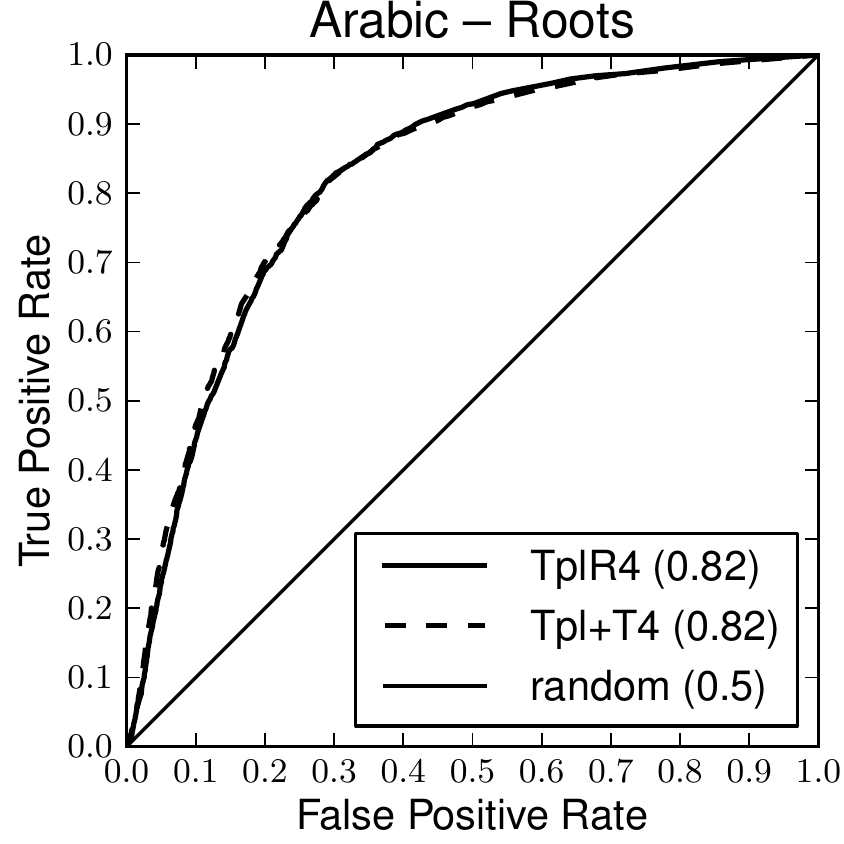}
\vspace{1ex}
\end{minipage} %
\begin{minipage}[b]{0.42\textwidth}
\centering
\includegraphics[width=\textwidth]{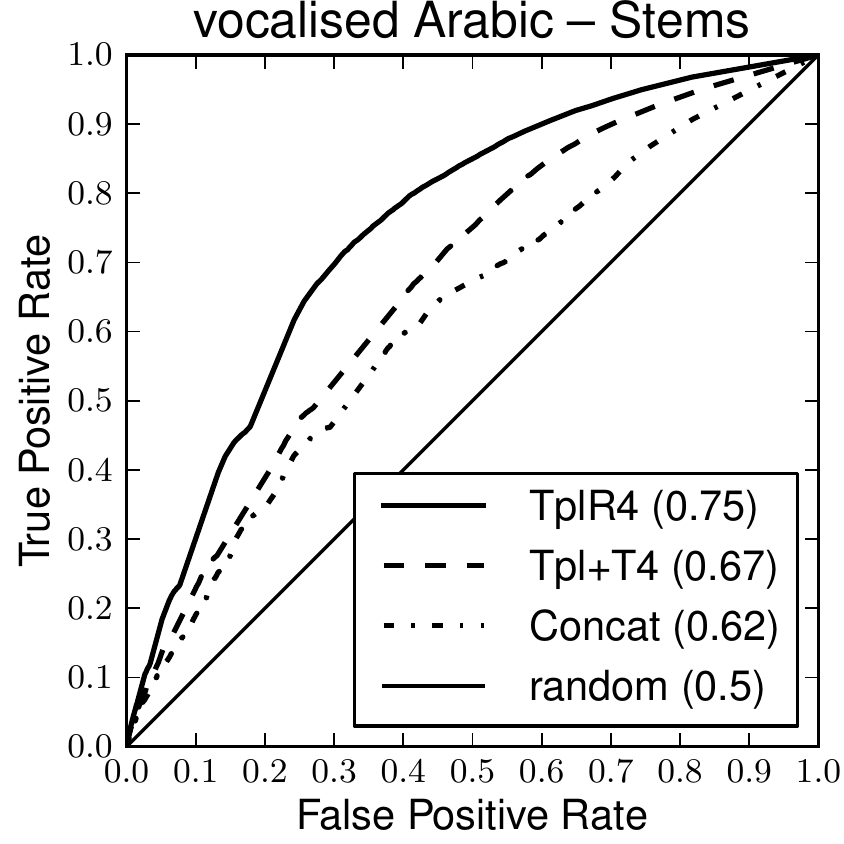}
\end{minipage} %
\hspace{0.1\textwidth}
\begin{minipage}[b]{0.42\textwidth}
\centering
\includegraphics[width=\textwidth]{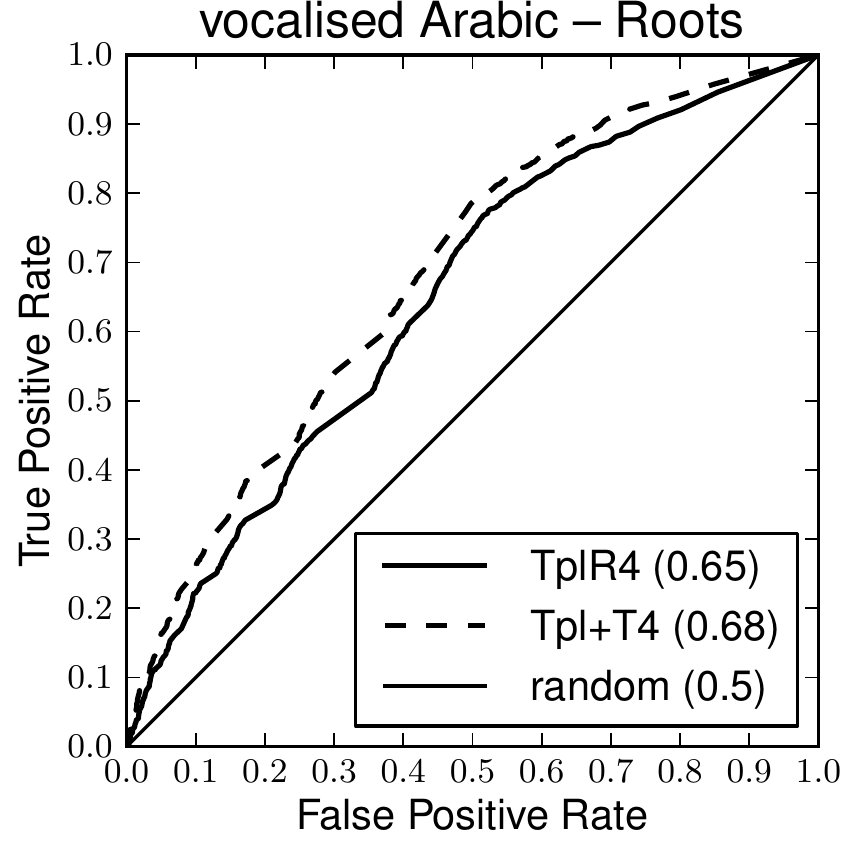}
\end{minipage} %
\caption[ROC curves for stem and root lexica]%
{ROC curves for predicting the stem lexicon (left) and root lexicon
  (right) for the Hebrew and Arabic data sets, as described in
  \S\ref{sec:pymcfg:lexicon-task-prob}.  The area under each curve (AUC) is
  given in parentheses, as computed with the trapezium rule.
}
\label{fig:mcfg:roc}
\end{figure}

\autoref{fig:mcfg:roc} shows the ROC plots and AUC values for inducing stem and
root lexica from our data sets using different adaptor grammars.
The first main outcome is that the prediction of stems benefits from the
use of the root-templatic SRCG rules, consistent with the results in the
MAP-based evaluation. This effect is again most pronounced for the vocalised
data sets \bwvoc and \heb. The second main result is that the prediction of root
lexica for \heb and \bw is substantially better than the random baseline: the
SRCG-based models would rank a true root above a non-root string with
probability 0.77 and 0.82 on these respective data sets. For
 the vocalised \bwvoc, the corresponding value is 0.68. As a means for
 predicting whether or not particular strings are root morphemes of the
 language, our method thus performs better than might be expected from the low
 root-identification results in the per-word analysis in
 \S\ref{sec:pymcfg:results-root-ident}.

\begin{table}[bth]
\centering
\input{tex/heb_samples}
\caption[Analysis of top triliteral Hebrew roots]%
{Top five Hebrew triliteral roots hypothesised by the grammar \tplonetrilitTFour as 
    ranked by model probability, along with examples of their occurrence in the
    analysed words. The first four are legitimate roots, while the fifth is not a
    valid root.\\
    The middle two columns give examples of words where the model found the
    correct segmentation points (\segTP) and the predicted root characters coincide with the true root
    characters (\HGRT{bold}).
    The penultimate column shows examples where the root is found despite
    segmentation mistakes, marking false positive (\segFP{}) and 
    false negative (\segFN{}) segmentation points.
    Examples in the final column feature segmentation and root mistakes---false
    positive root characters appear in \HRT{grey}, while false negatives 
    are \GRT{boxed}.
    }
\label{fig:mcfg:heb_samples}
\end{table}

\clearpage
\section{Summary}
\label{sec:pymcfg:summary}
This chapter introduced a new approach to the unsupervised learning of
non\=/concatenative morphology. We showed how the simple Range Concatenating Grammar
formalism can be used to encode non-concatenative morphological processes which
do not involve character deletion or insertion when composing morphemes into a
surface word form. This approach can account for infixation and circumfixation,
but our focus was to instantiate and evaluate the general idea on the
root-template-based morphology employed in Semitic languages.  For a learning
framework, we made the natural extension of non-parametric Bayesian adaptor
grammars to the SRCG formalism, which makes it possible to model explicitly the
tendency of morphemes (contiguous and discontiguous ones) to be reused in
different words. Aside from our application to morphology, this learning
approach could also be useful in other applications of SRCGs.

Our experimentation on Hebrew, Quranic Arabic and two versions of a synthetic
data set containing morphotactically valid standard Arabic words shows that
modelling discontiguous sub-structures of words improves the segmentation of
words into morphemes.
A simple root identification method correctly identified
triliteral roots for about two-thirds of the words where they are expected, for
two of the aforementioned data sets. On vocalised Arabic, absolute performance
across the different tasks was generally lower, and very low for the Quran.
Finally, we found that our SRCG-based models could be used to infer a lexicon of
triliteral root morphemes.
This mixed outcome flows from the fact that we deliberately sought to limit the
amount of language-specific knowledge built into the grammars in order to test
to what extent a high-level notion of intercalated morphology is beneficial in
these kind of unsupervised models. For the Hebrew and synthetic Arabic data
sets, this was balanced against exercising more control over the word shapes we
included in the evaluation, whereas for the Quran we applied no informed
preprocessing or standardisation whatsoever.

%% file: ch-pysrcag/tex/example_tree.tex
	\tikz{
	\begin{scope}[frontier/.style={distance from root=92.5pt}]
	\Tree 
	[. Word(wakitAbi)
	  [.Pre(wa) \vphantom{k}w a ]
	  [.Stm(kitAb) 
	    [ .\node(A){Root(k,t,b)}; ]
	    [ .\node(B){Template(i,A)}; ]
	  ]
	  [.Suf(i) \vphantom{k}i ]
	]
	\end{scope}
	\begin{scope}[every node/.style={minimum size=2.8ex},xshift=-30pt, yshift=-88pt, start chain, node distance=0.1mm]
	\node[on chain](k) {\vphantom{k}k};
	\node[on chain](i1) {\vphantom{k}i};
	\node[on chain](t) {\vphantom{k}t};
	\node[on chain](a2) {\vphantom{k}A};
	\node[on chain](b) {\vphantom{k}b};
	\end{scope}
	\begin{scope}[dashed]
	\draw (k)--(A.south);
	\draw (i1.north)--(B.south);
	\draw (t.north)--(A.south);
	\draw (a2.north)--(B.south);
	\draw (b.north)--(A.south);
	\end{scope}
	}

%% file: ch-pysrcag/tex/modelrules.tex
\begin{tabular}{rl}
\cfgone $=$  & {$\left\{%
    \begin{aligned}%
      \nt{Word}      &\arrowg \nta{Pre}\; \nta{Stem}\; \nta{Suf} \mid
                              \nta{Pre}\; \nta{Stem}             \mid 
                              \nta{Stem}\;\nta{Suf}              \mid 
                              \nta{Stem} \\
      \nta{Pre} \mid &\nta{Stem} \mid \nta{Suf} \arrowg \nt{Char}^{+} \\
    \end{aligned}%
    \right.$} \\ \midrule
\multirow{4}{*}{\cfgmulti$=$} & \cfgone $\cup$\ \ldots \\ & {$\left\{%
    \begin{aligned}%
      \nt{Word} &\arrowg \nt{Pres}\; \nta{Stem}\; \nt{Sufs} \mid
                         \nt{Pres}\; \nta{Stem}             \mid 
                         \nta{Stem}\;\nt{Sufs}              \\
      \nt{Pres} &\arrowg \nta{Pre}^{+} \\
      \nt{Sufs} &\arrowg \nta{Suf}^{+} \\
    \end{aligned}%
    \right.$} \\ \midrule
\tplonetrilit $=$& {$\left\{%
  \begin{aligned} %
    \nta{Stem}(abcdef)    &\arrowg \nta{R3}(a,c,e)\; \nt{T3}(b,d,f) \\
    \nta{Stem}(abcde)     &\arrowg \nta{R3}(a,c,e)\; \nt{T2}(b,d)   \\
    \nta{Stem}(abcd)      &\arrowg \nta{R3}(a,c,d)\; \nt{T1}(b)     \\
    \nta{Stem}(abc)       &\arrowg \nta{R3}(a,b,c)                  \\%
    \nta{R}n \mid \nt{T}n &\arrowg \text{(recursion described
      in~\S\ref{sec:mcfg:complex-stem-rules})}                      \\
  \end{aligned}%
  \right.$} \\ \midrule
\tplonetrilitTFour $=$ & \tplonetrilit $\cup$\ %
  $\left\{\nta{Stem}(abcdefg) 
            \arrowg \nta{R3}(b,d,f)\; \nt{T4}(a,c,e,g)\right\}$  \\  \midrule
\tplonetrilitTtoThree $=$ & \tplonetrilit $\cup$\ %
  $\left\{\nt{T1} \arrowg  \nt{Char}\; \nt{Char}\mid
                           \nt{Char}\; \nt{Char}\; \nt{Char} \right\}$ \\ \midrule
\tplquad $=$ & \tplonetrilitTFour $\cup$\ %
    $\left\{\nta{Stem}(abcdefgh) 
            \arrowg \nta{R4}(b,d,f,h)\; \nt{T4}(a,c,e,g)\right\}$  \\  \midrule
\tplmultitrilit $=$& \cfgmulti $\cup$ \tplonetrilit \\ \midrule
\end{tabular}

%% file: ch-pysrcag/tex/segmentation-results.tex
\pgfplotstabletypeset[col sep=comma,header=false,
  every head row/.style={ before row=\toprule,after row=\midrule},
  every last row/.style={after row=\bottomrule},
  every row no 1/.style={after row=\midrule},
  numCol/.style={numeric type, precision=1, fixed, zerofill},
  display columns/0/.style={string type, column type=l, column name={} },%
  display columns/1/.style={column name={\bwvoc}, numCol},%
  display columns/2/.style={column name={\bw}, numCol},%
  display columns/3/.style={column name={\heb}, numCol},%
  display columns/4/.style={string type, column type=l, column name={} },%
  display columns/5/.style={column name={\qu}, numCol, assume math mode=false},%
]{ 
\morfessor,            55.57, 40.04, 24.20, \quad \morfessor,              44.34
\cfgone,               47.36, 64.22, 60.05, \quad \cfgmulti,               19.64
\tplonetrilit,         60.42, 71.91, 77.26, \quad \tplmultitrilit,         22.53
\tplonetrilitTtoThree, 60.52, 72.20, 77.41, \quad \tplmultitrilitTtoThree, 25.72
\tplonetrilitTFour,    64.49, 71.59, 77.14, \quad \tplmultitrilitTFour,    24.81
\tplquad,              74.54, 73.66, 78.59, ,
}

%% file: ch-pysrcag/tex/extraction-example.tex
\def\jEM{\mathrm{exact\_match}}%
\begin{figure}[t]
Let $w=$abcdef and the inferred parse tree $\tilde{t}(w)=\Bigg($
\input{tex/extraction-example-tree}
$\Bigg)$.

\begin{tabular}{cll}
  & Method~A &  Method~B \\ \cmidrule{2-3}
  & $\tilde{h}^A_3(\tilde{t}(w))=\{acd\}$ & $\tilde{h}^B_3(\tilde{t}(w))=\{acd,bef\}$ \\
  & $\tilde{h}^A_4(\tilde{t}(w))=\emptyset$ & $\tilde{h}^B_4(\tilde{t}(w))=\emptyset$ \\ \midrule
  \multirow{2}{*}{Suppose $g_3(w)=\{acd\}$.} &
   $\tilde{h}^A_3 \cap g_3 = \{acd\}$ & $\tilde{h}^B_3 \cap g_3 = \{acd\}$  \\
  & $\therefore \jEM_3(w)=1$ & $\therefore \jEM_3(w)=1$  \\ \midrule
  \multirow{2}{*}{Suppose $g_3(w)=\{bef\}$.} &
   $\tilde{h}^A_3 \cap g_3 = \emptyset $ & $\tilde{h}^B_3 \cap g_3 = \{bef\}$  \\
  & $\therefore \jEM_3(w)=0$ & $\therefore \jEM_3(w)=1$  \\ \bottomrule
\end{tabular}
\caption[Illustration of root extraction methods]{%
  Illustration of the root extraction methods and scoring descibed in
  \S\ref{sec:pymcfg:root-ident-method}. For brevity, we present a single
  learning outcome for the word $w$ (above table) and consider two alternative gold
  standard references $g_3(w)$ (bottom two rows). 
}
\label{tbl:mcfg:extract-example}
\end{figure}

%% file: ch-pysrcag/tex/extraction-example-tree.tex
\begin{minipage}[c]{3.5cm}
\scalebox{0.85}{ %
\begin{tikzpicture}
\Tree 
[.\node{}; \edge[dashed];
[.\node{ \textbf{Stem} abcdef };
[.\node{\textbf{R3} a $\cdot$ c $\cdot$ d}; \edge[roof,dashed]; \node{};]
    [.\node{\textbf{T3} b $\cdot$ e $\cdot$ f};  \edge[roof,dashed]; \node{};]
   ]   ]  
\end{tikzpicture}
}%
\end{minipage}

%% file: ch-pysrcag/tex/root-ident-results.tex
\pgfplotstabletypeset[col sep=comma,header=false,
  every row no list/.code 2 args={
          \pgfplotsforeachungrouped \rownumber in {#1}     {
              \pgfplotsset{/pgfplots/table/every row no \rownumber/.append style={#2}}
          }
      },
  every head row/.append style={
        before row/.add={\toprule}{},
        after row/.add={\midrule}{%
        \multicolumn{4}{l}{\emph{Random baselines}}\\
        },
  },
  every row no 3/.append style={after row={%
     \multicolumn{4}{l}{\emph{Method~A: Extract \nta{R3}}} \\
  }},
  every row no 7/.append style={after row={%
     \multicolumn{4}{l}{\emph{Method~B: Extract \nta{R3} and \nt{T3}}} \\
  }},
  every last row/.append style={after row/.add={\midrule}{\midrule%
    \arraybackslash%
    Evaluated ($N'$) 
     & $27783$ 
     & $27494$ 
     & $2738$  
    \\
    \bottomrule
}},
  numCol/.style={numeric type, precision=1, fixed, zerofill, column type=r},
  display columns/0/.style={string type, column type=l, column name={} },%
  display columns/1/.style={column name={\bwvoc}, numCol},%
  display columns/2/.style={column name={\bw}, numCol},%
  display columns/3/.style={column name={\heb}, numCol},%
  every row no list={0,1,2,3,4,5,6,7,8,9,10,11}{before row=\quad},
]{ 
Uniform guesser,       1.468,         8.618,      3.738
Stem-constrained,      6.574,         43.634,     14.19
Character-constrained, 10.258,        10.346,     30.336
Doubly constrained,    41.74,         45.152,     79.912
\tplonetrilit,         0.76,          66.62,      72.9
\tplonetrilitTtoThree, 0.74,          64.5,       9.17
\tplonetrilitTFour,    0.66,          67.06,      74.18
\tplquad,              0.89,          63.98,      69.54
\tplonetrilit,         17.18,         66.91,      72.94
\tplonetrilitTtoThree, 0.76,          64.53,      9.17
\tplonetrilitTFour,    19.79,         67.38,      74.65
\tplquad,              13.91,         64.05,      69.54
}

%% file: ch-pysrcag/tex/root-ident-results-quad.tex
\pgfplotstabletypeset[col sep=comma,header=false,
  every row no list/.code 2 args={
          \pgfplotsforeachungrouped \rownumber in {#1}     {
              \pgfplotsset{/pgfplots/table/every row no \rownumber/.append style={#2}}
          }
      },
  every head row/.append style={
        before row/.add={\toprule}{},
        after row/.add={\midrule}{%
        \multicolumn{4}{l}{\emph{Random baselines}}\\
        },
  },
  every row no 3/.append style={after row={%
     \multicolumn{4}{l}{\tplquad} \\
}},
  every last row/.append style={after row/.add={\midrule}{%
    \arraybackslash%
    Evaluated ($N'$) 
      & $2295$
      & $2276$
      & $19$  
    \\
    \bottomrule
}},
  numCol/.style={numeric type, precision=1, fixed, zerofill, column type=r},
  display columns/0/.style={string type, column type=l, column name={} },%
  display columns/1/.style={column name={\bwvoc}, numCol},%
  display columns/2/.style={column name={\bw}, numCol},%
  display columns/3/.style={column name={\heb}, numCol},%
  every row no list={0,1,2,3,4,5,6}{before row=\quad},
]{ 
Uniform guesser,                        0.696,         8.408,      2.104
Stem-constrained,                       2.72,          40.658,     3.158
Character-constrained,                  10.526,        10.816,     15.792
Doubly constrained,                     39.382,        40.562,     66.314
\emph{Method A:} Extract \nta{R4},                       0.09,          10.37,      0
\emph{Method B:} Extract \nta{R4} and \nt{T4},           14.07,         10.37,      0
}

%% file: ch-pysrcag/tex/trees.tex
\begin{tabular}{c   l@{\hspace{1.5cm}}  c}
  \cfgone parse && \tplonetrilitTFour parse \\ \cmidrule(lr){1-1} \cmidrule(lr){3-3}
\begin{minipage}[b]{3.2cm}
\scalebox{0.85}{ %
\begin{tikzpicture}
\Tree [.Word
          [.Pre \edge[roof]; {w l $>$} ]
          [.Stem \edge[roof]; {s t A r} ]
        [.Suf \edge[roof]; {k m}  ] ] 
\end{tikzpicture}
}\vspace{8pt}
\end{minipage}
&&
\begin{minipage}{4.5cm}
  \vspace{1ex}
\scalebox{0.85}{ %
\begin{tikzpicture}
\Tree 
          [.\node{\textit{Pre(w l)} ... \textbf{Stem} ... \textit{Suf(k m)}};
            [.\node{\textbf{T2} $>$ $\cdot$ A};
              [.\node{\textbf{T1}}; \edge[roof]; { $>$ }  ]                                 
              [.\node{\textbf{T1}}; \edge[roof]; { A } ] ]                               
            [.\node{\textbf{X2} s  t $\cdot$ r};
              [.\node{\textbf{R3} s $\cdot$ t $\cdot$ r};
                [.\node{\textbf{R2} s $\cdot$ t};
                  [.\node{\textbf{R1}}; \edge[roof]; { s }   ]                             
                  [.\node{\textbf{R1}};  \edge[roof]; { t }  ] ]                           
                [.\node{\textbf{R1}};  \edge[roof]; { r }  ] ] ] ]                        

\end{tikzpicture}
} %
\end{minipage}
\end{tabular}

%% file: ch-pysrcag/tex/trees_bwvoc.tex
\begin{tabular}{c   l@{\hspace{1.5cm}}  c}
  \cfgone parse && \tplquad parse \\ \cmidrule(lr){1-1} \cmidrule(lr){3-3}
\begin{minipage}{3.1cm}
\scalebox{0.85}{ %
\begin{tikzpicture}
\Tree [.Word
[.Pre \edge[roof]; {l i d a} ]
[.Stem \edge[roof]; {n o t i l l A} ]
] 
\end{tikzpicture}
} 
\end{minipage} %
&&
\begin{minipage}{6.5cm}
\scalebox{0.85}{ %
\begin{tikzpicture}
\Tree 
[.\node{\textit{Pre(l i)} ... \textbf{Stem}(danotill) ... \textit{Suf(A)}};
[.\node{\textbf{R4} d $\cdot$ n $\cdot$ t $\cdot$ l};
        [.\node{\textbf{R3} d $\cdot$ n $\cdot$ t };
            [.\node{\textbf{R2} d $\cdot$ n };
                [.\node{\textbf{R1}}; \edge[roof]; { d }  ]
                [.\node{\textbf{R1}}; \edge[roof]; { n }  ] ]
            [.\node{\textbf{R1}}; \edge[roof]; { t } ] ]
    [.\node{\textbf{R1}}; \edge[roof]; { l } ] ]
[.\node{\textbf{T4} o $\cdot$ a $\cdot$ i $\cdot$ l};
        [.\node{\textbf{T3} o $\cdot$ a $\cdot$ i };
        [.\node{\textbf{T2} o $\cdot$ a };
        [.\node{\textbf{T1}}; \edge[roof]; { o }  ]
        [.\node{\textbf{T1}}; \edge[roof]; { a }  ] ]
        [.\node{\textbf{T1}}; \edge[roof]; { i } ] ]
        [.\node{\textbf{T1}}; \edge[roof]; { l } ] ] ]
\end{tikzpicture}
}
\end{minipage}
\end{tabular}

%% file: ch-pysrcag/tex/heb_samples.tex
\begin{tabular}{rc ll l l} \toprule 
  && 
  && \multicolumn{2}{c}{\bf Mistaken Examples} 
\\ \cmidrule(lr){5-6}
  $P(s,R3)$ & \textbf{Root} $s$
  & \multicolumn{2}{c}{\bf Correct Examples} & Bad Segmentation & Bad Root 
\\ \midrule
0.00649 & spr  \checkmark
   & 
     \HGRT{{s}}a\HGRT{{p}}a\HGRT{{r}}\segTP{}ti 
   &
     ye\segTP{}\HGRT{{s}}a\HGRT{{p}}\HGRT{{r}}\segTP{}u 
   & 
      me\segFP{}\HGRT{{s}}a\HGRT{{p}}\HGRT{{r}}\segTP{}im
   &
      hi\segFP{}\HGRT{{s}}\HRT{t}a\HGRT{{p}}\segFP{}a\GRT{r}\segFN{}t
\\ 
0.00527 & lbs \checkmark
   & 
     \HGRT{{l}}a\HGRT{{b}}a\HGRT{{S}}\segTP{}t 
   &
     ti\segTP{}\HGRT{{l}}\HGRT{{b}}e\HGRT{{S}}\segTP{}i
   & 
      le\segFN{}ha\segFP\HGRT{{l}}\HGRT{{b}}i\HGRT{{S}} 
   &
\\
0.00524 & ptx \checkmark 
   & 
     \HGRT{{p}}a\HGRT{{t}}a\HGRT{{x}}\segTP{}ti 
   &
     ti\segTP{}\HGRT{{p}}\HGRT{{t}}e\HGRT{{x}}\segTP{}i
   & 
     ni\segFP{}\HGRT{{p}}\HGRT{{t}}a\HGRT{{x}}\segTP{}at
   &
     li\segTP{}\HGRT{{p}}\GRT{t}\HRT{o}a\HGRT{{x}}  
\\
0.00521 & rxc \checkmark 
   & 
     yi\segTP{}\HGRT{{r}}\HGRT{{x}}a\HGRT{{c}} 
   &
     \HGRT{{r}}o\HGRT{{x}}e\HGRT{{c}}\segTP{}et 
   & 
     le\segFN{}hit\segFP{}\HGRT{{r}}a\HGRT{{x}}e\HGRT{{c}}
   &
\\
0.00510 & !al $\times$ 
  &&
  & 
     le\segFN{}ha\segFP{}\HRT{!al}\segFP{}ot
  &
     \HGRT{{!}}\HRT{a}\GRT{c}\HGRT{{l}}\segFP{}an\segFN{}it
\\
\bottomrule
\end{tabular}    

%% file: conclusions/conclusions.tex
\chapter{\label{ch:conclusion}Conclusion} 

This dissertation presented original work on the problem of capturing sub-word
structure in probabilistic models of language. The central
approach was to use the essence of certain morphological processes as a
basis for extending three different types of models in order to make them better suited
to morphologically rich languages. 
We applied this approach both to language models (LMs) which define
probabilities of word sequences, and to unsupervised models that seek to induce
the morphemes of individual words from unannotated text.

In \autoref{ch:hpylm} we presented a hierarchical Bayesian model of word sequences which accounts for
productive compound formation. 
We exploited the right-head\-ed\-ness of German compounds to structure the
statistical dependencies between a compound word and its immediately preceding
context, while using an embedded language model to generate the modifiers within
each compound independently of the word's context.
The main findings from our evaluation were as follows:
\begin{enumerate}
  \item The compounding language model succeeded in reducing the data sparsity
    due to compounding by expressing compounds in terms of their parts---in our
    data set, the number of distinct elements in the base distribution at the
    root of the hierarchical model is reduced to less than half the number of
    unsegmented word types.
    The model obtained small improvements in test set
    perplexity compared to a baseline language model that views compounds as
    atomic elements.
  \item The use of the compounding language model as a feature in a machine
    translation system led to translations that were judged to be equivalent
    in overall quality with that from a baseline language model, as measured with \bleu.
    Finer-grained measurement showed that the compounding model did, however,
    improve the translation system's precision in outputting compound words,
  \item Structuring the statistical dependencies according to the
    right-headedness of German compounds played a key role in the aforementioned
    outcomes. This was established by observing a degradation in perplexity and
    \bleu when using a model variant that posits left-headedness.
\end{enumerate}

In \autoref{ch:dist} we shifted the attention to another type of language
model to consider how sub-word structure could be leveraged in another setting.
Based on automatically induced distributed feature representations of words and
smooth scoring functions, distributed language models (DLMs) circumvent the need for
smoothing by means of back-off as employed in traditional \ngram LMs, and as we
refined further in the compounding model. Our contribution in this regard is two-fold: We show that DLMs can benefit substantially from
explicitly modelling sub-word structure, and we determine that it is possible
and beneficial to use a normalised DLM as part of a machine translation decoder.

We formulated a variant of the log bilinear (LBL) language model
\citep{Mnih2012} that incorporates a simple method of composing word vectors
from morpheme vectors. Morpheme vectors are learnt as part of the model,
leveraging a separate unsupervised model to obtain the morphological
segmentation itself.
The method is inspired directly by recent positive results in modelling compositional morphology
\citep{Luong2013,Lazaridou2013} and the observation that word-based DLMs already
capture non-trivial, morphologically relevant regularities \citep{Mikolov2013a}.
The main findings from our monolingual evaluation are as follows:
\begin{enumerate}
  \item The morpheme-based LBL achieved substantial reductions in test set
    perplexity for six languages of varying morphological complexity.
    The model affects a soft tying of parameters across words that share
    morphemes, which contributed to the largest perplexity
    improvements being generally on rarer words. This is an important outcome for
    modelling morphologically richer languages, where rare words tend to form a
    larger portion of the vocabulary.
  \item The availability of morpheme vectors allowed for the composition of
    high-quality word vectors for out-of-vocabulary words, as measured in a word
    similarity rating task on multiple individual languages.
\end{enumerate}

The integration of correctly normalised LBLs into a machine translation decoder
was made tractable by partitioning the vocabulary into word classes and
decomposing the probability model accordingly. This is a known method of
decreasing the computational cost involved in computing probabilities with DLMs.
To our knowledge, this work is however the first to deploy it successfully
in translation---previous uses of DLMs in translation have been limited to rescoring
n-best lists or lattices, or have used unnormalised DLMs.
Our evaluation findings were as follows:
\begin{enumerate}
  \item Use of a LBL LM feature in the log-linear translation model
    (in addition to a modified Kneser-Ney (MKN) LM feature) provided consistent
    improvements in \bleu when translating into six languages of varying
    morphological complexity,
    compared to a baseline system using only the MKN LM.
  \item The morpheme-based LBL provided further small increases in \bleu, though
    the improvements cannot be separated entirely from the instability of the
    optimiser used to tune the translation model feature weights.
\end{enumerate}

As a third and final instance of capturing sub-word structure in probabilistic
models of language, we switched perspective to the question of \emph{acquiring}
such sub-word structure in the first place, rather than relying on an external
source as we did in the sequence-based models.  Our contribution in this regard
is to suggest that simple Range Concatenating Grammars (SRCGs) are useful for
encoding both concatenative and non-concatenative morphological processes in a
single formalism. The grammar-based approach allows for unsupervised learning to
be done using the non-parametric Bayesian framework of adaptor grammars, which
we instantiate for the SRCG formalism.
We applied this approach to Semitic morphology, where words are often formed by
first interspersing the characters of root morphemes into templatic structures
and then attaching further affixes in a concatenative fashion. As such, it
offers a useful test-bed for the notion that joint modelling of concatenative
and non-concatenative morphological processes could have 
a mutually disambiguating effect in morphological analysis.
Our evaluation gave rise to the following main findings:
\begin{enumerate}
  \item Modelling discontiguous sub-word structure improved the segmentation of
    Arabic and Hebrew words into linear morphemes, as measured by the F-score
    in comparison to baseline grammars that ignore discontiguous
    sub-structure.
    This holds for both orthographic variants of our Arabic data set (with and
    without vocalisation).
  \item A simple method for extracting root morphemes from the inferred
    parse trees of words produced mixed results. On two of the data sets it
    correctly identified about two-thirds of triliteral roots for the words
    where such roots are expected, but for the other data sets and for
    the identification of quadriliteral roots in all our data sets, the accuracy
    was low.
  \item The intercalating adaptor grammars also produced drastic F-score
    increases for the lexica of stems induced from data, relative to the purely
    concatenating baseline grammars. This is related to the improved
    segmentation performance.
  \item Considered as binary classifiers of strings, the intercalating adaptor
    grammars obtained moderately high performance in discriminating between
    string tuples that are triliteral roots and those that are not.
\end{enumerate}
A characteristic of our unsupervised morphology learning work was that we erred on
the side of underspecification when it came to engineering rules to
fit the data sets and languages used, and this is reflected in the mixed results.
Our primary intention was to test the extent to which a high-level notion of
non-concatenative morphology could reasonably be expressed in adaptor grammars.
This paves the way for interesting future applications, and creates a framework
within which it is straightforward to encode finer-grained knowledge of
particular non-concatenative processes.

In conclusion, this dissertation showed that the explicit inclusion of sub-word
structure in probabilistic language models is beneficial for morphologically
complex languages in settings ranging from unsupervised morphology learning and
word similarity rating to the estimation of word-sequence probabilities and
their application in machine translation.

\section{Extensions and further research}
Possible directions for further research emanating from the work presented in this
dissertation include the following: 
\begin{enumerate}
  \item The LMs developed in Chapters \ref{ch:hpylm} and \ref{ch:dist} are in
    principle capable of providing probabilities for OOV words by leveraging
    their sub-word representations encoded in the models. For a machine
    translation system to benefit maximally from that ability, the system itself
    also requires the ability to hypothesise novel word forms in its translation
    output. An interesting direction for future research is to combine our LMs
    with the 
    recently proposed methodology of \citet{Chahuneau2013SynthPhrases}
    where novel inflections of words can be hypothesised by the translation
    system. 
  \item Both LMs made use of deterministic segmentations of words into
    sub-parts. An alternative would be to learn the segmentation jointly with the rest
    of the model, \eg as done by \cite{Mochihashi2009a} in the context of Chinese
    word segmentation. 
  \item The method from Chapter \ref{ch:dist} for composing word vectors from
    morpheme vectors through addition is readily implementable in bilingual or
    multilingual distributed models \citep[\eg][]{ZouSocher2013}.
    Aside from the parameter tying effect that should also be useful for rare
    words in that setting, an interesting question is whether a morpheme-based
    bilingual model can capture the tendency of a given language to encode
    certain grammatical properties with distinct words while another language
    uses bound morphemes for the same. 
  \item Our approach to modelling discontiguous sub-word structure in Chapter
    \ref{ch:pymcfg} presents interesting opportunities for extending the line of
    work that approaches unsupervised morphology learning from the perspective of
    language acquisition. \citet{Frank2013} recently considered the mutually
    disambiguating roles of morphology and syntax in this context, and
    a similar study on languages employing non-concatenative morphology could
    benefit from our work.
  \item The use of SRCGs in Chapter \ref{ch:pymcfg} affords modelling
    convenience but comes in exchange for high polynomial parsing time
    complexity. 
    As with adaptor grammars generally, the emphasis is on being
    able to explore different model assumptions easily. 
    Yet full recursive power is not necessary for many morphological processes,
    so that a relevant question 
    is whether finite-state transformations or approximations
    could be used to cast a given adaptor grammar into a
    more efficient computational framework. There is a rich literature on this
    for context-free grammars \citep[\eg][]{Nederhof2000}.

  \item The burden of defining the appropriate set of rewrite
    rules for the base distribution grammar could be overcome by using a semi-supervised
    approach that searches for optimality over different instantiations of
    so-called metagrammars \citep{Sirts2013}.

\end{enumerate}

%% file: preamb/backmatter.tex
%
%
\setlength{\baselineskip}{16pt}
	
\bibliography{library_filtered} 
\bibliographystyle{myplainnat}